\documentclass[sn-basic]{sn-jnl}

\usepackage{graphicx}
\usepackage{multirow}
\usepackage{amsmath,amssymb,amsfonts}
\usepackage{amsthm}
\usepackage{mathrsfs}
\usepackage[title]{appendix}
\usepackage{xcolor}
\usepackage{textcomp}
\usepackage{manyfoot}
\usepackage{booktabs}
\usepackage{algorithm}
\usepackage{algorithmicx}
\usepackage{algpseudocode}
\usepackage{listings}
\usepackage{array,colortbl}

\usepackage{listings}
\usepackage[normalem]{ulem}
\usepackage{xspace}
\usepackage{float}
\usepackage{tikz}
\usepackage{verbatim}
\usepackage{tabularx}
\usepackage{enumitem}
\usepackage{wrapfig}
\usepackage{multicol}
\usepackage[most]{tcolorbox}
\usepackage[framemethod=tikz]{mdframed}
\usepackage{subcaption}
\usepackage{caption}
\usepackage{multirow}
\usepackage{pifont}
\usepackage{epigraph}
\usepackage{bm}
\usepackage{bbding}
\usepackage{makecell}
\usepackage{colortbl}

\definecolor{mygray}{gray}{.9}
\definecolor{mypink}{rgb}{.99,.91,.95}
\definecolor{mycyan}{cmyk}{.3,0,0,0}
\DeclareMathOperator*{\argmax}{argmax}
\usepackage{tikz}
\usepackage[edges]{forest}
\definecolor{hidden-draw}{RGB}{205, 44, 36}
\definecolor{hidden-blue}{RGB}{194,232,247}
\definecolor{hidden-orange}{RGB}{243,202,120}
\definecolor{hidden-yellow}{RGB}{242,244,193}
\definecolor{tree-level-1}{RGB}{245,20,85}
\definecolor{tree-level-2}{RGB}{246,86,118}
\definecolor{tree-level-3}{RGB}{248,177,193}
\definecolor{tree-leaf}{RGB}{176,230,198}

\definecolor{Self}{RGB}{255,0,128}
\definecolor{Ensemble}{RGB}{0,127,255}
\definecolor{Iterative}{RGB}{153,51,255}

\definecolor{exemplar1}{RGB}{136,98,148}
\definecolor{exemplar2}{RGB}{148,210,242}
\definecolor{knowledge1}{RGB}{249,219,152}
\definecolor{knowledge2}{RGB}{255,245,220}

\newcolumntype{C}[1]{>{\centering}m{#1}}
\usepackage{subcaption}
\usepackage{pgfplots}
\pgfplotsset{compat=1.17}
\usetikzlibrary{pgfplots.groupplots}

\usepgfplotslibrary{colorbrewer} 

\usepgfplotslibrary[colorbrewer] 

\usetikzlibrary{pgfplots.colorbrewer} 

\usetikzlibrary[pgfplots.colorbrewer]

\newtheorem{definition}{Definition}%

\begin{document}

\title[]{A Survey of Reasoning with Foundation Models}

\author[1]{\fnm{Jiankai} \sur{Sun}}

\author[1]{\fnm{Chuanyang} \sur{Zheng}}

\author[2]{\fnm{Enze} \sur{Xie$^\mathsection$}}

\author[2]{\fnm{Zhengying} \sur{Liu$^\mathsection$}}

\author[1]{\fnm{Ruihang} \sur{Chu}}

\author[1]{\fnm{Jianing} \sur{Qiu}}

\author[1]{\fnm{Jiaqi} \sur{Xu}}

\author[3]{\fnm{Mingyu} \sur{Ding}}

\author[4]{\fnm{Hongyang} \sur{Li}}

\author[1]{\fnm{Mengzhe} \sur{Geng}}

\author[2]{\fnm{Yue} \sur{Wu}}

\author[1]{\fnm{Wenhai} \sur{Wang}}

\author[2,6]{\fnm{Junsong} \sur{Chen}}

\author[11]{\fnm{Zhangyue} \sur{Yin}}

\author[2]{\fnm{Xiaozhe} \sur{Ren}}

\author[5]{\fnm{Jie} \sur{Fu}}

\author[5]{\fnm{Junxian} \sur{He}}

\author[1]{\fnm{Wu} \sur{Yuan}}

\author[3]{\fnm{Qi} \sur{Liu}}

\author[3]{\fnm{Xihui} \sur{Liu}}

\author[1]{\fnm{Yu} \sur{Li}}

\author[7]{\fnm{Hao} \sur{Dong}}

\author[1]{\fnm{Yu} \sur{Cheng}}

\author[7]{\fnm{Ming} \sur{Zhang}}

\author[1]{\fnm{Pheng Ann} \sur{Heng}}

\author[8,4]{\fnm{Jifeng} \sur{Dai}}

\author[3,4]{\fnm{Ping} \sur{Luo}}

\author[9]{\fnm{Jingdong} \sur{Wang}}

\author[10]{\fnm{Ji-Rong} \sur{Wen}}

\author[11]{\fnm{Xipeng} \sur{Qiu}}

\author[5]{\fnm{Yike} \sur{Guo}}

\author[12]{\fnm{Hui} \sur{Xiong}}

\author[2]{\fnm{Qun} \sur{Liu}}

\author[2]{\fnm{Zhenguo} \sur{Li}}

\affil[1]{The Chinese University of Hong Kong}

\affil[2]{\orgname{Huawei Noah's Ark Lab}}

\affil[3]{\orgname{The University of Hong Kong}}

\affil[4]{\orgname{Shanghai AI Lab}}

\affil[5]{\orgname{Hong Kong University of Science and Technology}}

\affil[6]{\orgname{Dalian University of Technology}}

\affil[7]{\orgname{Peking University}}

\affil[8]{\orgname{Tsinghua University}}

\affil[9]{\orgname{Hefei University of Technology}}

\affil[10]{\orgname{Renmin University of China}}

\affil[11]{\orgname{Fudan University}}

\affil[12]{\orgname{Hong Kong University of Science and Technology (Guangzhou)}}

\footnotetext{Project Lead. Email: \url{{xie.enze, liuzhengying2}@huawei.com}}

\abstract{

Reasoning, a crucial ability for complex problem-solving, plays a pivotal role in various real-world settings such as negotiation, medical diagnosis, and criminal investigation. It serves as a fundamental methodology in the field of Artificial General Intelligence (AGI). With the ongoing development of foundation models, e.g., Large Language Models (LLMs), there is a growing interest in exploring their abilities in reasoning tasks. In this paper, we introduce seminal foundation models proposed or adaptable for reasoning, highlighting the latest advancements in various reasoning tasks, methods, and benchmarks. We then delve into the potential future directions behind the emergence of reasoning abilities within foundation models. We also discuss the relevance of multimodal learning, autonomous agents, and super alignment in the context of reasoning. By discussing these future research directions, we hope to inspire researchers in their exploration of this field, stimulate further advancements in reasoning with foundation models, and contribute to the development of AGI.
\footnote{We maintain a continuously updated reading list to benefit future research, featuring relevant papers and popular benchmarks on reasoning. GitHub: \url{https://github.com/reasoning-survey/Awesome-Reasoning-Foundation-Models}}~\footnote{Preliminary release. We are committed to maintaining the quality and recency of this work.}
}


\maketitle
\newpage
\tableofcontents
\newpage
\section{Introduction}\label{sec:intro}

\epigraph{\textit{``Humans have always done nonmonotonic reasoning, but rigorous monotonic reasoning in reaching given conclusions has been deservedly more respected and admired.''}}{John McCarthy (2004)}

\begin{figure}[tbp]
	\begin{center}
	
\includegraphics[width=\columnwidth]{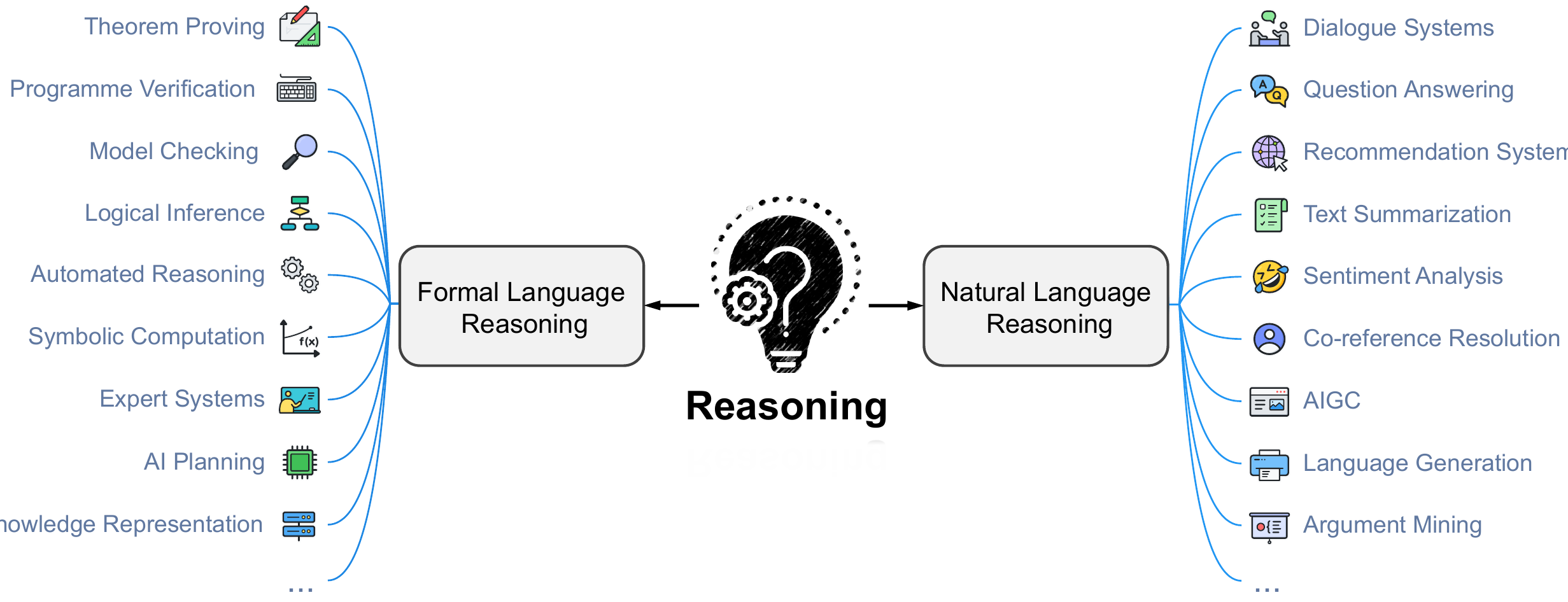}
	\end{center}
	\caption{Two broad types of language reasoning and examples of the supported tasks.}
\label{fig:intro_twocls}
\end{figure}

Reasoning is an essential aspect of artificial intelligence, with applications spanning various fields, such as problem-solving, theorem proving, decision-making, and robotics~\citep{manning2022human}. 
\emph{Thinking, Fast and Slow}~\citep{daniel2017thinking} elucidates a dual-system framework for the human mind, consisting of ``System 1'' and ``System 2'' modes of thought. ``System~1'' operates rapidly, relying on instincts, emotions, intuition, and unconscious processes. In contrast, ``System 2'' operates slower, involving conscious deliberation such as algorithmic reasoning, logical analysis, and mathematical abilities. Reasoning plays a crucial role as one of the key functions of ``System~2''~\citep{bengio2017consciousness,weston20232}.
Reasoning can be categorized into two broad types: formal language reasoning and natural language reasoning~\citep{reiter1975formal,berzonsky1978formal,teig2016bringing,yu2023natural,zhao2023survey,li2023diffusion}. On one hand, as shown in Figure~\ref{fig:intro_twocls}, formal language reasoning is often used in areas like formal verification of software and hardware systems, theorem proving and automated reasoning~\citep{reiter1975formal,berzonsky1978formal}. On the other hand, natural language reasoning enables more intuitive human-computer interactions and supports tasks like question answering~\citep{shao2023prompting,jiang2021can}, information retrieval~\citep{zhu2023large,ai2023information}, text summarization~\citep{liu2023learning}, and sentiment analysis~\citep{yu2023natural,araci2019finbert,barbieri2021xlm}.

\begin{figure}[tbp]
	\begin{center}
		
\includegraphics[width=1.0\columnwidth]{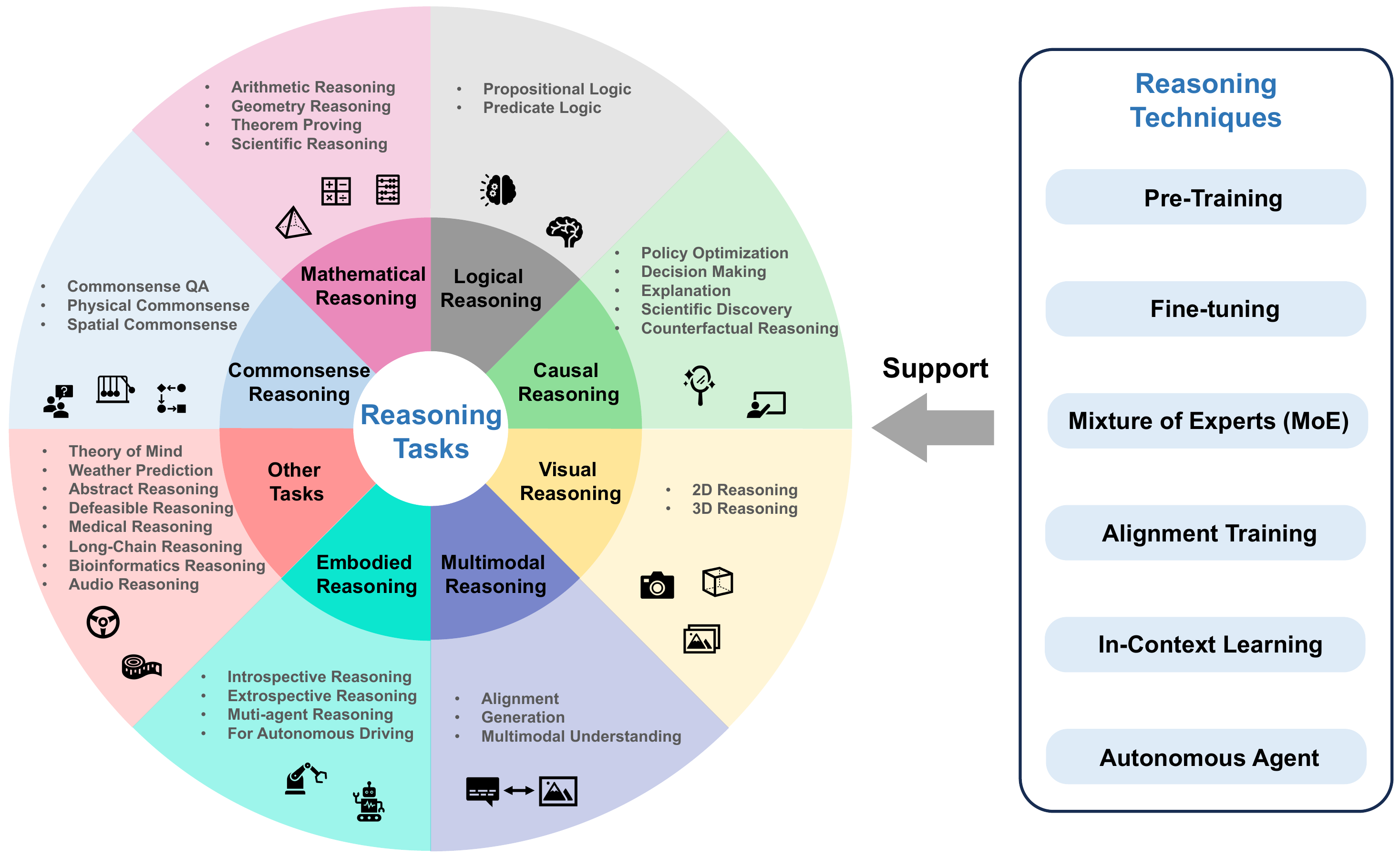}
	\end{center}
	\caption{Left: Overview of the reasoning tasks introduced in this survey, as detailed in Section~\ref{sec:tasks}. Right: Overview of the reasoning techniques for foundation models, as detailed in Section~\ref{sec:technique}.}
\label{fig:intro_overview}
\end{figure}

Since their inception, foundation models~\citep{bommasani2021opportunities} have demonstrated remarkable efficacy across various domains, including natural language processing~\citep{qiao2022reasoning}, computer vision~\citep{wang2023review}, and multimodal tasks~\citep{li2023large}. However, the burgeoning interest in general-purpose artificial intelligence has sparked a compelling debate regarding whether foundation models can exhibit human-like reasoning abilities. Consequently, there has been a surge of interest in studying the reasoning capabilities of foundation models. While previous surveys have explored the application potential of foundation models from different perspectives~\citep{gu2023systematic,wang2023review,yin2023survey,zong2023selfsupervised,lou2023instruction,charalambous2023new,aligning_llm_human,wang2023interactive,wang2023survey}, there remains a need for a systematic and comprehensive survey that specifically focuses on recent advancements in multimodal and interactive reasoning, which emulates human reasoning styles more closely. Figure~\ref{fig:intro_overview} presents an overview of reasoning with regard to tasks and techniques that this article will discuss.

Foundation models typically consist of billions of parameters and undergo \mbox{(pre-)training} using self-supervised learning~\citep{jain2023bring} on a broad dataset~\citep{bommasani2021opportunities}. Once (pre-)trained, foundation models can be adapted to solve numerous downstream tasks through task-specific fine-tuning, linear probing, or prompt engineering, demonstrating remarkable generalizability and impressive accuracy~\citep{bommasani2021opportunities,qiu2023large}. In contrast to the soft attention mechanisms utilized in conventional transformers, System 2 Attention (S2A) harnesses the capabilities of Large Language Models (LLMs) to facilitate linguistic reasoning. This method improves the factuality and objectivity of long-form content generation.
By integrating logical rules and principles into the learning process~\citep{mao2023agentdriver}, these models can perform complex tasks such as deduction and inference. This allows them to make decisions based on explicit knowledge~\citep{mao2023agentdriver} and logical reasoning, rather than relying solely on statistical patterns~\citep{yang2023logical}. As a rapidly growing field in artificial intelligence research, reasoning with foundation models aims to develop models capable of understanding and interacting with complex information in a more human-like manner. Built upon a foundation of logical reasoning and knowledge representation, these models make it possible to reason about abstract concepts and make decisions based on logical rules.

First, reasoning with foundation models enables the application of prior knowledge and domain expertise. Logical rules can be derived from expert knowledge or formalized from existing ontologies or knowledge graphs. By leveraging this prior knowledge, models can benefit from a better understanding of the problem domain and make more informed decisions.
Second, reasoning with foundation models can enhance the robustness and generalization capabilities. By incorporating the information contained in massive amounts of data, models can better handle situations facing limited data or encountering unseen scenarios during deployment. This enables models to be more reliable and sturdy for robust, real-world usage.

In contrast to current surveys that have primarily focused on specific aspects of foundation models, such as prompts~\citep{qiao2022reasoning}, hallucination ~\citep{rawte2023survey}, deductive reasoning~\citep{huang2022towards}, logical reasoning~\citep{friedman2023large,yang2023logical}, causal reasoning~\citep{kiciman2023causal, stolfo2022causal}, health informatics~\citep{qiu2023large}, or AI agents~\citep{xi2023rise}, this paper takes a broader perspective, aiming to connect various research efforts in this area in a cohesive and organized manner. 
As Figure~\ref{fig:intro_overview} shows, we provide a concise overview of various reasoning tasks, including \textbf{Commonsense Reasoning, Mathematical Reasoning, Logical Reasoning, Causal Reasoning, Visual Reasoning, Audio Reasoning, Multimodal Reasoning, Embodied Reasoning, Defeasible Reasoning, and beyond}. By doing so, we provide a comprehensive overview highlighting the interconnections and relationships between different aspects of the field to inspire more research efforts to actively engage with and further the advances of reasoning with foundation models.

In summary, we have conducted a survey of over 650 papers on foundation models, primarily focusing on research from the past two years.
We discuss different tasks, approaches, techniques, and benchmarks used in these models. We also explore various application domains that can benefit from reasoning with foundation models, such as question-answering, automated reasoning, and knowledge representation. 
We also discuss the challenges and limitations of current reasoning with foundation models and potential directions for future research.
By understanding the advancements and challenges in this field, researchers can explore new avenues for developing intelligent systems that can reason and make decisions in a more human-like and interpretable manner. Overall, this paper aims to provide a comprehensive understanding of reasoning with foundation models, its current state, and future possibilities.

\section{Background}\label{sec:background}
This section introduces background knowledge about foundation models for reasoning. We will delve into key aspects such as what reasoning is, recent progress in general foundation models, the architectural design of foundation models, the training methodologies employed, and the transfer learning paradigm that enables their applications for reasoning tasks. 
By elucidating these fundamental aspects, we hope our readers will understand the underlying principles and techniques driving reasoning with foundation models, setting the stage for the subsequent exploration of recent advancements and methodologies in this field.

\subsection{Definition of Reasoning}\label{sec:definition}

\begin{table*}[]
    \centering
    \begin{tabular}{l|p{19pc}}
\toprule
Context & Lee found the Northeast to be way too
cold. Lee decided to move to Florida.
 \\ \toprule
Question & How would you describe Lee? \\ \midrule
\multirow{3}{*}{Answers} & a) happy \\
& b) likes cold weather \\ 
& \textbf{c) likes the heat} \\
\bottomrule
\end{tabular}
    \caption{An Example of Commonsense Reasoning Problem from Social IQA~\citep{sap-etal-2019-social}. The correct answer is in bold.}
    \label{tab:commonsense_reasoning_example}
\end{table*}

\begin{table*}[]
    \centering
    \begin{tabular}{l|p{19pc}}
\toprule
Problem & A farmer has 3 types of fruits in his garden: apples, oranges, and pears. He has twice as many apples as oranges and three times as many pears as apples. If he has 24 oranges, how many pieces of fruit does he have in total? \\\midrule
Expression & $x = 24 \times 2 + 24 \times 3 \times 2 + 24$ \\\midrule
Solution & 216 \\\bottomrule
\end{tabular}
    \caption{A Sample Math Word Problem (MWP).}
    \label{tab:math_reasoning_example}
\end{table*}

When the term ``reasoning" is brought up, its precise meaning is often unclear to people. To clarify, let us first establish a clear definition of reasoning. ``Reasoning'' is a broad and multifaceted concept that manifests in various contexts. It encompasses cognitive processes and logical thinking employed to analyze information, make deductions, draw conclusions, and formulate coherent arguments. Reasoning can be observed in diverse domains, such as scientific inquiry, problem-solving, decision-making, and everyday discourse. Its fundamental purpose is to enable individuals to connect pieces of information, evaluate relationships, and arrive at informed judgments or solutions. By exploring the different facets and dimensions of reasoning, we can gain a comprehensive understanding of its significance and explore the mathematical formalisms and techniques employed to elucidate and enhance this fundamental aspect of human cognition.

In addition to its broad conceptual nature, the term ``reasoning'' carries specific definitions within various fields. Let us briefly touch upon the definitions of reasoning in the domains of philosophy, logic, and Natural Language Processing (NLP)~\citep{clark2020transformers,huang2022towards,yang2022language,young-etal-2022-abductionrules,yu2023natural}.

\paragraph{Philosophy}
\begin{definition}
(Cognitive reasoning). Cognitive reasoning refers to modeling the human ability to draw meaningful conclusions despite incomplete and inconsistent knowledge involving among others the representation of knowledge where all processes from the acquisition and update of knowledge to the derivation of conclusions must be implementable and executable on appropriate hardware~\citep{furbach2019cognitive}.
\end{definition}
\paragraph{Logic}
\begin{definition}
(Logical reasoning). 
Logical reasoning involves a process of thought where conclusions are methodically drawn based on premises and the relationships between these premises, ensuring that the conclusions are logically implied or necessitated by them~\citep{Nunes2012}.
\end{definition}
\paragraph{NLP}
\begin{definition}
(Natural language reasoning). Natural language reasoning is a process of integrating multiple knowledge (e.g., encyclopedic knowledge and commonsense knowledge) to derive some new conclusions about the (realistic or hypothetical) world. Knowledge can be derived from sources that are both explicit and implicit. Conclusions are assertions or events assumed to be true in the world, or practical actions~\citep{yu2023natural}.
\end{definition}

\begin{table*}[]
    \centering
\begin{tabular}{ll}
        \toprule
        \multicolumn{2}{c}{\textbf{Example}} \\
        \midrule
        Fact1 & This animal is a robin. \\
        Rule & All robins are birds. \\
        Fact2 & This animal is a bird. \\
        \midrule
        \textbf{Reasoning Type} & \textbf{Representation} \\
        \midrule
        Deduction & (Fact1 + Rule $\rightarrow$ \textcolor{red}{Fact2}) \\
        Abduction & (Fact1 + \textcolor{red}{Rule} $\leftarrow$ Fact2) \\
        Induction & (Fact1 + Fact2 $\rightarrow$ \textcolor{red}{Rule}) \\
        \bottomrule
    \end{tabular}
    \caption{Illustration of deductive reasoning, abductive reasoning, and inductive reasoning. In this example, the black text represents the given knowledge, while the red text represents the inferred knowledge. The term ``Fact'' indicates specific information, while ``Rule'' denotes a general principle or guideline.}
\label{tab:deduction_abduction_induction}
\end{table*}

We can also get a better understanding of what reasoning is, by categorizing them from different perspectives, as shown in the next sections.

\subsubsection{Deductive, Abductive, and Inductive Reasoning}

Before delving into recent developments, let us first review the traditional perspectives on reasoning, which categorizes it into three primary types: inductive reasoning, deductive reasoning, and abductive reasoning. This categorization has long been recognized and provides a framework for understanding the different modes of reasoning. By examining each type, we can better understand their distinctive characteristics and applications. So, let us take a closer look at these traditional categories to enhance our comprehension of reasoning processes.

Table~\ref{tab:deduction_abduction_induction} provides an example to explain these three reasoning types, respectively. Deductive reasoning is a logical process that derives specific conclusions from general principles or premises. It follows a top-down approach, starting with general principles and applying logical rules to reach specific conclusions. Deductive reasoning aims to provide logically valid and conclusive results. 

Inductive reasoning involves drawing general conclusions or patterns based on specific observations or evidence. It moves from specific instances to broader generalizations.
Inductive reasoning does not guarantee absolute certainty but provides probable conclusions based on available evidence~\citep{wang2023hypothesis}.

Abductive reasoning is the process of making plausible explanations or hypotheses to account for observed facts or data. It involves inferring the best possible explanation from incomplete or limited information. Abductive reasoning is often used in problem-solving and hypothesis generation.

In commonly used terms of reasoning, for a non-fallacious argument (an argument consisting of a premise and a conclusion)~\citep{Flach2000}, a deductive argument is classified as such when the premise can offer conclusive support for the conclusion. In other words, if all the premises of the argument are true, it would be impossible for the conclusion to be false. On the other hand, an inductive argument is characterized by the premise providing only partial support for the conclusion~\citep{salmon1989scientific}. In the case of inductive arguments, the conclusions extend or surpass the information contained in the premises~\citep{salmon1989scientific}.  Unlike deductive arguments that provide conclusive proof or inductive arguments that offer partial support, abductive arguments aim to provide the most reasonable explanation for a given situation, even if it may not be the only possible explanation.

Typically, in the trio of reasoning types, which includes deduction, abduction, and induction, the most extensively studied and explored is deduction, while research on abduction and induction has remained relatively limited and under-explored~\citep{Flach2000,yang2023logical}. 
Encouragingly, progress has been made recently in the field of inductive reasoning. \citet{sinha2019clutrr} propose the CLUTRR dataset for classifying kinship relations in short stories using Natural Language Understanding (NLU). Inductive Relation Induction~\citep{yang2022language} investigates the prediction of relation that involves unseen entities. \citet{misra2022property} focus on classifying synthetic language sentences using neural networks, whereas \citet{yang2021learning} have studied rule induction using quasi-natural language (symbolic rather than natural language). 

Other taxonomies of reasoning tasks include: 
\begin{enumerate}[label=(\alph*)]
    \item \textbf{Formal Reasoning vs. Informal Reasoning}~\citep{evans2004informal,teig2016bringing}: This taxonomy is based on the nature or formality of the reasoning process. Formal reasoning involves following strict rules, logical frameworks, or formal systems to derive conclusions and often relies on mathematical or deductive reasoning. Informal reasoning, on the other hand, is less structured and more intuitive, relying on personal experiences, common sense, and heuristics.

    \item \textbf{Neural Reasoning vs. Symbolic Reasoning vs. Neural-Symbolic Reasoning}~\citep{garcez2008neural,garcez2015neural,garcez2022neural}: This taxonomy is based on the underlying computational framework used for reasoning. Neural reasoning refers to approaches that utilize neural networks or deep learning models for reasoning tasks. Symbolic reasoning involves using symbolic representations, logic-based inference rules, or symbolic manipulation for reasoning. Neural-symbolic reasoning combines elements of both neural networks and symbolic reasoning, aiming to integrate their respective strengths.

    \item \textbf{Backward Reasoning vs. Forward Reasoning}~\citep{al2015comparison}: This taxonomy is based on the direction of the reasoning process. Backward reasoning starts from a goal or desired outcome and works backward by applying rules or evidence to determine the necessary conditions or steps to reach that goal. Forward reasoning starts with initial premises or evidence and progresses step-by-step to derive new conclusions or reach a final outcome.

    \item \textbf{Single-step Reasoning vs. Multi-step Reasoning}~\citep{song2018explore,yu2023natural}: This taxonomy is based on the complexity or number of steps involved in the reasoning process. Multi-step reasoning refers to tasks that require multiple sequential or interconnected steps to arrive at a solution or conclusion. It involves chaining together intermediate steps or inferences to reach the final result.

    \item \textbf{Deductive Reasoning vs. Defeasible Reasoning}~\citep{yu2023natural,koons2005defeasible,pollock1987defeasible,pollock1991theory}: The classification criterion for this type of reasoning is based on the nature of the reasoning process and the handling of exceptions or conflicting information. Defeasible reasoning involves reasoning under uncertainty or with incomplete information, where conclusions can be overridden or defeated by new evidence or exceptions. It allows for the revision or re-evaluation of conclusions based on additional information or context.

    \item \textbf{Unimodal Reasoning vs. Multimodal Reasoning}~\citep{sowa2003laws,oberlander1996proof}: This taxonomy is based on the input modalities used in the reasoning process. Unimodal reasoning refers to reasoning tasks that involve a single modality of information or input, for example, reasoning tasks that are based solely on language information. Multimodal reasoning, on the other hand, involves integrating and reasoning with multiple modalities of information simultaneously. This could include combining visual, language, textual, auditory, or other types of input for the reasoning process. 
\end{enumerate}

In addition to the categorization mentioned above, there are several other ways to classify or categorize information and reasoning, including factual reasoning~\citep{byrne1999deductive}, counterfactual reasoning~\citep{bottou2013counterfactual}, plausible (defeasible) reasoning~\citep{collins1989logic}, default reasoning~\citep{Brewka2012}, and abstract reasoning~\citep{yu2021abstract}.

\subsubsection{Mathematical Representation}
By acknowledging the above diverse definitions and perspectives, we gain a richer understanding of reasoning as a multifaceted concept that spans philosophical inquiry, formal logic, and practical applications in fields such as NLP. In this section, we will explore the commonalities and distinct characteristics of reasoning across these domains and investigate the mathematical methodologies employed to advance our understanding and implementation of reasoning processes.
Here are examples of illustrating reasoning in different mathematical frameworks:

\paragraph{Propositional Logic}

Logical proposition: Let $p$ and $q$ be logical propositions. We can represent their conjunction (AND) as $p \wedge q$.
Modus Ponens: If $p \rightarrow q$ and $p$ are true, then we can conclude $q$. This can be represented as $(p \rightarrow q) \wedge p \rightarrow q$.

\paragraph{Predicate Logic}

Quantifier and Predicate: Let $P(x)$ be a predicate representing ``$x$ is a prime number.'' The existential quantifier ($\exists$) can be used to express the existence of a prime number, such as $\exists x P(x)$.
Universal Quantifier: Let $Q(x)$ be a predicate representing ``$x$ is an even number.'' The universal quantifier ($\forall$) can be used to express that all numbers are even, such as $\forall x Q(x)$.

\paragraph{Set Theory}

Set Intersection: Let $A$ and $B$ be sets. The intersection of $A$ and $B$ is denoted as $A \cap B$.
Set Complement: Let $A$ be a set. The complement of $A$ is denoted as $A'$.

\paragraph{Graph Theory}

Graph Representation: Let $G = (V, E)$ be a graph, where $V$ represents the set of nodes and $E$ represents the set of edges.
Shortest Path: Let $d(u,v)$ represent the shortest path between nodes $u$ and $v$ in a graph. The shortest path problem can be formulated as finding the minimum value of $d(u,v)$ for all pairs of nodes.

\paragraph{Conditional Probability} Let $P(A)$ represent the probability of event $A$ and $P(B)$ represent the probability of event $B$. The conditional probability of $A$ given $B$ is denoted as $P(A|B)$ and can be calculated using Bayes' theorem.

\paragraph{Formal Systems}

Axiomatic System: Let $S$ be an axiomatic system with a set of axioms and a set of inference rules. A formal proof within the system can be represented as a sequence of statements, where each statement is either an axiom or derived using the inference rules.

These mathematical expressions provide a glimpse into how reasoning can be expressed mathematically in different frameworks. However, it is important to note that the complexity of reasoning problems often requires more elaborate mathematical expressions and formalisms.

Despite these traditional categorizations and rigorous mathematical representations, with the advent of foundation models, researchers have increasingly moved away from strict adherence to these restrictions. Instead, they have embraced a more flexible approach to reasoning, considering its various forms and applications in different scenarios.

In contemporary research, reasoning has evolved to encompass a wide range of tasks and contexts. For instance, Commonsense Reasoning has emerged as a vital area for study, aiming to endow AI systems with the ability to understand and reason about everyday situations, incorporating common knowledge and contextual understanding. An example illustrating Commonsense Reasoning is shown in Table~\ref{tab:commonsense_reasoning_example}.
Similarly, Mathematical Reasoning has garnered significant attention, particularly in the context of foundation models. Researchers are exploring ways to enhance models' mathematical reasoning abilities, including solving math word problems. An example showcasing Mathematical Reasoning, specifically a Math Word Problem, is presented in Table~\ref{tab:math_reasoning_example}.

These examples highlight the diverse manifestations of reasoning in different application domains. The focus has shifted from rigid categorizations to addressing specific reasoning challenges and designing models capable of tackling them effectively. By embracing this more flexible and application-driven perspective, researchers aim to broaden the scope of reasoning and advance the development of AI systems capable of exhibiting human-like reasoning capabilities across a wide array of tasks and contexts.

\subsection{Foundation Models and Recent Progress}

\begin{figure}[tbp]
	\begin{center}
		
\includegraphics[width=1.0\columnwidth]{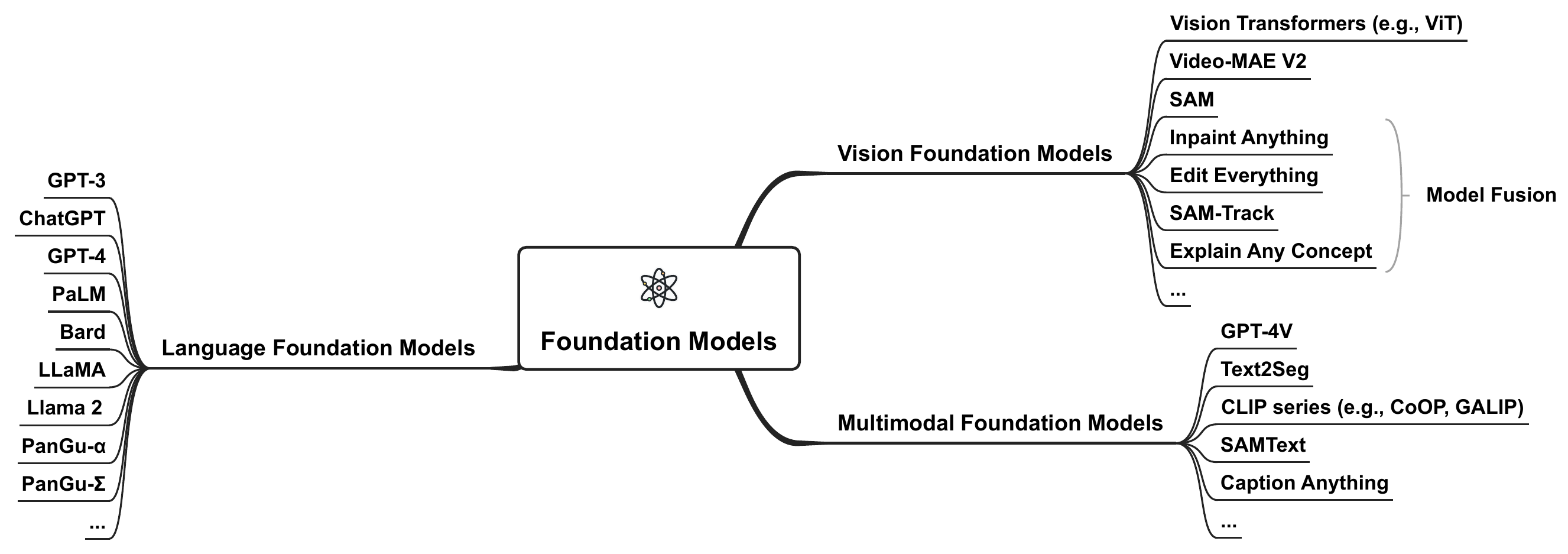}
	\end{center}
	\caption{Foundation models can be mainly categorized into language, vision, and multimodal foundation models, each of which is an actively researched area.}
\label{fig:bg_foundation_model}
\end{figure}

In recent years, the field of artificial intelligence has witnessed rapid development of foundation models. Foundation models have revolutionized various domains, including but not limited to computer vision, natural language processing, and speech recognition. Next, we introduce three main categories of the foundation model and their representative works, as summarized in Figure~\ref{fig:bg_foundation_model}.

\subsubsection{Language Foundation Models and Language Prompt}

Foundation models, such as GPT-3~\citep{brown2020language}, herald breakthroughs in natural language understanding and generation tasks first. These models have shown the ability to understand and generate coherent, contextually appropriate responses in natural language and have achieved significant progress in various language-related tasks, including text completion, translation, dialogue, summarization, question answering, and beyond. 

Recently, with the advancements in research and refined training methodologies, a variety of advanced large-scale language models~\citep{zhao2023survey}\footnote{\url{https://github.com/RUCAIBox/LLMSurvey}} have emerged. Prominent among them are GPT-4~\citep{openai2023gpt4}, which powers ChatGPT, and PaLM~\citep{chowdhery2022palm}, a crucial component of Bard. Additionally, LLaMA~\citep{touvron2023llama} and Llama 2~\citep{touvron2023llama2} have gained popularity as a collection of open-source large language models, varying in parameters from 7B to 65B.
The focus on multilingual support has also become a key area of interest in foundation modeling research. For instance, PanGu-$\alpha$~\citep{zeng2021pangu}, pre-trained on 1.1 TB of Chinese data and has 200 billion parameters, shows robust language modeling capabilities. Taking the concept further, PanGu-$\Sigma$~\citep{ren2023pangu} utilizes techniques like Random Routed Experts (RRE) and Expert Computation and Storage Separation (ECSS) to develop a system that trains a trillion-parameter language model, leading to a significant 6.3x increase in training throughput through heterogeneous computing.

\subsubsection{Vision Foundation Models and Visual Prompt}
Following the remarkable success of foundation models in the language domain, its implications transcend to the realms of the vision field as well. 

Vision Transformer (ViT)~\citep{dosovitskiy2020image} applies the Transformer framework to computer vision tasks, achieving impressive performance in classification and retrieval tasks by leveraging self-attention mechanisms.
Swin Transformer~\citep{liu2021swin} introduces a hierarchical structure with shifted windows, improving the efficiency of processing high-resolution images. It has demonstrated strong performance across various computer vision tasks such as image classification, object detection, and semantic segmentation.
Methods like MAE~\citep{he2022masked}, BEIT~\citep{bao2021beit}, and CAE~\citep{chen2023context} propose masked modeling as an efficient self-supervised learning strategy to learn general-purpose visual representations.
VideoMAE~V2~\citep{wang2023videomae} is an enhanced version of VideoMAE~\citep{tong2022videomae}, with a billion parameters, designed for video understanding tasks. It utilizes self-supervised learning to learn temporal and spatial dependencies, excelling at tasks like action classification and action detection.
As multitask vision foundation models, Florence~\citep{yuan2021florence} and Florence-2~\citep{ding2022davit,xiao2023florence} can be easily adapted for a variety of computer vision tasks, such as classification, retrieval, object detection, visual question answering (VQA), image captioning, video retrieval, and action recognition, etc. 
Segment Anything Model (SAM)~\citep{kirillov2023segany} excels at producing object masks from input prompts like partial masks, points, or boxes. It has the capability to generate masks for all objects in an image. 
SAM is trained on a vast dataset that includes 11 million images and 1.1 billion masks. Notably, SAM demonstrates zero-shot performance across a wide range of segmentation tasks. As a zero-shot anomaly segmentation, Segment Any Anomaly+ (SAA+)~\citep{cao2023segment} introduces hybrid prompt regularization, leveraging domain-specific expertise and contextual information from the target image to enhance the adaptability of foundational models. By incorporating these elements into the regularization prompt, SAA+ strengthens the prompt's robustness, enabling more precise identification of anomalous regions. Furthermore, \citet{wang2023scaling} have also revealed the potential of incorporating domain expert knowledge as prior support in addressing segmentation challenges in complex scenes.

\paragraph{Model Fusion: Enhancing Visual Task through Combination}

There is a recent trend in the field of computer vision to combine different pre-trained Vision Foundation Models, each specializing in specific tasks, in order to tackle complex visual tasks more effectively. These approaches take advantage of the increasing power and diversity of these foundation models, leveraging their individual strengths to achieve superior performance in challenging visual tasks.

Inpaint Anything~\citep{yu2023inpaint} presents three essential functionalities in image inpainting, namely Remove Anything, Fill Anything, and Replace Anything, which are achieved through the synergistic combination of various foundational models.
It leverages click prompts for automatic segmentation, utilizes state-of-the-art inpainting models like LaMa~\citep{suvorov2021resolution} and Stable Diffusion~\citep{rombach2022high} for filling masked regions, and employs AI models with text prompts to generate specific content for filling or replacing voids.

Edit Everything~\citep{xie2023edit} presents a generative system that combines SAM~\citep{kirillov2023segany}, CLIP~\citep{radford2021learning}, and Stable Diffusion~\citep{rombach2022high} to enable image editing guided by both image and text inputs. 
Initially, Edit Everything~\citep{xie2023edit} employs SAM to segment the original image into several fragments. Subsequently, the process of image editing is guided by text prompts, leading to a transformation that adjusts the source image to correspond with the target image as described in the given text prompts.

SAM-Track~\citep{cheng2023segment} introduces a video segmentation framework that integrates Grounding-DINO~\citep{liu2023grounding}, DeAOT~\citep{yang2022decoupling}, and SAM~\citep{kirillov2023segany} to facilitate interactive and automated object tracking and segmentation across multiple modalities. The framework allows interactive prompts, including click-prompt, box-prompt, and text-prompt, in the initial frame of the video to guide SAM's segmentation process. Explain Any Concept (EAC)~\citep{sun2023explain} presents an approach for concept explanation, utilizing SAM for initial segmentation and introducing a surrogate model to enhance the efficiency of the explanation process.

\subsubsection{Multimodal Foundation Models}
As foundation models continue to exhibit impressive performance on individual modalities, such as language and images, a natural extension arises: Can these models effectively handle multimodal data? This question arises from the recognition that real-world scenarios often involve multiple modalities, such as text, images, and audio, which collectively provide a more comprehensive and nuanced understanding of the data.

Text2Seg~\citep{zhang2023text2seg} introduces a vision-language model that leverages text prompts as input to generate segmentation masks. The model operates by using a text prompt to generate bounding boxes with Grounding DINO~\citep{liu2023grounding}, which guides SAM in generating segmentation masks.
CLIP~\citep{radford2021learning} learns joint representations of images and text. It achieves this by aligning visual and textual information, enabling cross-modal understanding, and demonstrating impressive capabilities in various vision and language tasks.
Similarly, methods~\citep{uniter,oscar,vinvl,Lit,yao2021filip,jia2021scaling,huo2021wenlan,fei2022towards}, like ALIGN~\citep{jia2021scaling} and WenLan~\citep{huo2021wenlan}, align image and text representations by learning a common feature space.
CoOp (Context Optimization)~\citep{zhou2022learning} presents a straightforward technique to customize CLIP-like vision-language models for downstream tasks. CoOp employs learnable vectors to represent the context words in a prompt while maintaining the pre-trained parameters in a fixed state.
GALIP (Generative Adversarial CLIPs)~\citep{tao2023galip} is another advancement, specifically developed for the task of text-to-image generation.
In CLIP Surgery~\citep{li2023clip}, heatmaps are first generated based on text prompts. Point prompts, which are then sampled from these heatmaps, are then inputted into SAM~\citep{kirillov2023segany} for further processing.
Following this, a similarity algorithm utilizing CLIP~\citep{radford2021learning} is employed to produce the final segmentation map.
SAMText~\citep{he2023scalable} presents a flexible approach for creating segmentation masks tailored to scene text. This method initiates by deriving bounding box coordinates from the annotations present in an existing scene text detection model.
These coordinates then prompt SAM to generate masks. Caption Anything~\citep{wang2023caption} presents a foundational model-enhanced framework for image captioning that enables interactive multimodal control from both visual and linguistic aspects. 
By combining SAM~\citep{kirillov2023segany} with ChatGPT, users gain the flexibility to manipulate images using a variety of prompts, including points prompts or bounding boxes prompts, during interaction. 
It additionally leverages Large Language Models (LLMs) to refine instructions, ensuring they accurately reflect the user's intended meaning and remain consistent with their intention.
GPT-4V(ision) empowers users to interpret and analyze user-provided image inputs~\citep{gpt4v_system_card}.

The potential for foundation models to excel in multimodal tasks (text-to-image, text-to-code, and speech-to-text) opens up exciting possibilities in various domains. By seamlessly integrating and processing information from different modalities, these models can enhance tasks such as image captioning, visual question answering, and audio-visual scene understanding. Moreover, multimodal foundation models hold promise in applications that require reasoning and decision-making based on multiple sources of information. By harnessing the power of multimodal data, these models have the potential to unlock new levels of understanding, context awareness, and performance across a wide range of domains, including robotics~\citep{roya2023_robotics_survey}, healthcare~\citep{qiu2023large}, autonomous vehicles~\citep{zhou2023_drivellm_survey}, and multimedia analysis.

\subsubsection{Potential for Applications in Reasoning}

\begin{figure}[tbp]
	\begin{center}
		
\includegraphics[width=1.0\columnwidth]{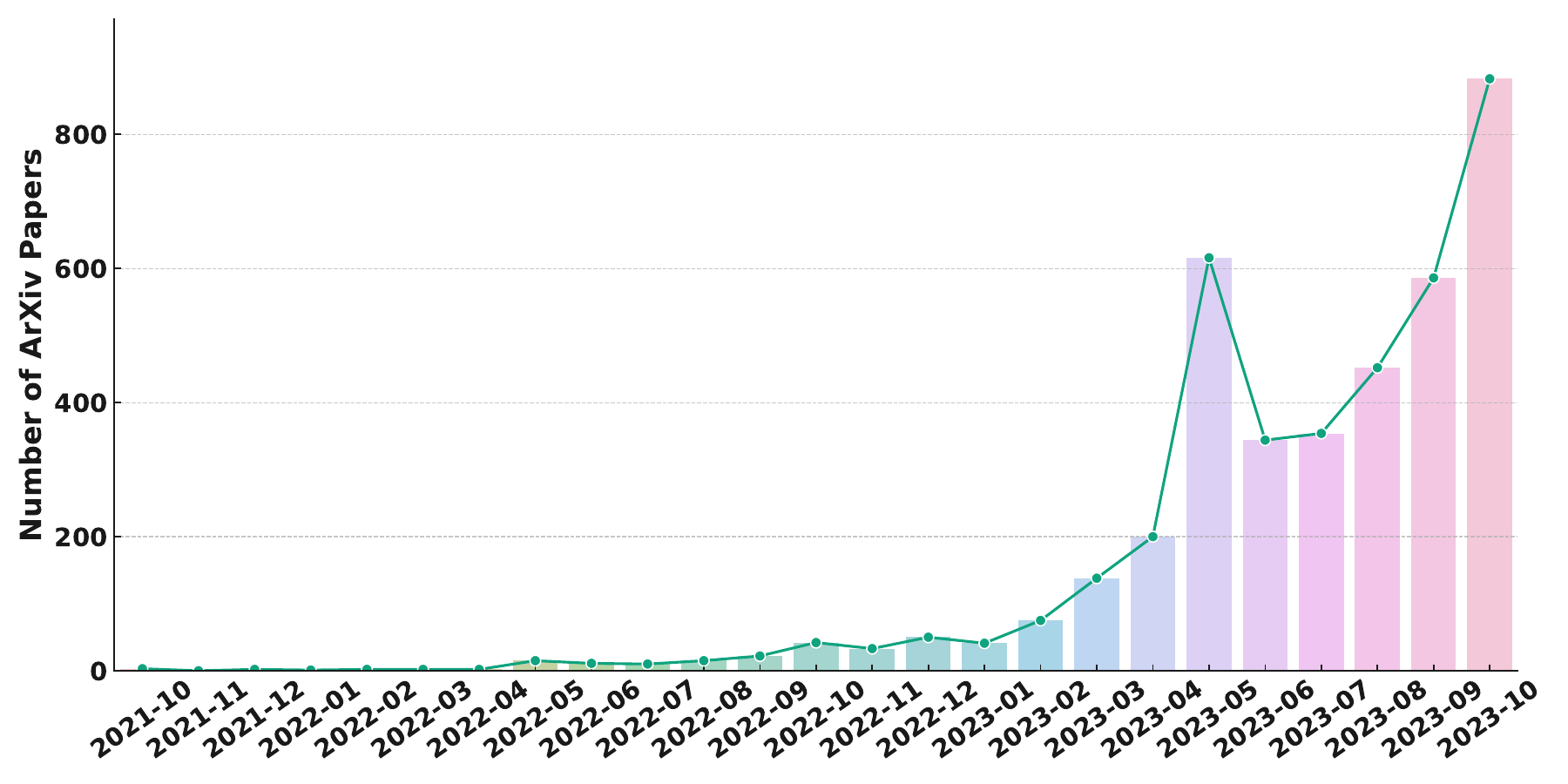}
	\end{center}
	\caption{Number of arXiv Papers on ``Reasoning with Large Language Models'' over the past two years. It depicts a rising trend in the research interest, with the number of articles surging notably in the months of 2023.}
\label{fig:arxiv_paper}
\end{figure}

Reasoning with foundation models is an emerging field.
Recently there has been an influx of research that attempts to apply foundation models to reasoning tasks, and promising results have been achieved. The statistics are presented in Figure~\ref{fig:arxiv_paper}.
\citet{laban2023llms} identify challenges in evaluating complex tasks with Large Language Models (LLMs) and highlight the need for improved evaluation benchmarks. \citet{shi2023language} demonstrate that multilingual language models can go beyond language and perform tasks like commonsense reasoning and semantic judgment in a word-in-context setting. 
 Language models serve as multilingual reasoners employing chain-of-thought processes. Self-Taught Reasoner (STaR)~\citep{zelikman2022star} enhances a model's reasoning abilities by iteratively generating rationales and fine-tuning based on correct answers. 
MWP-BERT~\citep{liang2022mwp} leverages both BERT~\citep{kenton2019bert} (110M) and RoBERTa~\citep{liu2019roberta} (123M) pre-training to tackle Math Word Problem (MWP) solving. Meanwhile, Minerva~\citep{lewkowycz2022solving}, based on the PaLM~\citep{chowdhery2022palm} pre-trained language model, boasts an impressive parameter size of up to 540B. Minerva demonstrates strong performance by accurately answering nearly a third of over two hundred undergraduate-level problems in various disciplines like chemistry, biology, economics, physics, and other sciences that involve quantitative reasoning.
Zero-shot-CoT~\citep{kojima2022large} demonstrates impressive performance across a range of reasoning tasks, including arithmetic challenges such as MultiArith~\citep{patel-etal-2021-nlp}, GSM8K~\citep{cobbe2021gsm8k}, AQUA-RAT~\citep{ling-etal-2017-program}, SVAMP~\citep{patel-etal-2021-nlp}, symbolic reasoning, and other logical reasoning tasks like Date Understanding~\citep{srivastava2023beyond}, Tracking Shuffled Objects~\citep{srivastava2023beyond}, all without the necessity for handcrafted few-shot examples.
Employing just one prompt template, this approach indicates the zero-shot potential and the high-level, multi-task cognitive capacities of LLMs, while also emphasizing the significant prospects for additional research in this field.

However, there is still a need for intelligent systems that can perform more sophisticated forms of reasoning, beyond simple pattern recognition.

\section{Reasoning Tasks}
\label{sec:tasks}
In this section, we provide a concise overview of various reasoning tasks, as Figure~\ref{fig:intro_overview} shows. Here, we present distinct categories of reasoning approaches and tasks:
\begin{itemize}
    \item Commonsense Reasoning (Section~\ref{sec:commensense_reasoning}): Exploring the capacity to infer and apply everyday, intuitive knowledge.
    \item Mathematical Reasoning (Section~\ref{sec:mathematical_reasoning}): Focusing on the ability to solve mathematical problems and derive logical conclusions.
    \item Logical Reasoning (Section~\ref{sec:logical_reasoning}): Examining the process of drawing inferences and making decisions based on formal logic.
    \item Causal Reasoning (Section~\ref{sec:causal_reasoning}): Investigating the understanding of cause-and-effect relationships and their implications.
    \item Multimodal Reasoning (Section~\ref{sec:multimodal_reasoning}): Involving reasoning across multiple data modalities, such as text, images, and sensory information.
    \item Visual Reasoning (Section~\ref{sec:visual_reasoning}): Focusing on tasks that require the interpretation and manipulation of visual data.
    \item Embodied Reasoning (Section~\ref{sec:embodied_reasoning}): Exploring reasoning in the context of embodied agents interacting with their environment.
    \item Other Reasoning Tasks (Section~\ref{sec:other_reasoning}): The discussion of reasoning extends across various contexts, including conceptual frameworks, such as abstract reasoning~\ref{subsubsec:abstract_reasoning}, defeasible reasoning~\ref{sec:defeasible_reasoning}, as well as applied fields such as medical reasoning~\ref{subsubsec:medical_reasoning}, bioinformatic reasoning~\ref{subsubsec:bio_reasoning}, among others. We also highlight the immense utility of long-chain reasoning in applications for researchers to explore~\ref{subsubsec:long-chain-reasoning}.
\end{itemize}

This comprehensive overview provides insights into the diverse landscape of reasoning tasks and approaches within the field.
A summary of seminal works in each reasoning sector can be found in Figure~\ref{tab:taxonomy_method_small}.
\tikzstyle{my-box}=[
    rectangle,
    % draw=hidden-draw,
    draw=Purples-J,
    rounded corners,
    text opacity=1,
    minimum height=1.5em,
    minimum width=5em,
    inner sep=2pt,
    align=center,
    fill opacity=.5,
    font=\bf\scriptsize,
]
% \tikzstyle{leaf}=[my-box, minimum height=1.5em,
%     fill=hidden-orange!60, text=black, align=left,font=\scriptsize,
%     inner xsep=2pt,
%     inner ysep=4pt,
% ]
\tikzstyle{leaf}=[my-box, minimum height=1.5em,
    fill=Purples-E, text=black, align=left,font=\bf\scriptsize,
    inner xsep=2pt,
    inner ysep=4pt,
]
\begin{figure*}[tp]
    \centering
    \resizebox{\textwidth}{!}{
        \begin{forest}
            forked edges,
            for tree={
                grow=east,
                reversed=true,
                anchor=base west,
                parent anchor=east,
                child anchor=west,
                base=left,
                font=\bf\small,
                rectangle,
                % draw=hidden-draw,
                draw=Purples-J,
                rounded corners,
                align=left,
                minimum width=4em,
                edge+={darkgray, line width=1pt},
                s sep=3pt,
                inner xsep=2pt,
                inner ysep=3pt,
                ver/.style={rotate=90, child anchor=north, parent anchor=south, anchor=center},
            },
            where level=1{text width=7em,font=\bf\scriptsize,}{},
            where level=2{text width=7em,font=\bf\scriptsize,}{},
            where level=3{text width=5.5em,font=\bf\scriptsize,}{},
            where level=4{text width=6.1em,font=\bf\scriptsize,}{},
            [
                Reasoning Tasks with Foundation Models, ver
                    [
                        Commonsense
                        [
                            QA
                            [
                                CQA~\citep{talmor-etal-2019-commonsenseqa}{,}
                                ConceptNet~\citep{speer2017conceptnet}{,}\\
                                CoS-E~\citep{rajani2019explain}{,} 
                                CAGE~\citep{rajani2019explain}{,}
                                etc.
                                , leaf, text width=39.7em
                            ]
                        ]
                        [
                            Physical
                            [
                            ESPRIT~\citep{rajani-etal-2020-esprit}{,}
                            PACS~\citep{yu2022pacs}{,}
                            PIQA~\citep{bisk2020piqa}{,} \\
                            NEWTON~\citep{wang2023newton}{,}
                            etc.
                            , leaf, text width=39.7em
                            ]
                        ]
                        [
                            Spatial
                            [Liu et al~\citep{liu-etal-2022-things}{,} etc.
                            , leaf, text width=39.7em
                            ]
                        ]
                    ]
                    [
                        Mathematical
                        [
                            Arithmetic
                            [
                                PromptPG~\citep{lu2022dynamic}{,}
                                etc.
                                , leaf, text width=39.7em
                            ]
                        ]
                        [
                            Geometry
                            [
                            Geoformer~\citep{chen-etal-2022-unigeo}{,}
                            Inter-GPS~\citep{lu2021inter}{,}
                            etc.
                            , leaf, text width=39.7em
                            ]
                        ]
                        % [
                        %     Math QA
                        %     [
                        %     Ling et al.~\citep{ling2017program}{,}
                        %     MT2Net~\citep{zhao-etal-2022-multihiertt}{,}
                        %     Talmor et al.~\citep{talmor2021multimodalqa}{,}
                        %     , leaf, text width=39.7em
                        %     ]
                        % ]
                        [
                            Theorem
                            [
                            LeanDojo~\citep{yang2023leandojo}{,}
                            etc.
                            , leaf, text width=39.7em
                            ]
                        ]
                        [
                            Scientific
                            [
                            SciBench~\citep{wang2023scibench}{,}
                            ScienceWorld~\citep{wang2022scienceworld}{,}\\
                            ScienceQA~\citep{lu2022learn}{,}
                            etc.
                            , leaf, text width=39.7em
                            ]
                        ]
                    ]
                    [
                        Logical
                        [
                            Propositional
                            [
                                Tomasic et al.~\citep{tomasic2021propositional}{,}
                                etc.
                                , leaf, text width=39.7em
                            ]
                        ]
                        [
                            Predicate
                            [
                            ILP~\citep{cropper2022inductive}{,}
                            etc.
                            , leaf, text width=39.7em
                            ]
                        ]
                    ]
                    [
                        Causal
                        [
                            Counterfactual
                            [
                                Li et al.~\citep{li2023counterfactual}{,}
                                Wu et al.\citep{wu2023reasoning}{,}
                                etc.
                                , leaf, text width=39.7em
                            ]
                        ]
                    ]
                    [
                        Visual
                        [
                            3D
                            [
                                3D-LLM~\citep{3dllm}{,}
                                3D-VisTa~\citep{3dvista}{,} 
                                etc.
                                , leaf, text width=39.7em
                            ]
                        ]
                    ]
                    [
                        Audio
                        [
                            Speech
                            [
                                SUPERB~\citep{yang21superb}{,}
                                SUPERB-SG~\citep{tsai2022superb}{,} \\
                Wav2Vec~\citep{baevski2020wav2vec}{,} Speech SIMCLR~\citep{jiang2020speech}{,}\\
                Unit BERT (HuBERT)~\citep{hsu2021hubert}{,} WavLM~\citep{chen2022wavlm}
                                etc.
                                , leaf, text width=39.7em
                            ]
                        ]
                    ]
                    [
                        Multimodal
                        [
                            Generation
                            [
                                Stable Diffusion~\citep{rombach2022high}{,}
                                DALL·E{,} 
                                Midjourney{,}\\
                                Flamingo-80B~\citep{alayrac2022flamingo}{,}
                                % Frozen~\citep{tsimpoukelli2021multimodal}{,}
                                MAGMA~\citep{eichenberg-etal-2022-magma}{,}\\
                                Kosmos-2~\citep{peng2023kosmos2}{,}
                                etc.
                                , leaf, text width=39.7em
                            ]
                        ]
                        [
                            Alignment
                            [
                            CLIP~\citep{radford2021learning}{,}
                            BLIP-2~\citep{li2023blip2}{,}
                            etc.
                            , leaf, text width=39.7em
                            ]
                        ]
                        [
                            Understanding
                            [
                            LLaVA~\citep{liu2023visual}{,}
                            DePlot~\citep{liu-2022-deplot}{,}
                            MatCha~\citep{liu-2022-matcha}{,} \\
                            DetGPT~\citep{pi2023detgpt}{,}
                            etc.
                            , leaf, text width=39.7em
                            ]
                        ]
                    ]
                    [
                        Embodied
                        [
                            Introspective
                            [
                            PAL~\citep{gao2023pal}{,}
                            ProgPrompt~\citep{singh2022progprompt}{,} \\
                            Code-as-Policies~\citep{codeaspolicies2022}{,} 
                            SayCan~\citep{Ahn2022DoAI}{,}
                            etc.
                            , leaf, text width=39.7em
                            ]
                        ]
                        [
                            Extrospective
                            [
                            Self-Ask~\citep{press2023measuring}{,} ReAct~\citep{yao2023react}{,}\\
                            ToolFormer~\citep{schick2023toolformer}{,}
                            LLM-Planner~\citep{song2023llmplanner}{,}\\
                            Statler~\citep{yoneda2023statler}{,}
                            EmbodiedGPT~\citep{mu2023embodiedgpt}{,}
                            % Dasgupta et al.~\citep{dasgupta2022collaborating}{,}
                            etc.
                            , leaf, text width=39.7em
                            ]
                        ]
                        [
                            Multi-agent
                            [
                            Zhang et al.~\citep{zhang2023building}{,}
                            Du et al.~\citep{du2023improving}{,} \\
                            Nascimento et al.~\citep{nascimento2023selfadaptive}{,}
                            Chen et al.~\citep{chen2023federated}{,}
                            etc.
                            , leaf, text width=39.7em
                            ]
                        ]
                    ]
                    [
                        Others
                        [
                            ToM
                            [
                            Kosinski et al.~\citep{kosinski2023theory}{,}
                            etc.
                            , leaf, text width=39.7em
                            ]
                        ]
                        [
                            Weather \\Prediction
                            [
                            MetNet-2~\citep{espeholt2022deep}{,}
                            Bi et al.~\citep{bi2023accurate}{,}
                            etc.
                            , leaf, text width=39.7em
                            ]
                        ]
                        [
                            Abstract \\Reasoning
                            [
                            Gendron et al.~\citep{gendron2023large}{,}
                            etc.
                            , leaf, text width=39.7em
                            ]
                        ]
                        [
                            Defeasible \\Reasoning
                            [
                            BoardgameQA~\citep{kazemi2023boardgameqa}{,}
                            etc.
                            , leaf, text width=39.7em
                            ]
                        ]
                        [
                            Medical \\Reasoning
                            [
                            Med PaLM 2~\citep{singhal2023towards}{,}
                            Med PaLM M~\citep{tu2023towards}{,}\\
                            VisionFM~\citep{qiu2023visionfm}{,} 
                            RETFound~\citep{zhou2023foundation}{,}
                            etc.
                            , leaf, text width=39.7em
                            ]
                        ]
                        [
                            Bioinformatics \\Reasoning
                            [
                            ProGen~\citep{madani2023large}{,}
                            RFdifusion~\citep{watson2023novo}{,}
                            etc.
                            , leaf, text width=39.7em
                            ]
                        ]
                    ]
                ]
            ]
        \end{forest}
    }
    \caption{Taxonomy of Reasoning Tasks with Foundation Models. Only the representative approaches for each type of task are listed.
    }
    \label{tab:taxonomy_method_small}
\end{figure*}
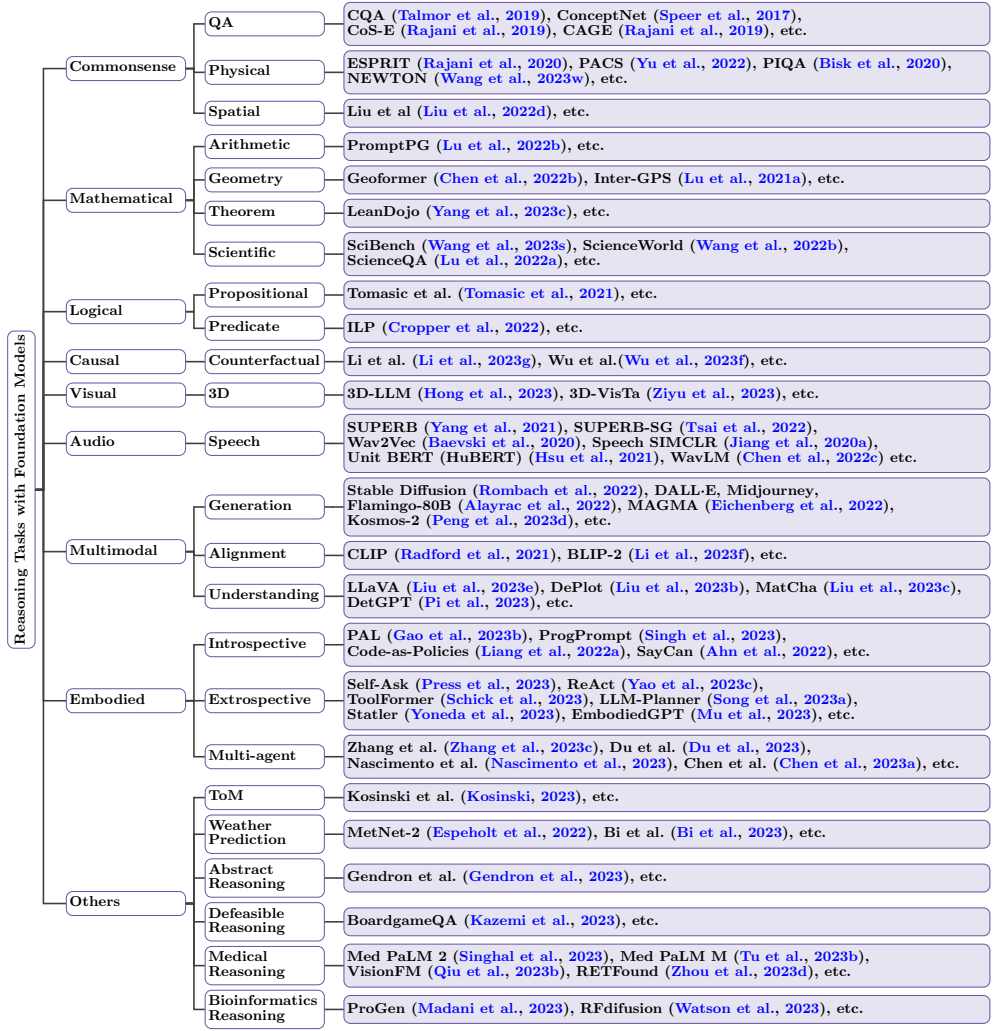

\subsection{Commonsense Reasoning}~\label{sec:commensense_reasoning}

Commonsense reasoning refers to the human-like capacity to make assumptions and inferences about the nature and characteristics of everyday situations that humans encounter on a regular basis\footnote{\url{http://www-formal.stanford.edu/leora/commonsense/}}.

Recent research indicates that language models are capable of acquiring certain aspects of common sense knowledge~\citep{zhao2023large,ye2023improving}. In the domain of structured commonsense reasoning, \citet{madaan-etal-2022-language} tackle the task by generating a graph based on natural language input. They formalize this problem as a code generation challenge, utilizing large language models that are prompted with code to construct the graph representation.
\citet{berglund2023reversal} also point out that language models often demonstrate a fundamental lapse in logical deduction, failing to generalize a common pattern in their training set, specifically, the likelihood of ``B is A" occurring if ``A is B" is present.
\citet{li2022systematic} take a systematic approach to evaluate the performance of large pre-trained language models on various commonsense benchmarks. They conduct zero-shot and few-shot commonsense evaluations across four different benchmarks, considering six different model sizes. Notably, their evaluation includes a remarkably large language model with 280 billion parameters. Multiple evaluation settings, such as different score functions and prompt formats, are explored to comprehensively assess the models' ability to capture and reason about commonsense knowledge.

Another direction in the field of commonsense reasoning involves combining pre-trained language models with commonsense-specific fine-tuning techniques. \citet{chang2021plain} propose several architectural variations, leverage external commonsense corpora, and employ commonsense-specific fine-tuning techniques for the Social IQA task~\citep{sap-etal-2019-social}. Through their work, they demonstrate that these optimizations can enhance the model's performance in tasks related to social intelligence.
Furthermore, \citet{yang2023bridging} introduce a two-stage framework designed to connect pre-training and fine-tuning in the task of commonsense generation. 

\begin{figure}[tbp]
	\begin{center}
\includegraphics[width=1.0\columnwidth]{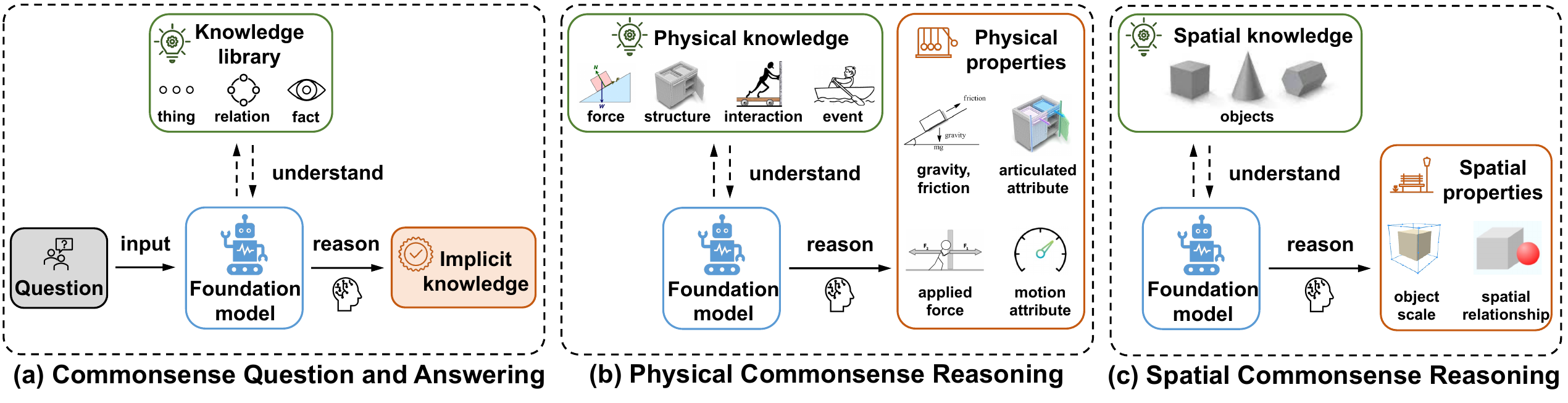}
	\end{center}
	\caption{Three areas of research of foundation models in commonsense reasoning. (a)~By understanding everyday knowledge, foundation models can reason about implicit knowledge from questions and deduce answers. (b) Foundation models infer a wide range of physical properties from general physical knowledge. (c) Foundation models reason about spatial properties from a set of objects.}
\label{fig:reason_commonsense}
\end{figure}

In addition to the above-mentioned works, there are other aspects of commonsense reasoning that have been explored. These include commonsense question answering~(QA), physical reasoning, spatial reasoning, and the corresponding benchmarks, as shown in Figure~\ref{fig:reason_commonsense}. These areas of research contribute to a deeper understanding of how language models can effectively capture and reason about commonsense knowledge in various contexts.

\subsubsection{Commonsense Question and Answering (QA)}
As a subfield of commonsense reasoning, Commonsense Question Answering (QA) focuses on developing systems capable of answering questions that require a deep understanding of everyday knowledge and human-like reasoning. Unlike traditional fact-based QA, where answers can be derived from explicit information, commonsense QA involves understanding and reasoning about implicit knowledge and everyday human reasoning, as depicted in Figure~\ref{fig:reason_commonsense}(a).

The Commonsense Question Answering (CQA) dataset~\citep{talmor-etal-2019-commonsenseqa} is a challenging multiple-choice dataset specifically designed for commonsense question answering. It is derived from ConceptNet~\citep{speer2017conceptnet} and consists of approximately 12,000 questions. Every question comes with one correct answer and four additional distractor answers.
In addition, the Commonsense Explanations (CoS-E) dataset~\citep{rajani2019explain} contains human commonsense explanations for the CQA dataset. The CoS-E dataset comprises two types of explanations: Selected explanations, which are text spans highlighted in the question that justify the answer choice, and open-ended explanations, which are free-form natural language explanations. 

Commonsense Auto-Generated Explanation (CAGE) model~\citep{rajani2019explain} is a framework that involves training a language model to generate useful explanations by fine-tuning it using both the problem input and human-generated explanations.

The development of effective commonsense QA systems is an active area of research, and ongoing advancements in language models, knowledge representation, and reasoning techniques continue to push the boundaries of commonsense understanding in machine intelligence.

\subsubsection{Physical Commonsense Reasoning}
Commonsense physical reasoning~\citep{ding2021dynamic}, shown in Figure~\ref{fig:reason_commonsense}(b), involves utilizing everyday knowledge about the physical world to reason and understand the behavior of objects and their properties. It encompasses reasoning about physical concepts, such as the properties of objects reasoning (gravity, mass, inertia, or friction), their affordances, and how they can be manipulated~\citep{Chu_2023_CVPR}.

\underline{E}xplaining \underline{S}olutions to \underline{P}hysical 
\underline{R}eason\underline{I}ng \underline{T}asks (ESPRIT) framework~\citep{rajani-etal-2020-esprit} combines commonsense physical reasoning with interpretability via natural language explanations. It operates in two stages: firstly, pinpointing key physical events in tasks, and secondly, crafting natural language descriptions for both the initial scene and these crucial events.
The framework aims to provide a unified approach to reasoning about commonsense physical concepts, such as gravity, friction, and collision, while also offering qualitative explanations using natural language.
PACS (Physical Audiovisual CommonSense)~\citep{yu2022pacs} is a dataset designed for physical audiovisual commonsense reasoning. It comprises 13,400 question-answer pairs, including 1,377 distinct questions and 1,526 videos for physical commonsense. By benchmarking unimodal and multimodal reasoning models, PACS identifies the limitations and areas of improvement in current models, thereby providing valuable opportunities to propel research in physical reasoning by examining multimodal reasoning approaches.
PIQA (Physical Interaction: Question Answering)~\citep{bisk2020piqa} is a dataset that focuses on multiple-choice question-answering in the domain of physical interactions. The task involves selecting the most appropriate solution from two given options based on a given question. The PIQA dataset consists of over 16,000 training QA pairs, with additional data reserved for development and testing. The questions in PIQA have an average length of 7.8 words, while both correct and incorrect solutions have an average length of 21.3 words. NEWTON~\citep{wang2023newton} is a comprehensive platform that serves as a repository, pipeline, and benchmark specifically created to assess the physical reasoning capabilities of LLMs.

CATER~\citep{girdhar2019cater} mainly focuses on physics-related visual scenes. CLEVRER~\citep{yi2019clevrer} is a video question-answering benchmark that targets the physical and causal relations grounded in dynamic videos of rigid-body collisions. CLEVRER-Humans~\citep{mao2022clevrer} further extends it to the causal judgment of physical events with human labels.
Physion~\citep{bear2021physion}, Physion++~\citep{tung2023physion++}, and ComPhy~\citep{chen2022comphy} evaluate objects with different latent physical properties (e.g., mass, friction, elasticity, and deformability) from dynamic videos rendered from physics engines.

Based on the above benchmarks, transformer-based foundational models~\citep{deepmind2020object,wu2022slotformer} and neuro-symbolic frameworks with differentiable physics~\citep{ding2021dynamic} are developed.
Aloe (Attention over Learned Object Embeddings)~\citep{deepmind2020object} integrates MONet~\citep{burgess2019monet} for unsupervised object segmentation with self-attention mechanisms, facilitating spatio-temporal physical reasoning about objects.
SlotFormer~\citep{wu2022slotformer}, a Transformer-based object-centric dynamics model, is designed to unsupervisedly decipher complex systems and interactions from videos. 
Utilizing a context encoding provided by Spatial Transformer~\citep{jaderberg2016spatial}, Generative Structured World Models (G-SWM)~\citep{lin2020improving} advance object-centric world modeling. They incorporate multimodal uncertainty and situational awareness through a core module known as Versatile Propagation (V-Prop).
These frameworks and datasets contribute to the advancement of commonsense physical reasoning by providing resources for model evaluation, interpretability, and understanding physical concepts through explanations and multimodal analysis.

Currently, the physical commonsense reasoning domain based on foundation models is relatively unexplored, offering a ripe avenue for research and development. This presents a unique chance for researchers and practitioners to delve into and expand the boundaries of what's possible with these models, potentially leading to groundbreaking advancements and innovations.

\subsubsection{Spatial Commonsense Reasoning}

As illustrated in Figure~\ref{fig:reason_commonsense}(c), spatial commonsense reasoning involves detecting the spatial position of objects and inferring the relationships between visual stimuli to understand the surrounding environment. Within the domain of spatial commonsense reasoning, two significant perspectives are object scales~\citep{arocaouellette2021prost} and spatial relationship~\citep{hudson2019gqa}.
\citet{liu-etal-2022-things} introduce a spatial commonsense benchmark, distinctly highlighting the relative sizes of objects and the spatial interactions between individuals and objects across various actions.
They investigate the performance of various models, including pre-trained vision-language models and image synthesis models. Interestingly, they find that the models for synthesizing images demonstrate better capabilities in learning accurate coherent knowledge of spatial relationships compared to other models. Furthermore, the spatial insights obtained through these models for synthesizing images also demonstrate their utility in enhancing natural language understanding tasks that necessitate spatial commonsense reasoning.

\subsection{Mathematical Reasoning}~\label{sec:mathematical_reasoning}

Mathematics distinguishes itself as a distinct language that relies on symbolic forms, and precision in meaning and possesses lower dimensionality compared to natural language. This unique characteristic allows us to demonstrate that meaning can be derived from a set of learned rule sets, as exemplified by the symbolic representations of mathematical concepts~\citep{floyd2004wittgenstein}.
Mathematical problems can be effectively programmed when they are represented using symbols and corresponding expressions. By formulating these problems in a computer language that can be translated into machine code, deep learning-based reasoning systems have the ability to train on and acquire the underlying rules~\citep{hinton1990connectionist,schmidhuber2015deep,friedman2023tokenization}.

Experimental findings suggest that the performance of  Large Language Models (LLMs) shows a weak correlation with question difficulty. 
\citet{ling-etal-2017-program} propose an approach to solve algebraic word problems in a way that not only generates the answer but also provides an explanation or rationale for the obtained result.
MT2Net~\citep{zhao-etal-2022-multihiertt} is a specialized model designed to tackle the MultiHiertt dataset~\citep{zhao-etal-2022-multihiertt}. It retrieves supporting facts from financial reports and generates executable reasoning programs to answer questions. This approach aims to provide a comprehensive and accurate solution for the given questions.

\subsubsection{Arithmetic Reasoning}

Math Word Problems (MWPs) are commonly used to evaluate the arithmetic reasoning abilities of language models. While these issues may appear uncomplicated to humans, language models frequently encounter challenges when it comes to tasks involving arithmetic reasoning~\citep{hendrycksmath2021,patel-etal-2021-nlp}.

Previous research has explored various approaches to address these challenges. Template-based statistical learning methods like KAZB~\citep{kushman-etal-2014-learning}, ZDC~\citep{Zhou2015LearnTS}, and similarity-based method SIM~\citep{huang2016well} have been utilized. \citet{wang-etal-2017-deep} employs a recurrent neural network (RNN) to convert math word problems into equation templates, eliminating the need for complex feature engineering. Additionally, they developed a hybrid model that integrates the RNN with a similarity-based retrieval system, further enhancing its performance. \citet{xie2019goal} introduces an innovative neural approach to construct expression trees in a goal-oriented manner for solving math word problems. \citet{shen-etal-2021-generate-rank} introduces a novel ranking task for math word problems and presents the Generate \& Rank framework, which combines a generative pre-trained language model with multi-task learning. This approach allows the model to learn from its errors and effectively differentiate between correct and incorrect expressions.
A notable finding is that employing chain-of-thought prompting, along with a language model containing an impressive 540 billion parameters, yields performance comparable to task-specific fine-tuned models across multiple tasks~\citep{wei2022chain}. Unlike traditional symbolic reasoning tasks such as program synthesis and knowledge graph reasoning, solving MWPs requires additional emphasis on numerical reasoning.
PromptPG~\citep{lu2022dynamic} takes a different approach by utilizing policy gradient techniques to learn the selection of in-context examples. By dynamically constructing appropriate prompts for each test example, PromptPG facilitates the solving of math word problems. This adaptive approach enhances the model's ability to handle numerical reasoning tasks effectively.  \citet{wang2023math} introduce MATH-SHEPHERD, a novel process-oriented math verifier that evaluates and assigns a reward score to each step in Large Language Models' (LLMs) solutions to math problems.

\subsubsection{Geometry Reasoning}

GeoS~\citep{seo-etal-2015-solving} provides a system for mapping geometry word problems into a logical representation, facilitating the process of problem-solving.
\citet{chen2021geoqa} introduce Neural Geometric Solver (NGS) as an approach to addressing challenges posed by geometric problems in the GeoQA benchmark~\citep{chen2021geoqa}. NGS adopts a holistic approach, adeptly parsing multimodal information and generating interpretable programs.
Geoformer~\citep{chen-etal-2022-unigeo} concurrently addresses calculation and proving problems through sequence generation. This approach demonstrates improved reasoning capabilities in both tasks by employing a unified formulation. Additionally, the authors propose the Mathematical Expression Pretraining (MEP) method, predicting mathematical expressions within problem solutions~\citep{chen-etal-2022-unigeo}. This technique enhances the model's ability to handle mathematical expressions effectively.
Inter-GPS~\citep{lu2021inter} formulates the geometry-solving task as a problem-goal-searching process. By incorporating theorem knowledge as conditional rules, Inter-GPS enables step-by-step symbolic reasoning, facilitating effective geometry problem-solving.

\subsubsection{Automated Theorem Proving}

Theorem proving is pivotal in both hardware and software verification~\citep{khan2020formal,li2005combining}. In the context of hardware verification, it has found successful application in the design of integrated circuits~\citep{khan2020formal,li2005combining}. In the realm of software verification, a notable achievement is the development of CertC, a verified C compiler~\citep{berghofer2004extracting}. It is worth mentioning that companies such as Intel have made significant investments in formal methods to ensure the 
absence of critical floating-point bugs in their processors. A prominent example of the consequences of such bugs is the costly Pentium FDIV bug in 1994, which resulted in a loss of \$500 million~\citep{harrison2010formal}. Consequently, theorem proving has played a pivotal role in verifying floating-point firmware~\citep{harrison2010formal}.
Traditionally, theorem proving has relied on highly trained human experts proficient in specific theorem proving tools and their respective application domains. However, the emergence of learnable automated theorem proving holds the potential to revolutionize hardware and software verification in two significant ways. First, it enhances the level of automation in theorem proving, making it less reliant on human expertise and manpower. Second, it increases the adaptability of these methods, broadening their utility and applicability through machine learning.

Researchers create Contemporary mathematical verification systems based on interactive theorem provers (ITPs), including Isabelle~\citep{paulson1994isabelle}, Lean~\citep{de2015lean}, Coq~\citep{barras1997coq}, and Metamath~\citep{megill2019computer}.
In recent years, various approaches have integrated machine learning with ITPs~\citep{yang2019learning, gauthier2021tactictoe}. Validated on various datasets (PISA~\citep{jiang2021lisa}, miniF2F~\citep{zheng2021minif2f}, LeanDojo~\citep{yang2023leandojo}, FIMO~\cite{liu2023fimo} and TRIGO~\citep{xiong2023trigo}), these approaches leverage advancements in language models~\citep{polu2020generative, han2021proof, polu2023formal, jiang2022thor, lample2022hypertree, mikula2023magnushammer} to recommend actions based on the current state of the proof, with a tree search identifying a sequence of correct steps using actions provided by the language model. Methods like Monte Carlo Tree Search (MCTS)~\citep{silver2018general, wu2021tacticzero, laurent2022learning} or dynamic-tree MCTS~\citep{wang2023dt} are employed for this purpose. Previous work has demonstrated the few-shot statement autoformalization capability of large language models (LLMs)~\citep{wu2022autoformalization}.
To investigate the applicability of these findings to proof autoformalization, DSP conducted a thorough analysis using Draft, Sketch, and Proof~\citep{jiang2022draft}. Subgoal-Learning~\citep{zhao2023decomposing} utilizes the subgoal-goal informal proof and demonstration selection. LeanDojo ~\citep{yang2023leandojo} is an open-source project for Lean~\citep{moura2021lean}, which contains toolkits, data, models, and benchmarks. Lyra~\citep{zheng2023lyra} proposes the use of Tool Correction to mitigate LLM hallucinations and Conjecture Correction to improve the quality of generated formal proof conjectures. Following the direction of Lyra, the LEGO-Prover~\citep{xin2023lego} employs a growing skill library containing verified lemmas as skills to enhance the capability of LLMs used in theorem proving.

\subsubsection{Scientific Reasoning}

Scientific reasoning encompasses the cognitive abilities and problem-solving skills required for formulating, evaluating, and refining hypotheses or theories. In the case of highly developed proficiency, it also involves critical reflection on the process of acquiring and evolving knowledge through these investigative activities~\citep{Morris12}. 
As mathematical reasoning forms the foundation of, we mention scientific reasoning here.

Scientific reasoning is closely relevant to AI for Science (AI4Science)~\citep{zhang2023artificial}. This relevance extends across a spectrum of fields, including physics, chemistry, quantum mechanics, and more. The integration of foundation models into these domains not only enhances our understanding but also opens up new avenues for exploration and innovation. The potential for foundation models to revolutionize traditional scientific methods, accelerate discoveries, and solve complex problems is immense, making them an indispensable tool in the modern scientific landscape. 
\citet{subramanian2023foundation} examine how various factors affect the transfer learning capabilities of foundational models, such as the size of pre-trained models, dataset scale, a blend of models, and parameters outside the training distribution. Their study finds that increasing the number of model parameters can enhance performance. Furthermore, the ``pre-train and fine-tune'' approach is highly effective for scientific reasoning tasks, particularly in physical systems governed by Partial Differential Equations (PDEs). \citet{horawalavithana-etal-2022-foundation} modify OpenAI's GPT-2 transformer decoder architecture to develop a 1.47 billion parameter general-purpose model specifically for chemistry. This large-scale model demonstrates proficiency not only in in-domain tasks but also in out-of-domain challenges. It is trained on a substantial corpus of 670GB of text data, encompassing approximately 53.45 million chemistry-focused scientific articles and abstracts.
IBM RXN for Chemistry ~\citep{GT4SD,manica2022gt4sd,das2021accelerated} utilizes foundational models for predicting chemical reactions and procedural methodologies in chemistry.
For a more comprehensive exploration of foundational models related to biology, please see Section~\ref{subsubsec:medical_reasoning} and Section~\ref{subsubsec:bio_reasoning}. We will not elaborate further on biology foundation models here.
Currently, most scientific reasoning research predominantly concentrates on fields like mathematics, physics, biology, and medicine~\citep{qiu2023large}. In contrast, foundational models in the quantum realm are comparatively scarce. 
Building scalable foundation models for quantum systems faces several challenges, including the intrinsic complexity of quantum mechanics, limited data availability, the absence of standardized methodologies, and constraints in quantum hardware capabilities. Despite these hurdles, venturing into this promising field presents an intriguing and potentially rewarding area of exploration.

Standardization aids in advancing the field of scientific reasoning. Proposing datasets or benchmarks is a process of standardization. Currently, datasets for scientific reasoning are mainly focused on fields such as mathematics, physics, and chemistry, examples of which include SciBench~\citep{wang2023scibench}, ScienceWorld~\citep{wang2022scienceworld}, and ScienceQA~\citep{lu2022learn}.
SciBench~\citep{wang2023scibench} is a specialized benchmark designed to evaluate the scientific reasoning capabilities, domain knowledge, and advanced calculation skills of LLMs in the context of college-level scientific problems. This comprehensive benchmark encompasses a meticulously curated collection of 695 problems carefully sourced from instructional textbooks. SciBench consists of two datasets. The first dataset constitutes an expansive collection of collegiate-level scientific problems sourced from mathematics, chemistry, and physics textbooks. Its primary objective is to evaluate the LLM's capacity to handle a diverse array of scientific topics and problem categories. The second dataset in SciBench, on the other hand, consists of problems sourced from computer science and mathematics undergraduate exams, forming a closed set. This closed set is intentionally crafted to gauge the LLMs' proficiency in solving precise problem-solving challenges within these particular fields.
ScienceWorld~\citep{wang2022scienceworld} is designed to evaluate agents' scientific reasoning capabilities within an interactive text environment. 
This environment simulates a standard elementary school science curriculum, featuring 30 high-level task types distributed across 10 different topics. The environment supports multiple states, allowing for diverse interactions and scenarios. By abstracting the world and incorporating a wide range of objects, ScienceWorld provides a complex interactive text environment for agents to navigate and reason therein.
It consists of 10 interconnected locations, each containing up to 200 types of objects. These objects span a range of categories, and common environmental items like furniture, books, and paintings. The environment provides a rich and diverse setting for agents to interact with. The action set within ScienceWorld consists of 25 high-level actions, covering actions related to the domain of science and common actions. Each step in ScienceWorld presents approximately 200,000 possible action-object pairs, although only a proportion of these pairs will have actual implications for the task at hand.
ScienceQA~\citep{lu2022learn} is a multimodal dataset comprising 21,208 multiple-choice science questions sourced from elementary and high school science curricula. The dataset offers a richer domain diversity by covering natural science, language science, and social science topics.

These resources provide valuable platforms for testing the capabilities of foundation models in complex scientific reasoning domains, allowing for a more structured approach to assessing their reasoning abilities. The focus on these traditional sciences highlights the need for expanding the scope of datasets to encompass a wider range of disciplines, potentially leading to more diverse and comprehensive advancements in scientific reasoning.

\subsection{Logical Reasoning}~\label{sec:logical_reasoning}

Logical reasoning, covering propositional and predicate logic (Table~\ref{table:logic_comparison}), is a rigorous form of thinking that involves using premises and their relations to derive conclusions that are implied by the premises~\citep{Nunes2012}. It can serve as a fundamental basis for various domains in computer science and mathematics.

Previous studies have explored the combination of neural networks and symbolic reasoning in neuro-symbolic methods~\citep{Mao2019NeuroSymbolic,pryor2023neupsl,tian2022weakly,cai2021abductive,pmlr-v155-sun21a,MANHAEVE2021103504,gupta2019neural}. However, these methods often face limitations such as specialized module designs that lack generalizability or brittleness caused by optimization difficulties.
In contrast, LLMs exhibit stronger generalization abilities when it comes to logical reasoning. The Logic-LM framework~\citep{pan2023logic} leverages LLMs and symbolic reasoning to enhance logical problem-solving~\citep{luo2023logiglue}. It begins by utilizing LLMs to convert natural language problems into symbolic formulations, which are then processed by deterministic symbolic solvers for inference. Additionally, a self-refinement stage is introduced, where error messages from the symbolic solver are utilized to revise the symbolic formalizations. \citet{bubeck2023sparks} demonstrate that the GPT-4 model can manifest logical reasoning abilities when addressing mathematical and general reasoning problems. These higher-order capabilities, often referred to as emergent properties, result from scaling the model with large datasets~\citep{wei2022emergent}. \citet{zhao2023explicit} employ language models for multi-step logical reasoning by integrating explicit planning into their inference procedure. This incorporation enables more informed reasoning decisions at each step by considering their future effects.
Furthermore, \citet{creswell2023selectioninference} propose the Selection-Inference (SI) framework, which employs pre-trained LLMs as general processing modules. The SI framework alternates between selection and inference steps to generate a sequence of interpretable, causal reasoning steps that lead to the final answer.

\begin{figure}[]
	\begin{center}
		
\includegraphics[width=0.8\columnwidth]{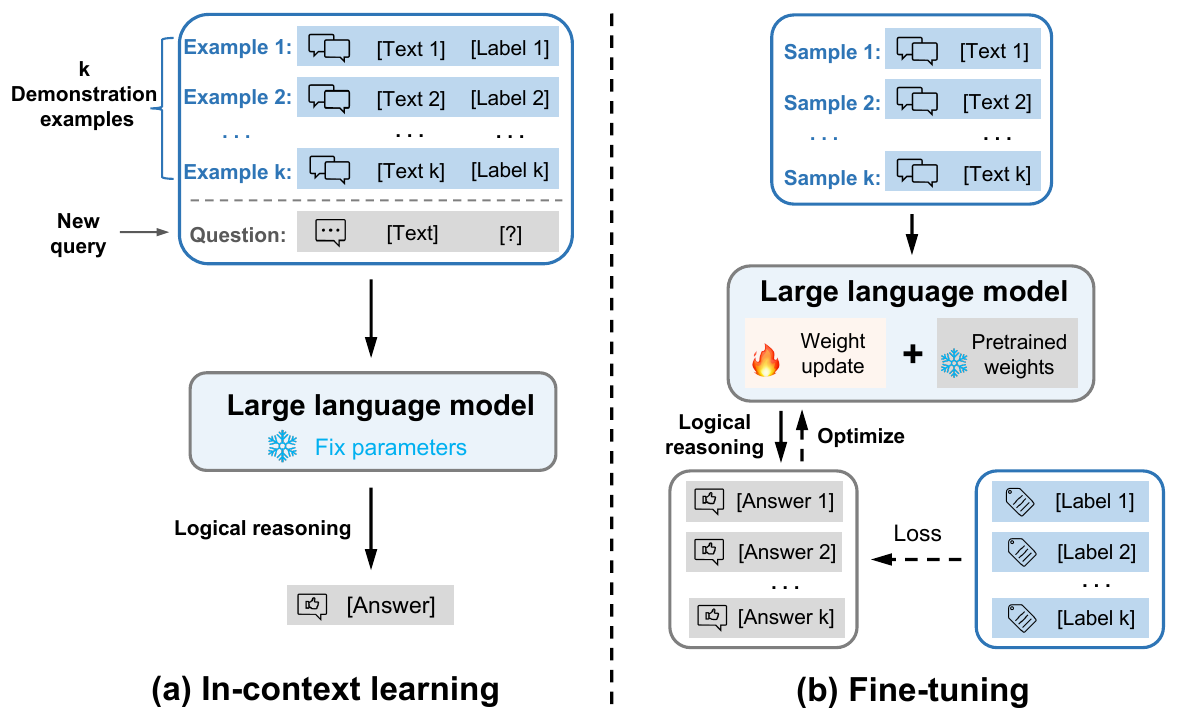}
	\end{center}
	\caption{Two main approaches to enhancing logical reasoning capabilities of large language models. (a) In-context learning leverages specific prompts as a demonstration to elicit logical reasoning. (b) Fine-tuning uses additional training samples to update the specialized model parameters.}
\label{fig:reason_icl_ft}
\end{figure}

Recent works leveraging LLMs for logical reasoning tasks can be categorized into two main approaches, as shown in Figure~\ref{fig:reason_icl_ft}. The first approach is in-context learning, where specific prompts are used to elicit step-by-step reasoning from LLMs. Notable methods in this category include chain-of-thought prompting~\citep{wei2022chain,wang2023selfconsistency} and the least-to-most prompting approach~\citep{zhou2022least}. These approaches enable reasoning directly over natural language, providing flexibility. However, the complexity and ambiguity of natural language can result in challenges such as unfaithful reasoning and hallucinations. The second approach is fine-tuning, where the reasoning capabilities of LLMs are optimized through fine-tuning or training specialized modules~\citep{clark2020transformers,tafjord-etal-2022-entailer,yang-etal-2022-generating}.

\subsubsection{Propositional Logic}

\begin{table}[]
\centering
\setlength{\tabcolsep}{2.5pt}
\resizebox{1\linewidth}{!}{
\begin{tabular}{l|p{5cm}|p{5cm}}
\toprule
    & Propositional Logic & Predicate Logic   \\
\midrule
Basic elements & Atomic propositions, Compound propositions  & Atomic propositions, Compound propositions, Variables, Quantifiers, Predicates \\
\midrule
Complexity         & Lower               & Higher             \\
\midrule
Expressive Power   & Limited             & More powerful      \\
\midrule
Applications       & Circuit design, Boolean algebra  & Natural language processing, Knowledge representation, Database queries \\
\midrule
Examples           & \( p \lor q \); \( p \land q \); \( \lnot p \); \(p \rightarrow q \) & \( \forall x, P(x) \); \( \exists x, P(x) \) \\
\bottomrule
\end{tabular}
}
\caption{Comparison between Propositional Logic and Predicate Logic in terms of basic elements, complexity, expressive power, and applications.}
\label{table:logic_comparison}
\end{table}

Propositional logic deals with declarative sentences that can be assigned a truth value, either true or false, without any ambiguity. There are two types of propositional logic: Atomic Propositions and Compound Propositions. Atomic propositions are basic statements that cannot be further broken down, while compound propositions are formed by combining atomic propositions using logical connectives such as conjunction (AND), disjunction (OR), and negation (NOT).

In the context of propositional logic resolution, \citet{tomasic2021propositional} performed fine-tuning on the GPT-2 and GPT-3 models, tailoring them for the purpose of simulating propositional logic resolution. This specialized training focuses on non-recursive rules that encompass conjunction, disjunction, and negation connectors. By leveraging these language models, they aimed to enhance the logical reasoning capabilities in propositional logic problems.

The use of language models for propositional logic resolution is intriguing because these models have demonstrated their ability to capture complex patterns and semantic relationships in natural language. By training them to understand and reason with propositional logic, researchers sought to improve their logical reasoning capabilities.

\subsubsection{Predicate Logic}

Predicate Logic, also known as First-order Logic, can be seen as an extension of propositional logic, allowing for more nuanced expressions. In Predicate Logic, predicates are used to represent properties and provide additional information about the subject of a sentence. It involves variables with a specified domain and encompasses objects, relations, and functions between those objects.

Inductive Logic Programming (ILP) is a specialized domain within the broader field of machine learning~\citep{cropper2022inductive}. ILP leverages first-order logic to represent hypotheses and data, making logical language a crucial component in knowledge representation and reasoning~\citep{de2010statistical}.

By incorporating predicate logical representations and reasoning, LLMs offer the potential for more interpretable and explainable models~\citep{liu-etal-2022-rlet}. It enables the discovery of logical patterns and rules from data, facilitating the extraction of human-understandable knowledge.

\subsection{Causal Reasoning}~\label{sec:causal_reasoning}

\begin{figure}[tbp]
	\begin{center}
		
\includegraphics[width=1.0\columnwidth]{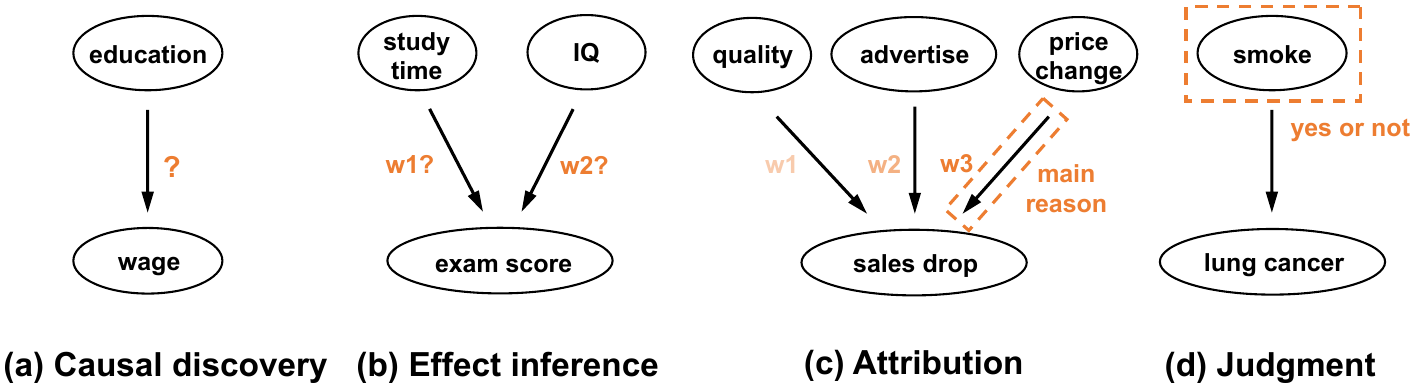}
	\end{center}
	\caption{Examples of causal graphs to reflect different casual reasoning tasks. (a) Causal discovery identifies the underlying causal relationships among variables in a given system. (b) Effect inference estimates the outcome (e.g., weight) of a specific intervention on a system based on known causal relationships. (c) Attribution determines the extent to which a particular cause is responsible for a given effect. (d) Judgment makes decisions based on the perceived consequences and implications of causal relationships.}
\label{fig:casual_reason}
\end{figure}

Causal reasoning refers to the process of understanding and explaining cause-and-effect relationships between events, actions, or variables~\citep{waldmann2013causal,liu2023magic}. Causal reasoning tasks can be categorized into causal discovery, effect inference, attribution, judgment, and other tasks~\citep{kiciman2023causal}. 
Causal discovery is the process of uncovering the directional cause-and-effect relationships between variables. Effect inference involves the characterization of the magnitude and pattern of a known or postulated causal connection~\citep{lyu2022can,wang2021inferbert,jin2023can}. Attribution, on the other hand, entails identifying the cause or causes behind a specific change. Judgment tasks expand on attribution tasks by encompassing the assignment of reward or blame for outcomes. Additionally, these tasks encompass various domains such as policy optimization, decision-making, explanation, scientific discovery, and more.

A causal graph, also known as a causal network or causal diagram, is a graphical representation of causal relationships between variables or events~\citep{balashankar-subramanian-2021-learning,scholkopf2021toward}. It is a visual tool used to depict cause-and-effect relationships and understand the causal structure of a system or phenomenon.
In a causal graph, variables or events are represented by nodes, and causal relationships between them are depicted by directed edges or arrows. In Figure~\ref{fig:casual_reason}, we use causal graphs to illustrate multiple reasoning tasks mentioned above.

\paragraph{Causal Discovery}
Causal discovery~\citep{peters2017elements} involves the task of identifying the causal graph~\citep{long2022can} that represents the underlying process responsible for generating observed data. LLMs have demonstrated competitive performance in discerning pairwise causal connections, although their effectiveness can vary and is influenced by the careful crafting of prompts. \citet{long2022can} investigate the limitations of GPT-3 in understanding causal relationships in the medical context.
Within the framework of Neuropathic Pain Diagnosis~\citep{tu2019neuropathic}, \citet{tu2023causaldiscovery} find that ChatGPT tends to make false negative mistakes. The performance of LLMs in causal discovery is not yet stable or consistent, and they may provide different answers to the same question, potentially due to internal model updates.
\citet{long2023causal} suggest that expert knowledge, including that of LLMs, may be incorrect. They propose leveraging imperfect experts, such as LLMs, to reduce uncertainty in the output of causal discovery algorithms. By incorporating the expertise of LLMs into the statistical analysis of objective data, they aim to improve the accuracy of causal structure learning. Advancing the current research on LLM-driven causal discovery, \citet{ban2023query} integrate knowledge-based LLM causal analysis with data-driven approaches to learning causal structures. They effectively combine the expertise of LLMs regarding existing causal mechanisms with the statistical analysis of objective data. They devise a specialized set of prompts aimed at deriving causal graphs from specific variables. By employing these prompts, they evaluate the impact of LLM-informed causality on deducing causal structures from data. Compared to text-only LLMs, Code-LLMs~\citep{liu2023magic} with code prompts are better in causal reasoning.

\paragraph{Type Causality and Actual Causality}
Type causality pertains to the inference of causal relationships between variables, which is evident in causal discovery and causal effect estimation. In contrast, actual causality~\citep{halpern2016actual} diverges from causal discovery by shifting the focus from variables and their interrelationships to individual events, with the goal of uncovering their specific causes.

CausaLM~\citep{feder2021causalm} has demonstrated that language models like BERT~\citep{kenton2019bert} can obtain a counterfactual representation of a particular concept of interest through the deliberate selection of auxiliary adversarial pre-training tasks. This counterfactual representation enables the prediction of the concept's actual causal effect on model's performance. On the other hand, \citet{zhang2023understanding} believe that current LLMs can address causal questions by leveraging existing causal knowledge, akin to combined domain experts. However, these models still struggle to provide satisfactory answers when it comes to discovering new knowledge or performing high-stakes decision-making tasks with a high level of precision. Current LLMs lack the ability to incorporate actual physical data measurements to establish a grounding for their available textual facts~\citep{zevcevic2023causal,willig2023probing}. As a result, they are unable to engage in actual, inductive inference, similar to classical (causal) structure discovery methods. This limitation raises a crucial societal discussion point regarding the process of learning from facts. It is arguable that the ideal goal should be \textit{understanding} rather than mere \textit{knowing}, as the latter lacks both generalization and justification.

In summary, while LLMs show promise in causal discovery, their performance is still inconsistent and sensitive to prompt engineering. Researchers are exploring ways to address these limitations.

\subsubsection{Counterfactual Reasoning} 

Counterfactuals involve a premise that is false in the real world but assumed to be true in a hypothetical scenario. For example, ``If cats were vegetarians,'' followed by an imaginary consequence like ``cats would love cabbages''~\citep{li2023counterfactual}. Counterfactual reasoning involves the consideration of hypothetical scenarios achieved by altering elements or conditions within an actual event or situation~\citep{kahneman1986norm,byrne2007rational}. It plays a fundamental role in understanding causality, enabling us to explore the potential outcomes that could have arisen under different circumstances. By subjecting language models to counterfactual testing, we can manipulate the factual accuracy and hypothetical nature of statements, thereby evaluating the models' capacity to discern and effectively utilize this information in making predictions. This testing approach enables us to gain insights into the models' aptitude for differentiating between actual and hypothetical scenarios and their ability to leverage this understanding for accurate and contextually appropriate responses.

In the context of language models, each reasoning task can be represented as a mapping function $f_w: X\rightarrow Y$, which maps an \textbf{input} $x\in X$ using a \textbf{world model} $w\in W$ to an \textbf{output} $y\in Y$. The world model encapsulates the conditions under which the function evaluation occurs, with the \textbf{default world} denoted as $w^{\textrm{default}}$. Hypothesis $h$ estimates $f^w$, while counterfactual worlds are represented as $w^{\textrm{cf}}$. The language model's implementation of $f_w$ for a given instance $x$ can be expressed as:
\begin{equation}
h(f, w, x) = \argmax_{y'}P_{{\textrm{LM}}}(y'|\textrm{prompt}_f(f,x), \textrm{prompt}_w(w)),
\end{equation}
where counterfactual reasoning is denoted as $h(f, w^{{\textrm{cf}}}, x)$ and factual reasoning as $h(f, w^{\textrm{default}}, x)$.

\citet{li2023counterfactual} utilize counterfactual conditionals to examine the ability of pre-trained language models (PLMs) in distinguishing between hypothetical and real-world scenarios. They explore how this capability interfaces with the models' utilization of pre-existing real-world knowledge and associative cues. Their findings reveal that when confronted with counterfactual situations, PLMs tend to generate completions that contradict established world knowledge. As an example, GPT-3 might have developed a nuanced comprehension of how linguistic cues, such as distinguishing between "If/had" and "Because," influence the connection between nearby lexical clues and the following words. This suggests that PLMs might prioritize the influence of immediate contextual cues over broader factual information when responding to counterfactual prompts. 
\citet{wu2023reasoning} introduce an evaluation framework that incorporates ``counterfactual'' task variations. They present a set of 11 counterfactual tasks and assess the capability and performance of GPT-4~\citep{openai2023gpt4}, Claude~\citep{anthropic2023introducing}, and PaLM-2~\citep{anil2023palm} on these tasks, considering both default and counterfactual conditions. The findings indicate that while current language models possess some degree of abstract task-solving abilities, their performance often relies on narrow, context-specific procedures that are not easily transferable across tasks. Notably, the models consistently exhibit a significant decrease in performance when confronted with counterfactual task variants compared to the default settings. These results emphasize the need for a careful interpretation of language model performance, taking into account different aspects of their behaviors and the challenges posed by counterfactual reasoning.

\subsection{Visual Reasoning}~\label{sec:visual_reasoning}

\begin{figure}[tbp]
	\begin{center}
		
\includegraphics[width=1.0\columnwidth]{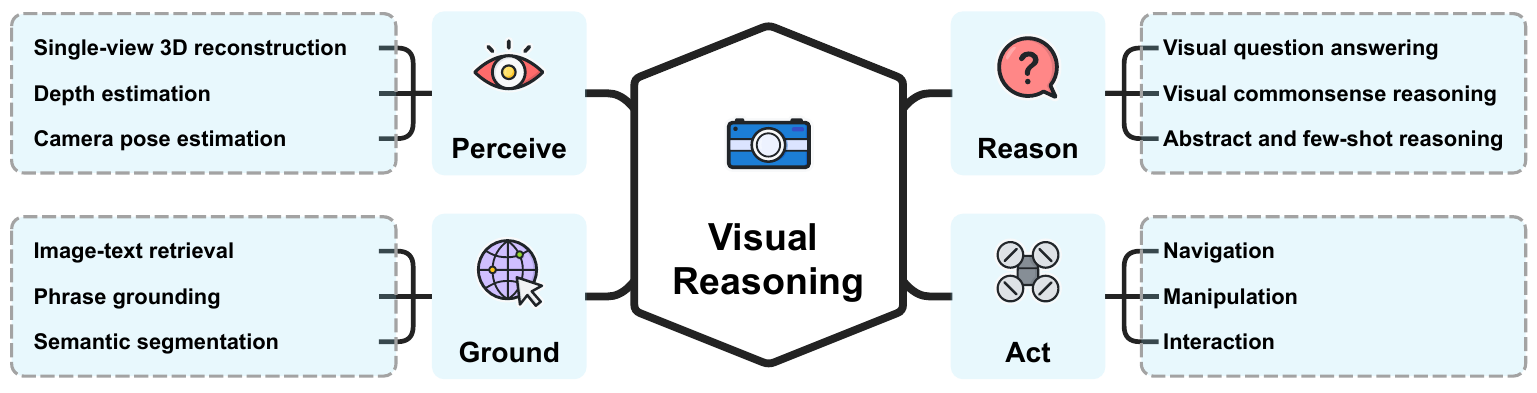}
	\end{center}
        \vspace*{2mm}
	\caption{Four functional domains of a general vision systmem G-VUE~\citep{huang2023perceive} and their corresponding visual tasks.}
\label{fig:visual_reasoning}
\end{figure}

Visual reasoning refers to the cognitive process of understanding, analyzing, and drawing conclusions from visual information. It involves the ability to perceive, interpret, and reason about visual stimuli such as images, scenes, or other visual representations~\citep{ding2023visual}.

General-purpose Visual Understanding Evaluation (G-VUE) is a comprehensive evaluation framework~\citep{huang2023perceive}. It aims to assess the full range of visual cognitive abilities. The framework is divided into four functional domains: Perceive, Ground, Reason, and Act. As shown in Figure~\ref{fig:visual_reasoning}, the framework encompasses a thoughtfully chosen collection of 12 tasks, including 3D reconstruction, visual reasoning, manipulation, and interaction, to represent these domains. G-VUE serves as a standardized and comprehensive platform for assessing the visual understanding capabilities of AI systems. By prioritizing diverse functional domains and carefully selecting tasks, the framework ensures the inclusion of a wide range of visual cognitive abilities. This enables more accurate evaluations of the strengths and weaknesses of AI systems.
VLGrammar~\citep{hong2021vlgrammar} is a model that employs compound probabilistic context-free grammars (PCFGs) to simultaneously induce language and image grammar. A novel contrastive learning framework is also proposed, which facilitates the joint learning of these two modules. AeNER~\citep{ding2021attention} introduces a general neural-network-based approach to dynamic visual spatio-temporal reasoning problems. This approach differs from bespoke methods like modular symbolic components, independent dynamics models, or semantic parsers. AeNER offers a more versatile and adaptable solution for addressing dynamic visual spatio-temporal reasoning challenges. LISA~\citep{lai2023lisa} proposes a new vision task termed reasoning segmentation, aiming to obtain the mask(s) of corresponding object(s) from an implicit query text. It requires world knowledge from LLMs along with image understanding to accurately locate the object of interest.

\subsubsection{3D Reasoning}
Specifically, 3D reasoning refers to the cognitive process of understanding, analyzing, and reasoning about 3D objects or spatial arrangements. 

3D-LLM~\citep{3dllm} is designed to process 3D point clouds along with their associated features. It demonstrates remarkable proficiency across a diverse spectrum of 3D-related tasks, encompassing dense captioning, 3D question answering, 3D grounding, 3D-assisted dialogue, navigation, task decomposition, and beyond.
PointLLM~\citep{xu2023pointllm} is an approach that extends LLMs to understand 3D point clouds, combining geometric, visual, and textual information to interact with and interpret 3D data. It shows superior performance in object classification and captioning tasks compared to 2D baselines.
On the other hand, 3D-VisTa~\citep{3dvista} is a pre-trained Transformer model specifically developed for aligning 3D vision and text. It proves to be highly valuable for 3D vision-language (3D-VL) tasks such as 3D visual grounding, dense captioning, and situated reasoning~\citep{ma2022sqa3d}.

Research into 3D reasoning with foundational models is at a fascinating juncture. Models like 3D-LLM, PointLLM, and 3D-VisTa have already showcased their effectiveness in diverse 3D tasks, blending geometric, visual, and textual data. Despite these advancements, the field is still burgeoning, with much room for exploration and enhancement. Future directions could include refining model capabilities for more intricate 3D scene interpretations, expanding applications in real-world scenarios such as navigation agents for visually impaired or blind people~\citep{qiu2022egocentric}, and bridging gaps in current methodologies. 

\subsection{Audio Reasoning}
\label{sec:audio_reasoning}

Audio reasoning pertains to the cognitive mechanism of comprehending, examining, and deriving conclusions from auditory data, of which speech is the major source. Speech representations learned in a self-supervised fashion provide a promising solution in this direction, where a single foundation model is trained and can be applied to a wide spectrum of downstream tasks~\citep{mohamed2022self}.

\begin{figure}[tbp]
    \begin{center}
        \includegraphics[width=1.0\columnwidth]{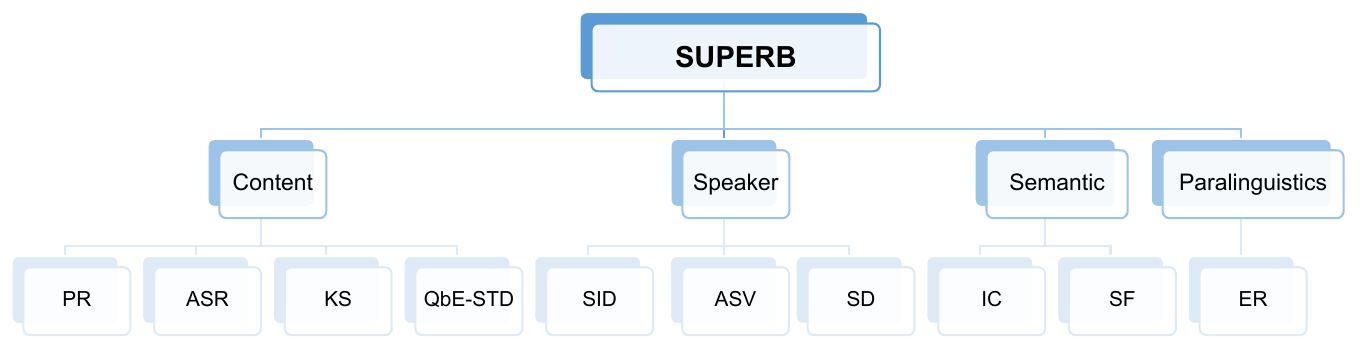}
    \end{center}
    \vspace*{2mm}
    \caption{Four evaluation areas of the SUPERB~\citep{yang21superb} focus on the discriminative abilities of foundation models and the corresponding tasks. PR: phone recognition, ASR: automatic speech recognition, KS: keyword spotting, QbE-STD: query by example spoken term detection, SID: speaker identification, ASV: automatic speaker verification, SD: speaker diarization, IC: intent classification, SF: slot filling,
    ER: emotion recognition.}
\label{fig:audio_superb}
\end{figure}

\subsubsection{Speech}

The field of speech processing can be broadly classified into two distinct categories: discriminative tasks and generative tasks. Discriminative tasks entail the process of making discrete decisions based on continuous speech, while generative tasks involve the generation of continuous speech from diverse input sources. The Speech processing Universal PERformance Benchmark (SUPERB)~\citep{yang21superb} is a widely adopted framework for evaluating the discriminative abilities of the foundation model. As demonstrated in Figure~\ref{fig:audio_superb}, it encompasses ten tasks covering four elements of speech: Content, Speaker, Semantics, and Paralinguistics.

The enhanced Speech processing Universal PERformance Benchmark (SUPERB-SG)~\citep{tsai2022superb} further introduces a framework to evaluate the generative abilities of the foundation model with five tasks: speech translation (ST), out-of-domain automatic speech recognition (OOD-ASR), voice conversion (VC), speech separation (SS), and speech enhancement (SE).

The foundation models for learning self-supervised speech representation can be categorized into three major types: 1) \textbf{generative models} that reconstruct the input speech sequence leveraging on restricted or corrupted views, for example, vector-quantized variational autoencoder (VQ-VAE)~\citep{van2017neural}, autoregressive predictive coding (APC)~\citep{chung2019unsupervised}, and masked acoustic model (MAM)~\citep{liu2020mockingjay}; 2) \textbf{contrastive models} that differentiate a target positive sample from distracting negative samples, for example, contrastive predictive coding (CPC)~\citep{oord2018representation}, Wav2Vec 2.0~\citep{baevski2020wav2vec}, and Speech SIMCLR~\citep{jiang2020speech}; and 3) \textbf{predictive models} that follow the settings similar to teacher-student learning~\citep{li2017large}, for example, Hidden Unit BERT (HuBERT)~\citep{hsu2021hubert}, WavLM~\citep{chen2022wavlm} and Data2Vec~\citep{baevski2022data2vec}. The Transformer-Encoder~\citep{dong2018speech} architecture and the Conformer-Encoder~\citep{gulati2020conformer} architecture are widely adopted in speech foundation models.

\subsection{Multimodal Reasoning}~\label{sec:multimodal_reasoning}

\begin{figure}[htbp]
	\begin{center}
		
\includegraphics[width=1.0\columnwidth]{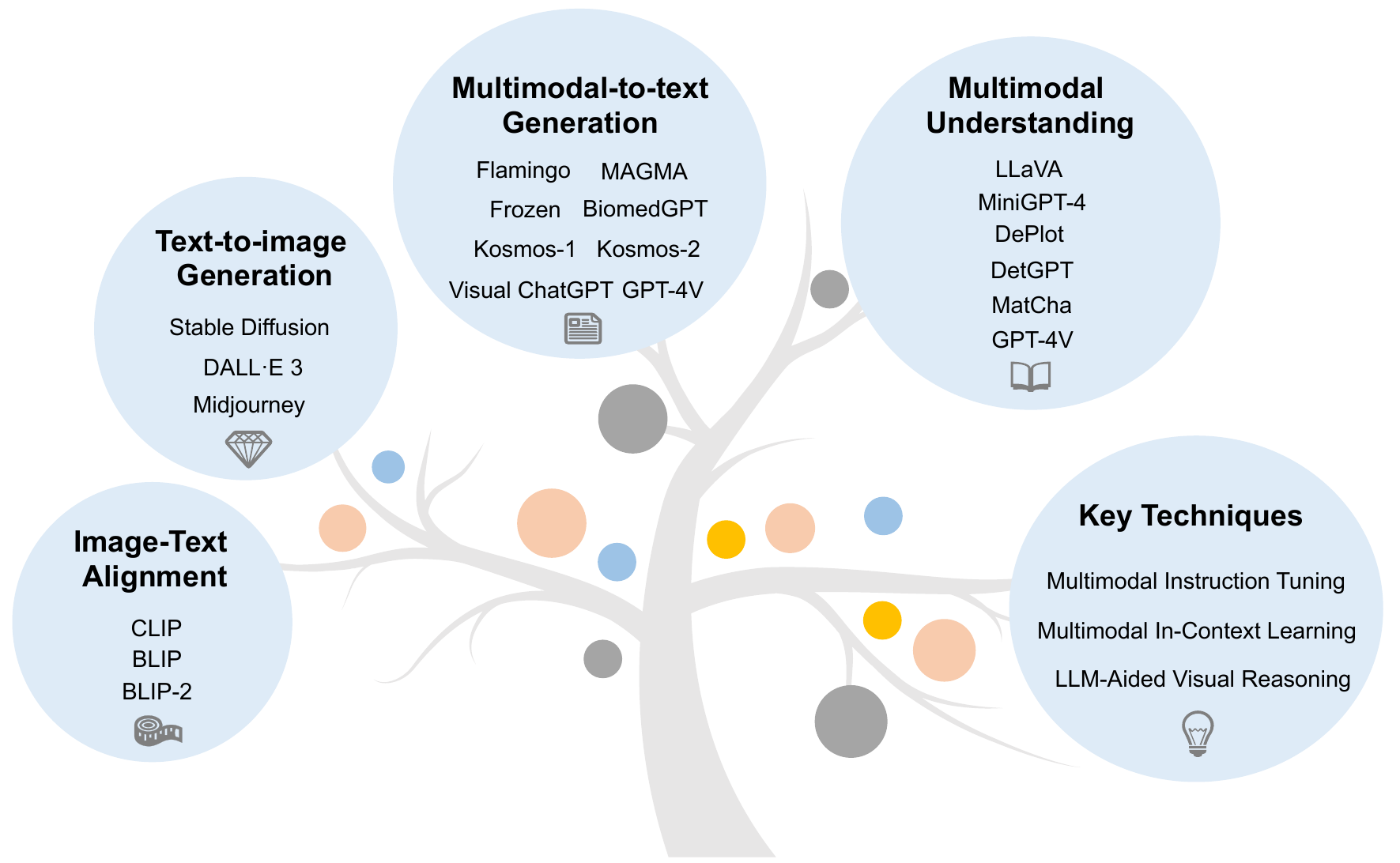}
	\end{center}
       
	\caption{Multimodal reasoning tasks can be broadly categorized into image-text alignment, text-to-image generation, multimodal-to-text generation, and multimodal understanding. Current multimodal foundation models mainly involves three key techniques to approach reasoning tasks, including multimodal instruction tuning, multimodal in-context learning, and LLM-aided visual reasoning. The figure style credits from tutorial~\citep{li2023large}.}
\label{fig:multimodal_reasoning}
\end{figure}

Multimodal reasoning refers to the cognitive process of integrating and reasoning across multiple modalities of information, such as text, images, videos, and other sensory inputs, to enhance understanding and perform complex reasoning tasks~\citep{yin2023survey,zong2023selfsupervised}\footnote{\url{https://github.com/atfortes/Awesome-Multimodal-Reasoning}}. 

In the pursuit of developing Artificial General Intelligence (AGI), multimodal reasoning represents a promising advancement over unimodal approaches for several reasons. Firstly, multimodal reasoning aligns more closely with the way humans perceive the world. Humans naturally receive inputs from multiple senses, which often complement and cooperate with each other. As a result, leveraging multimodal information is anticipated to enhance the intelligence of Multimodal Foundation Models. Secondly, multimodal reasoning provides a more user-friendly interface. By incorporating support for multimodal input, users can interact and communicate with intelligent assistants in a more flexible, diverse and potentially more intuitive manner, improving the overall user experience. Thirdly, multimodal reasoning facilitates a more comprehensive problem-solving capability. While unimodal language models typically excel in natural language processing (NLP) tasks, Multimodal Foundation Models have the potential to support a broader spectrum of tasks, making them more versatile and effective as task-solvers.
Key techniques and applications of Multimodal Foundation Models encompass various areas, including Multimodal Instruction Tuning (M-IT), which focuses on fine-tuning models based on multimodal instructions; Multimodal In-Context Learning (M-ICL), which leverages contextual information to enhance multimodal reasoning; and LLM-Aided Visual Reasoning (LAVR), which utilizes LLMs to enhance visual reasoning capabilities. Figure~\ref{fig:multimodal_reasoning} shows multiple multimodal reasoning tasks and the key techniques behind, which are introduced as follows.

\subsubsection{Alignment}
\paragraph{Image-Text Alignment}
CLIP~\citep{radford2021learning} utilizes a learning method that enables the creation of cohesive representations for both images and text. By aligning visual and textual information, CLIP fosters cross-modal comprehension and demonstrates exceptional proficiency across a wide range of vision and language tasks. In a similar vein, BLIP-2~\citep{li2023blip2} adopts a strategy to facilitate efficient cross-modal alignment without fine-tuning the vision encoder. Instead, it introduces a Querying Transformer (Q-Former) that extracts visual features from a fixed image encoder. These extracted query embeddings serve as soft visual prompts for the alignment process. Flamingo~\citep{alayrac2022flamingo} bridges pretrained vision and language backbones by token fusion with cross-attentions.

\subsubsection{Generation}
\paragraph{Text-to-image Generation}
Stable Diffusion~\citep{rombach2022high} integrates cross-attention layers to the model architecture, transforming diffusion models into robust and adaptable generative models for diverse conditional inputs like text and bounding boxes.
The application of latent diffusion models (LDMs) represents a significant breakthrough in image inpainting, while also delivering impressive results in unconditional content generation, super-resolution image generation, and other tasks. Notably, LDMs offer substantial reductions in computational demands compared to pixel-based diffusion models, while maintaining highly competitive performance. DALL·E\footnote{\url{https://openai.com/dall-e-3}} ~\citep{ramesh2021zero,ramesh2022hierarchical, betker2023improving} is an advanced AI system that has the capability to generate realistic images and artwork based on natural language descriptions. Likewise, Midjourney is another AI system that specializes in generating images based on natural language descriptions, which are referred to as ``prompts''. By leveraging the power of AI, Midjourney\footnote{\url{https://www.midjourney.com}} can translate textual prompts into visual compositions, providing a visual representation of the given description. ImageGen~\citep{saharia2022photorealistic} leverages the capabilities of expansive transformer language models for text comprehension and combines this with the efficacy of diffusion models for creating high-quality images. PixArt~\citep{chen2023pixart} is a Transformer-driven Text-to-Image (T2I) diffusion model. It rivals leading image generation systems such as Imagen, SDXL, and Midjourney in terms of quality, approaching the benchmarks set by commercial applications.

\paragraph{Multimodal-to-text Generation}
Flamingo-80B~\citep{alayrac2022flamingo} comprises a family of Visual Language Models (VLMs) equipped with in-context few-shot learning capabilities. These models undergo thorough evaluation across a wide array of tasks, including open-ended ones like visual question-answering and captioning, as well as closed-ended tasks such as multiple-choice visual question-answering.
Frozen~\citep{tsimpoukelli2021multimodal}  accomplishes few-shot learning ability within a multimodal context by preserving the language capabilities of a Language Model (LM) while incorporating visual information as a prefix. Frozen achieves this by freezing the LM and training a separate vision encoder to represent images. In the Frozen approach, visual information is represented as a sequence of embeddings, serving as a visual prefix.
MAGMA~\citep{eichenberg-etal-2022-magma} follows a similar approach to Frozen by incorporating a new image prefix encoder while keeping the language model frozen. It trains a series of Visual Language models capable of generating text autoregressively from combined visual and textual inputs. 
Visual ChatGPT~\citep{wu2023visual} and GPT-4~\citep{openai2023gpt4} represent advancements in extending chatbot capabilities to encompass multimodal applications that support both image and text prompts. Visual ChatGPT builds upon the foundation of ChatGPT and incorporates visual models. It incorporates a Prompt Manager that manages the histories of various visual foundation models, enabling a comprehensive multimodal conversation experience.
On the other hand, GPT-4 takes a different approach by accepting prompts that consist of both images and texts. This flexibility empowers users to specify vision and language tasks by generating text outputs in response to arbitrarily interlaced text and image prompts.
Microsoft has also proposed a series of Multimodal Foundation Models, including Kosmos-1~\citep{huang2023language} and Kosmos-2~\citep{peng2023kosmos2}. These models further contribute to the development of multimodal capabilities and facilitate rich interactions involving both images and text.
Furthermore, there are ongoing efforts to adapt GPT to specific domains, such as BiomedGPT~\citep{zhang2023biomedgpt}, which focuses specifically on biomedical research. These domain-specific adaptations aim to enhance the language model's performance and applicability within specialized fields.

\subsubsection{Multimodal Understanding}
Visual Instruction Tuning~\citep{liu2023visual} presents a groundbreaking approach that utilizes GPT-4 to generate multimodal language-image instruction-following data. This approach has the potential to reduce the reliance on manual annotation of large multimodal datasets. Expanding on this foundation, LLaVA (Large Language and Vision Assistant)~\citep{liu2023visual} represents an extensively trained, large-scale multimodal model. It seamlessly integrates a vision encoder with Vicuna~\citep{vicuna2023}, facilitating versatile visual and language comprehension for general-purpose applications. LLaVA excels across a diverse spectrum of tasks necessitating multimodal understanding, encompassing visual question-answering, image captioning, and instruction-following. Notably, it achieves impressive performance on Science QA~\citep{lu2022learn}, a multimodal reasoning dataset in the science domain.

In the domain of reasoning on charts, DePlot~\citep{liu-2022-deplot} presents a few-shot solution for visual language reasoning. It tackles the challenge through a two-step process: first, translating the plot into text, and then performing reasoning over the translated text. The authors also investigate the combination of DePlot with LLMs to further enhance performance.
MatCha (Math reasoning and Chart derendering pretraining)~\citep{liu-2022-matcha} introduces a comprehensive framework for visual language understanding in the chart domain. It highlights the importance of two critical components: understanding layout, including number extraction and organization, and mathematical reasoning. To enhance visual language understanding, the authors propose two complementary pretraining tasks: chart derendering, which involves generating the underlying data table or code used to create a given plot or chart, and math reasoning.

DetGPT~\citep{pi2023detgpt} revolutionizes object detection through its reasoning-based approach. It enables the automatic localization of objects of interest based on user-expressed desires, even in cases where the object is not explicitly mentioned. This innovative method incorporates reasoning capabilities to enhance the object detection process.
Q-Bench~\citep{wu2023q} demonstrates that the multimodal foundation models can perceive low-level visual attributes and provide image quality understanding.
LLaMA-VID~\citep{li2023llama} enhances LLMs for more efficient video and image understanding.
It represents each video frame with two tokens, which decreases the burden of processing long videos without sacrificing essential information.
To allow users to interactively control the focus of multimodal understanding, Prompt Highlighter~\citep{zhang2023prompt} highlights specific prompt spans and effectively guides autoregressive generation to produce more targeted outputs.

Integrating diverse data types such as text, images, tables, and audio presents distinct challenges for multimodal foundation models compared to their unimodal counterparts. A primary obstacle lies in effectively merging these varied data formats, a task complicated by issues like inconsistency and incompleteness in datasets, where mismatches between image content and corresponding descriptions, or missing data, can adversely affect model performance. Additionally, multimodal foundation models typically demand substantial computational resources for training. Exploring efficient training methods for these models thus emerges as a valuable area of research, crucial for advancing the capabilities of multimodal AI systems. These multimodal foundation models are also instrumental in learning universal representations applicable to fields like materials science, chemistry, and biology~\citep{GT4SD,manica2022gt4sd}.

\subsection{Agent Reasoning}~\label{sec:embodied_reasoning}

Agent reasoning, is an important capability for the Autonomous Language Agents, which refers to a cognitive process that integrates perception, action, and interaction with the physical environment or simulated environment to support reasoning and problem-solving. Autonomous Agents in the context of Large Language Models have the ability to perform a wide range of tasks, such as task decomposition, generating code, answering questions, engaging in dialogue, providing recommendations, and more. Autonomous Agents, often known as AI Agents, harness the power of Large Language Models to autonomously perform tasks, utilizing their extensive knowledge, reasoning skills, and vast informational resources~\citep{alibali2014gesture}.

Several works have investigated the use of language for planning purposes~\citep{jansen-2020-visually,li2022pre,sharma2021skill,zeng2023socratic,huang2022language,Ahn2022DoAI,mu2023embodiedgpt,hu2023tree,zhou2023generalizable}. Recent methods in task planning utilize pre-trained autoregressive foundation models to break down abstract, high-level instructions into executable, low-level step sequences for an agent, applying a zero-shot approach~\citep{huang2022language,Ahn2022DoAI}.
Specifically, \citet{huang2022language} prompt GPT-3~\citep{brown2020language} and Codex~\citep{chen2021evaluating} to create actions for agents, where each action step is semantically converted into a permissible action through Sentence-RoBERTa~\citep{liu2019roberta,reimers2019sentence}. In contrast, SayCan~\citep{Ahn2022DoAI} grounds the actions and language by combining the probability of each candidate action, as determined by FLAN~\citep{wei2021finetuned}, with the action's value function. The latter acts as a surrogate for measuring affordance~\citep{shah2021value}. However, both approaches assume the successful execution of each proposed step by the agent, without considering potential intermediate failures in dynamic environments or accounting for the performance of lower-level policies. 
SwiftSage~\citep{lin2023swiftsage} is a framework influenced by the dual-process theory of human cognition, tailored for superior performance in action planning within intricate interactive reasoning tasks. This framework is structured around two main components: the SWIFT module and the SAGE module.
The SWIFT module represents fast and intuitive thinking and is responsible for action planning based on the oracle agent's action trajectories. It is implemented as a small encoder-decoder language model that has been fine-tuned specifically for this purpose.
On the other hand, the SAGE module emulates deliberate thought processes and utilizes LLMs such as GPT-4 for subgoal planning and grounding. This module leverages the power of language models to perform more sophisticated reasoning tasks within the framework. Another noteworthy approach in this regard is \underline{R}easoning vi\underline{a} \underline{P}lanning (RAP)~\citep{hao2023reasoning}, which capitalizes on the language model's dual role as both a world model and a reasoning agent. RAP incorporates a well-founded planning algorithm, specifically based on Monte Carlo Tree Search, to facilitate strategic exploration within the expansive realm of reasoning. The effectiveness of RAP is evaluated across various tasks, including plan generation, mathematical reasoning (e.g., GSM8K~\citep{cobbe2021gsm8k}), and logical reasoning (e.g., PrOntoQA~\citep{SaparovHe2023}). The evaluations demonstrate RAP's proficiency in addressing diverse reasoning challenges, effectively showcasing its versatility as a capable reasoning agent. 
MimicPlay~\citep{wang2023mimicplay} introduces a method for learning robotic policies from human play data, utilizing emergent human and video prompts to direct low-level visuomotor control.

Introspective Reasoning, Extrospective Reasoning, Embodied Reasoning, and Multiagent Reasoning, along with their interconnected aspects, play pivotal roles in the advancement of agent reasoning systems~\citep{qin2023tool}. These components contribute to the development of higher-level cognitive abilities, such as self-awareness, adaptability, and effective collaboration. These capabilities are essential for the creation of intelligent systems that can successfully operate in complex and dynamic environments, seamlessly interact with humans, and engage in cooperative or competitive scenarios with other agents. 
We believe that combining foundational models with classical methods in robotics may create new opportunities, such as integrating classic approaches to perception~\citep{chu2021icm}, mapping~\citep{9123682}, completing~\citep{chu2023diffcomplete}, grasping~\citep{lisimultaneous}, planning~\citep{mao2023agentdriver}, interaction~\citep{jiao2020intuitive}, and control.
Safety is a crucial aspect of embodied intelligent systems. In this context, PlanCP~\citep{sun2023conformal} suggests the application of conformal prediction to diffusion dynamic models.

\subsubsection{Introspective Reasoning}

\begin{figure}[tbp]
	\begin{center}
		
\includegraphics[width=1.0\columnwidth]{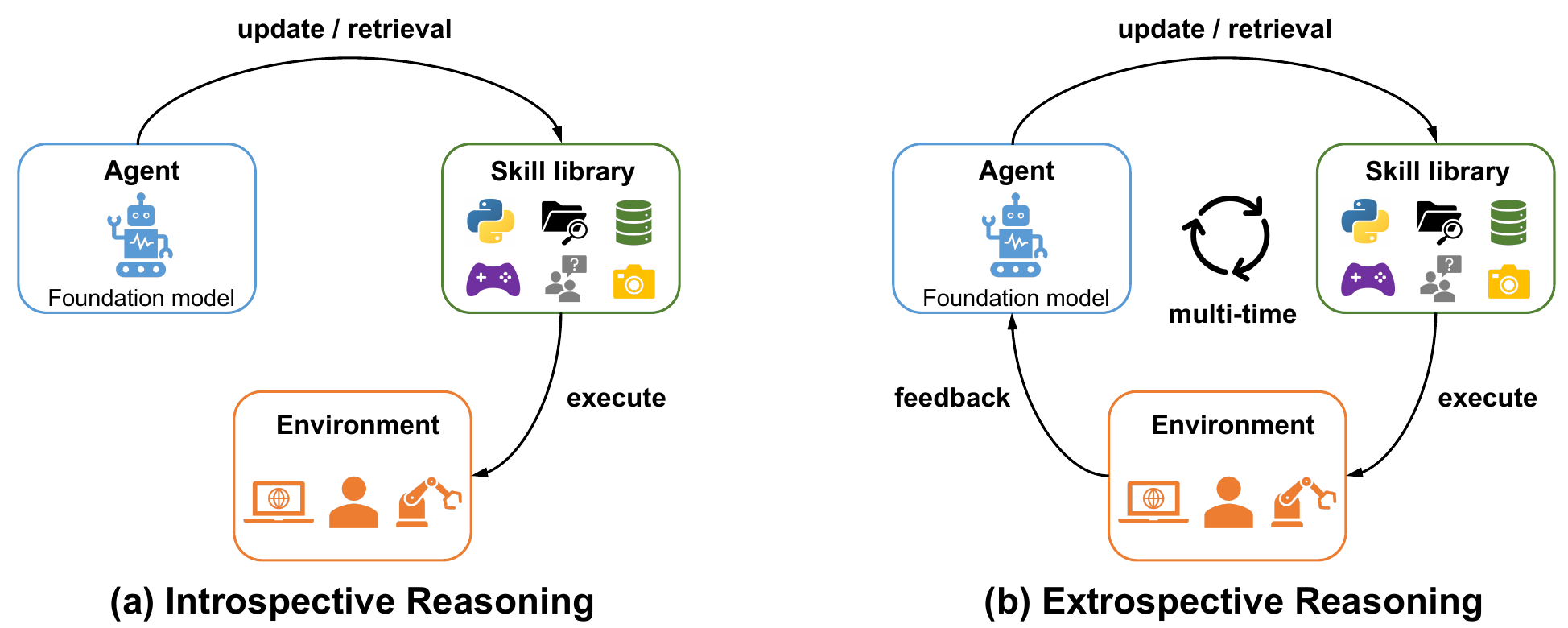}
	\end{center}
       
	\caption{Difference between introspective reasoning and extrospective reasoning. Introspective reasoning does not require interaction with the environment, while extrospective reasoning leverages observation and feedback from the external environment to adapt plans. The figure style credits from work~\citep{qin2023tool}.}
\label{fig:intro_extro}
\end{figure}

Introspective reasoning, illustrated in Figure~\ref{fig:intro_extro}(a), relies solely on internal knowledge and reasoning to generate a static plan of tool use without interacting with the environment~\citep{Leake2012}. Several related works in the field of introspective reasoning with LLMs include Program-Aided Language Models (PAL)~\citep{gao2023pal}, ProgPrompt~\citep{singh2022progprompt}, and Code-as-Policies~\citep{codeaspolicies2022}.

PAL~\citep{gao2023pal} utilizes an LLM for the comprehension of natural language problems and the generation of intermediate reasoning steps in the form of executable programs. Nonetheless, the actual execution of solution steps is delegated to a programmatic runtime, such as a Python interpreter. This approach enables PAL to harness the language understanding capabilities of the LLM while making use of a distinct runtime environment for executing the generated programs.
ProgPrompt~\citep{singh2022progprompt} presents a structured LLM prompt akin to programming, crafted to facilitate the generation of plans in diverse situational settings, encompassing different robot functionalities and tasks. This structure involves prompting the LLM with program-style descriptions of accessible actions and objects in a given environment, along with sample programs for execution. 
Code-as-Policies~\citep{codeaspolicies2022} introduces a robot-oriented framework for Language Model Generated Programs (LMPs). These LMPs are capable of depicting both reactive policies, like impedance controllers, and waypoint-oriented strategies. The versatility of Code-as-Policies is demonstrated across multiple real robot platforms, showcasing its applicability in diverse robotic scenarios.

Introspective reasoning may have limitations in dynamic and uncertain environments where external feedback and interaction with the environment are crucial for effective planning. It may struggle to adapt plans to changing circumstances or handle unexpected events without external information.

\subsubsection{Extrospective Reasoning}
Introspective reasoning, despite its simplicity, lacks the ability to adjust or modify a plan based on intermediate execution results. In contrast, extrospective reasoning operates by generating plans incrementally. As shown in Figure~\ref{fig:intro_extro}(b), it accomplishes this by iteratively interacting with the environment and incorporating feedback obtained from previous executions. Extrospective reasoning actively incorporates external information gathered through interactions with the environment. This allows extrospective reasoning to adapt and refine its plans based on real-time feedback and the observed outcomes of previous actions~\citep{acay2007extrospection}.

By actively engaging with the environment and utilizing feedback, extrospective reasoning offers a more flexible and responsive approach to generating plans, which is particularly suitable for complex and dynamic situations where the ability to adapt and learn from experience is crucial.
Several related works in the field of extrospective reasoning with LLMs include Self-Ask~\citep{press2023measuring}, ReAct~\citep{yao2023react},  ToolFormer~\citep{schick2023toolformer}, and LLM-Planner~\citep{song2023llmplanner}. Self-Ask ~\citep{press2023measuring} proactively generates and responds to its own follow-up queries before addressing the original question. Meanwhile, ReAct~\citep{yao2023react} leverages large language models to concurrently produce reasoning traces and task-specific actions. This dual approach enhances the interaction between these elements, with reasoning traces aiding in the development, monitoring, and modification of action plans, as well as managing unexpected situations. Conversely, actions facilitate the model's engagement with and acquisition of supplementary data from external entities like knowledge bases or environments. ToolFormer~\citep{schick2023toolformer} is designed to intelligently determine the appropriate APIs to utilize, the timing for their invocation, the specific arguments to provide, and how to effectively integrate the obtained results into subsequent token predictions. LLM-Planner~\citep{song2023llmplanner} utilizes the capabilities of large language models for efficient few-shot planning in the context of embodied agents.

In addition to the above-mentioned research, Statler~\citep{yoneda2023statler} provides a framework equipping LLMs with a persistent, memory-like representation of the world state. It utilizes two forms of general LLMs: a world-model reader and a world-model writer, both of which interact with and update the world state. This addition of a memory-like element to the framework significantly boosts the reasoning abilities of LLMs, allowing them to process information over extended time periods, free from the constraints typically imposed by context length limitations.
The explicit representation of the world state empowers LLMs to retain and access relevant information, facilitating more comprehensive and contextually aware reasoning processes.
\citet{dasgupta2022collaborating} propose a collaborative system that combines the complementary reasoning abilities of LLMs. The system has three components: the Planner, the Actor, and the Reporter. The Planner is a pre-trained language model responsible for generating commands that guide the actions of a simple embodied agent, referred to as the Actor. The Reporter acts as a communication bridge between the Planner and the Actor, relaying relevant information to the Planner to inform its decision-making process for issuing subsequent commands. By harnessing the strengths of each component, this collaborative system aims to enhance the overall reasoning and decision-making capabilities of LLMs, allowing for more effective and context-aware interactions between language-based instructions and the embodied agent.
Inner Monologue~\citep{huang2022inner} investigates the capacity of LLMs to reason effectively in embodied contexts by leveraging natural language feedback without additional training. The authors propose that by incorporating environmental feedback, LLMs can develop an inner monologue that augments their capability to process and plan within robotic control scenarios. This development enables LLMs to gain a more comprehensive understanding of the environment and enhances their adaptability to dynamic circumstances. 

The iterative nature of extrospective reasoning enables it to dynamically adjust its plan based on the evolving state of the environment and the outcomes of executed actions. This adaptive process enhances the effectiveness and efficiency of planning, as it leverages the knowledge gained from experience to continually improve future decision-making.

\subsubsection{Embodied Reasoning}
Recent research has highlighted the successful application of LLMs in robotics domains~\citep{Ahn2022DoAI,zeng2023socratic,huang2022inner,codeaspolicies2022,ding2023embodied}. 
Moreover, planning can be considered a form of temporal reasoning, adding to the significance of integrating LLMs into robotics.
Gato~\citep{reed2022generalist} functions as a multimodal, multi-task, and multi-embodiment generalist policy. It leverages supervised learning with an impressive parameter count of 1.2 billion. This technology has been acknowledged as a form of ``general-purpose'' artificial intelligence, representing a significant advancement towards the realization of artificial general intelligence. Robotic Transformer 1 (RT-1)~\citep{brohan2022rt} is trained on a comprehensive real-world robotics dataset consisting of over 130,000 episodes that encompass more than 700 tasks. This extensive dataset was collected over a period of 17 months using a fleet of 13 robots from Everyday Robots. RT-1 demonstrates promising properties as a scalable, pre-trained model, showcasing its ability to generalize based on factors such as data size, model size, and data diversity. The utilization of large-scale data collected from real robots engaged in real-world tasks contributes to RT-1's robustness and its potential for generalization in practical scenarios.
Expanding upon the capabilities of RT-1, Robotic Transformer~2 (RT-2)~\citep{rt22023arxiv} further enhances the model's understanding of the world, resulting in more efficient and accurate execution of robotic tasks. By incorporating the chain of thought reasoning, RT-2 achieves multi-stage semantic reasoning abilities. This expansion equips RT-2 with a set of emerging capabilities derived from extensive training on a vast internet-scale dataset. Prominent advancements encompass a marked improvement in the model's ability to generalize to unfamiliar objects, the capacity to understand commands absent from its original training data, and the capability to engage in basic reasoning when responding to user instructions. These enhancements enhance RT-2's performance and broaden its capacity to tackle a more extensive array of tasks with increased sophistication. 
After that, RT-X~\citep{padalkar2023open} further extends RT-1 and RT-2 to cross-embodiment settings and shows better transferabilities and zero-shot capabilities.
RoboFlamingo~\citep{li2023roboflamingo} leverages pre-trained Vision-Language Models (VLMs) to achieve sophisticated single-step vision-language comprehension. It incorporates an explicit policy head to effectively capture sequential historical data. This design grants it the flexibility needed for implementing open-loop control strategies and is finely tuned for efficient deployment on resource-constrained platforms.

Embodied reasoning plays a vital role in the development of intelligent robots. As humans, we are educated to comprehend the world by employing numerical/physical laws and logical principles. The question arises: can we empower robots with the same capacity? Numerous everyday tasks necessitate simple reasoning based on visual perception and natural language understanding. If we aspire to have robot companions capable of collaborating with us, it is essential for them to possess the ability to understand and reason over both visual information and natural language input. The ultimate objective of creating smart robots is to enable them to act in a manner that is comparable to, or even surpasses, human capabilities~\citep{XU2021100179}. This entails embodying human-like reasoning and performance in robots, aiming to bridge the gap between humans and machines. By enabling robots to understand and reason over visual and linguistic inputs, we move closer to achieving the goal of developing robots that can effectively interact and collaborate with humans.

\subsubsection{Multi-agent Reasoning}

\begin{figure}[tbp]
	\begin{center}
		
\includegraphics[width=1.0\columnwidth]{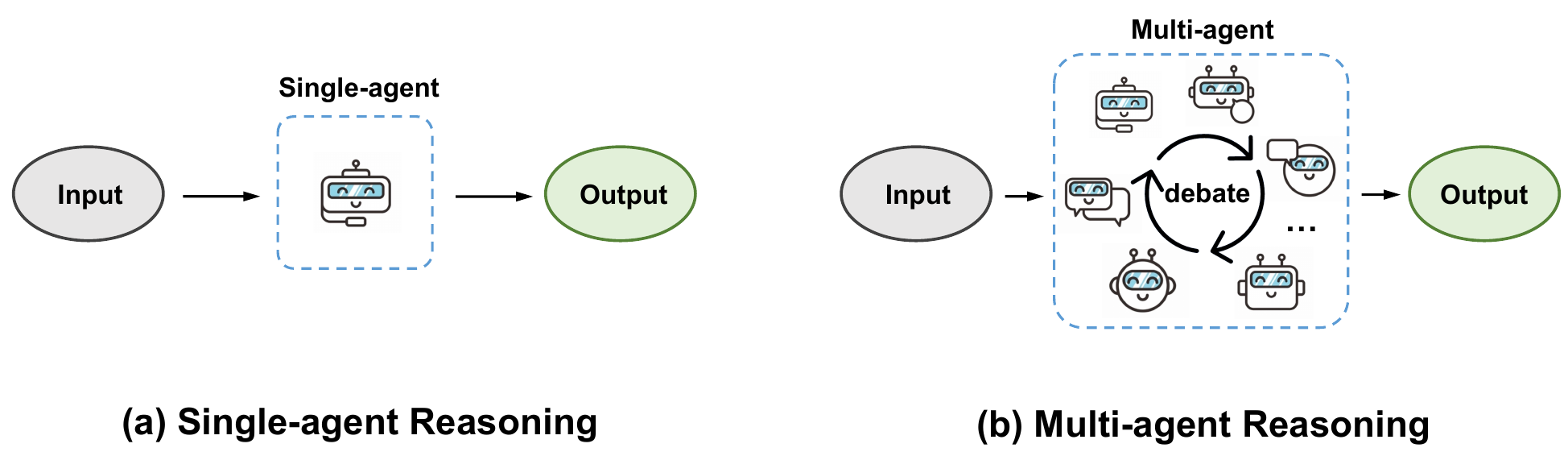}
	\end{center}
        
	\caption{Difference between single-agent reasoning and multi-agent reasoning.}
\label{fig:multi_agent}
\end{figure}

Multi-agent reasoning refers to the cognitive process by which multiple autonomous agents or entities engage in reasoning, decision-making, and communication within a shared environment or context. Compared with reasoning with a single agent, it involves the ability of individual agents to perceive, interpret, and reason about the actions, goals, beliefs, and intentions of other agents, and to adjust their own behaviors accordingly. Their differences are briefly summarized in Figure~\ref{fig:multi_agent}.

Recent studies have introduced the concept of multi-agent debate as a promising method to elevate reasoning abilities and ensure factual accuracy across diverse scenarios. In the work by \citet{zhang2023building}, they introduce a framework that leverages the capabilities of Large Language Models (LLMs) to foster cooperative interactions among multiple agents within embodied environments. This innovative approach empowers embodied agents to efficiently strategize, communicate, and collaborate with both other agents and humans, thereby enhancing their proficiency in accomplishing intricate, long-term tasks.
In a similar vein, \citet{du2023improving} propose a methodology that involves multiple instances of language models engaging in debates. Through iterative rounds of reasoning and response generation, these models collectively work towards reaching a common final answer. This approach has demonstrated significant improvements in mathematical and strategic reasoning across various tasks. 

In contrast to the aforementioned studies,
\citet{nascimento2023selfadaptive} propose the integration of LLMs, such as GPT-based technologies, into multi-agent systems (MASs). They introduce the concept of incorporating LLMs into MASs to create self-adjusting agents. This integration is achieved through an LLM-based MAPE-K (Monitoring, Analyzing, Planning, Executing, and Knowledge) model~\citep{DONASCIMENTO2017161,ibm2004practical}, which enables the agents to adapt and adjust their behaviors based on the knowledge and insights gained from LLMs. 

Federated Learning (FL) has gained prominence as a technology enabling the collaborative development of communal models while safeguarding data that remains decentralized. \citet{chen2023federated} introduce the idea of a federated LLM, encompassing three crucial elements: pre-training of federated LLMs, fine-tuning of these models, and the engineering of prompts specific to federated LLMs.
This approach harnesses the potential of federated learning to enhance multi-agent reasoning by leveraging LLMs. 

These research efforts demonstrate the efficacy of multi-agent debate approaches in enhancing reasoning abilities and factual accuracy. By leveraging the power of large language models and enabling cooperative interactions between agents, these studies contribute to the advancement of AI systems capable of complex reasoning and improved performance across various domains.

\subsubsection{Reasoning in Autonomous Driving}

Reasoning within the domain of autonomous driving spans across perception~\citep{li2023_bev_survey,li2023_toponet,9922340,sun2023connected}, safety~\citep{zhou2023_drivellm_survey}, explainability~\citep{echterhoff2023_concept_gridlock,sha2023languagempc,pmlr-v155-sun21a,pmlr-v155-huang21a} and system level~\citep{chen2023_e2e_survey}.
\citet{chen2023_e2e_survey} propose the frontiers and challenges for end-to-end autonomous driving, where logical reasoning with LLMs could have substantial impacts on different driving scenarios. \citet{zhou2023_drivellm_survey} review some recent work on LLMs with regards to driving. It suggests that by integrating
language data, vehicles as well as transportation systems can carry out reasoning and interact with real-world environments with a higher level of intelligence.

We believe that the common sense and world knowledge inherited from foundation models could unleash the substantial effectiveness of algorithms onboard to handle corner cases and enhance explainability and safety. Below, we survey this emergent topic from two perspectives.

DriveGPT4~\citep{xu2023_drivegpt4} represents a groundbreaking endeavor that harnesses LLMs to comprehend an interpretable end-to-end autonomous driving system. This pioneering effort not only showcases remarkable qualitative but also quantitative achievements when benchmarked against challenging standards. GPT-Driver~\citep{mao2023gptdriver} and Agent Driver~\citep{mao2023agentdriver} introduce the approach by utilizing LLMs as cognitive agents to operate a tool library. This implementation enhances driving behavior by incorporating explainability into the decision-making process.
MotionLM~\citep{seff2023_motionlm} cast multi-agent motion prediction as a language modeling task. The continuous trajectories are represented as sequences of discrete motion tokens.
Among the many other attempts, one particular challenge is how to utilize logical reasoning (e.g., chain of thought) to rationalize and explain driving behaviors. \citet{echterhoff2023_concept_gridlock} propose
a new view using the concept
bottlenecks for control command predictions.
\citet{tan2023_traffic_generation}  turn to language as a source of supervision to obtain dynamic traffic scenarios, surpassing prior work 
in terms of realism and fidelity.
nuPrompt~\citep{wu2023_nuprompt} is the first object-centric language prompt set for 3D, multi-view, and multi-frame driving scenes. It is equipped with diverse pairs of instance-prompt data and validated in the object tracking task.

\subsection{Other Tasks and Applications}~\label{sec:other_reasoning}
\subsubsection{Theory of Mind (ToM)}

The development of Theory of Mind (ToM)-like ability in models is speculated to have occurred naturally and independently as a consequence of their advancing language skills~\citep{kosinski2023theory}. Another explanation suggests that models were able to solve ToM tasks by uncovering and utilizing undiscovered language patterns, rather than explicitly employing ToM. While this alternative explanation may seem ordinary, it is actually remarkable as it implies the existence of undisclosed language regularities that enable the resolution of ToM tasks without the direct engagement of ToM.

\subsubsection{Weather Forecasting}\label{sec:llm_weather}
Weather forecasting plays a crucial role in both scientific research and societal applications. As an application of scientific reasoning, weather forecasting involves the use of reasoning skills to analyze data, identify patterns, and make predictions about future weather conditions. 

MetNet-2~\citep{espeholt2022deep} is a neural network specifically designed for high-resolution precipitation forecasting with up to a 12-hour lead time. This model excels in accurately predicting raw precipitation targets and outperforms state-of-the-art physics-based models currently used in the Continental United States.
In another study, ~\citet{bi2023accurate} present Pangu-Weather, an AI-based approach designed to achieve accurate global weather forecasts in the medium range. This method utilizes 3D deep networks that incorporate earth-specific priors, allowing for the effective handling of complex weather data patterns. To mitigate accumulation errors encountered in medium-range forecasting, a hierarchical temporal aggregation strategy is employed. By undergoing training on an extensive dataset spanning 39 years of global weather information, Pangu-Weather exhibits exceptional deterministic forecasting performance across all assessed variables when compared to the operational integrated forecasting system of the European Centre for Medium-Range Weather Forecasts (ECMWF). This underscores the remarkable effectiveness of Pangu-Weather in delivering precise global weather forecasts, offering valuable insights and advantages for a multitude of applications that heavily depend on weather-related information.

\subsubsection{Medical Reasoning}~\label{subsubsec:medical_reasoning}

Reasoning is also common in medicine. For example, clinicians reason the potential causes of a patient's symptoms and then advise which examinations to take and what treatment is the best following the diagnosis~\citep{qiu2023large}.

With a wide medical knowledge spectrum, foundation models can conduct expert-level reasoning in the context of medicine. For example, Med PaLM 2~\citep{singhal2023towards}, a biomedical large language model (LLM), scored 86.5\% in answering medical questions on the MedQA benchmark; GPT-4 passed the US Medical Licensing Exam (USMLE) with a score of 86.7\%.
Breakthroughs in medical reasoning brought by LLMs also inspire reasoning carried out in other medical modalities, such as medical images. For example, VisionFM~\citep{qiu2023visionfm}, a foundation model for ophthalmic image analysis, demonstrates impressive reasoning skills in predicting the presence of intracranial tumors from fundus photographs, surpassing both intermediate- and senior-level clinicians. RETFound~\citep{zhou2023foundation} shows remarkable performance in reasoning systemic diseases from ocular images.
LLaVA-Med~\citep{li2023llava} adapts LLaVA~\citep{liu2023visual} to align biomedical vocabulary and learn open-ended conversational semantics, which enables the interpretation of biomedical images and achieves promising performance for biomedical visual question answering. ELIXR~\citep{xu2023elixr} incorporates a language-aligned image encoder to perform a range of vision-language reasoning tasks for chest X-ray images. \citet{tu2023towards} develop a multimodal biomedical foundation model, Med-PaLM~M, to simultaneously explore clinical language, imaging, and genomics data, as well as introduce a multimodal biomedical benchmark, MultiMedBench.
Given the multimodal nature of medicine, it is anticipated that medical reasoning will be further augmented by increasingly intelligent multimodal foundation models~\citep{yang2023dawn}.

However, unlike in other domains, reasoning in medicine has to take more caution~\citep{yan2023multimodal}. Rigorous verification and examination should be conducted to ensure the biomedical reasoning outcome is factually grounded, and regulations should be established and enforced to provide legitimate and safe use of foundation models for biomedical reasoning.

\subsubsection{Bioinformatics Reasoning}
\label{subsubsec:bio_reasoning}

Reasoning in bioinformatics involves analyzing and interpreting complex languages of biology and gaining insights into the processes related to life. This includes understanding genetic sequences, protein functions, and cellular mechanisms through the analysis of large-scale datasets. Foundation models are reshaping various perspectives for biological reasoning, such as predicting protein structures and designing sequences in drug discovery~\citep{savage2023drug}.

In the field of biotechnology, numerous studies highlight the efficacy of foundation models in reasoning and analyzing DNA~\citep{nguyen2023hyenadna}, RNA~\citep{wang2023uni}, and protein~\citep{jumper2021highly}.
A notable example is AlphaFold~\citep{jumper2021highly}, which employs a transformer network architecture to precisely predict protein structures.
ProGen~\citep{madani2023large} and its subsequent ProGen2~\citep{nijkamp2022progen2} develop a suite of large protein language models, akin to natural language models, for generating protein sequences.
RFdifusion~\citep{watson2023novo} adopts a denoising diffusion approach in protein structure design, demonstrating significant advancements across various protein design tasks.
In the context of protein-ligand interactions, \citet{li2023druggpt} train the GPT-2 model on protein-ligand binding data, utilizing language model capabilities for ligand design.
Prot2Text~\citep{abdine2023prot2text} combines graph neural networks with LLMs to predict protein functions in a free-text format.
\citet{chen2023towards} introduce a framework powered by LLMs for condition recommendation in chemical synthesis, which aids drug discovery. This framework is designed to search the most recent chemical literature, utilizing in-context learning capabilities and employing multi-LLM debate strategies to enhance effectiveness.
For RNA analysis, Uni-RNA~\citep{wang2023uni} exhibits exceptional performance in structural and functional predictions, including RNA high-order structure map prediction, by leveraging large-scale pre-training on extensive RNA sequences.
Additionally, HyenaDNA~\citep{nguyen2023hyenadna} utilizes the long-range modeling and in-context learning strengths of LLMs and is pre-trained on human reference genome data, yielding significant achievements in genomic tasks.
GeneGPT~\citep{jin2023genegpt} enhances LLMs by integrating the National Center for Biotechnology Information (NCBI) API, which improves answering questions related to genomics.

\subsubsection{Code Generation}
Code generation, also referred to as program synthesis or generating code from a natural language description (NL2Code)~\citep{zan2023large}, is the process or technology that converts inputs in natural language into computer code. NL2Code represents a significant step towards more intuitive and accessible programming, leveraging foundation model to bridge the gap between natural language and computer code. 

PyMT5, delineated in the work of~\citet{clement2020pymt5}, stands as a Python-based text-to-text transfer transformer, adept at translating between diverse combinations of Python method features. This singular model is capable of generating entire methods from natural language documentation strings and summarizing code into various common docstring styles. Similarly, IntelliCode Compose~\citep{svyatkovskiy2020intellicode} is a versatile multilingual code completion tool, proficient in predicting code token sequences and generating syntactically correct code lines. GPT-Neo~\citep{gpt-neo} exemplifies an implementation of GPT-2 and GPT-3-like models, with a focus on distributed support through Mesh Tensorflow. This approach is further extended in GPT-J and GPT-NeoX-20B, as detailed in~\citet{wang2021gpt} and~\citet{black-etal-2022-gpt} respectively.
PLBART~\citep{ahmad2021unified} is a model pre-trained on an extensive corpus of Java and Python functions, coupled with natural language text, utilizing a denoising autoencoding approach. CodeT5~\citep{wang2021codet5} distinguishes itself as a unified pre-trained encoder-decoder Transformer, enhancing the semantic understanding of developer-assigned identifiers.  LaMDA~\citep{thoppilan2022lamda} emerges as a dialog-specialized family of Transformer-based models, pre-trained on a substantial volume of dialog data and web text.

CodeParrot~\citep{tunstall2022natural} is a GPT-2 based model trained for Python code generation, while Codex~\citep{chen2021evaluating} showcases a GPT language model fine-tuned on a vast array of public code from GitHub.~\citet{chandel2022training} delve into the practicality of a Data Science assistant empowered by a transformer model, JuPyT5, trained on public Jupyter Notebook repositories, and introduce a new evaluation metric, DSP. PolyCode~\citep{xu2022systematic} is a GPT-2 based model with substantial coding proficiency across multiple languages, trained on a large code dataset. AlphaCode~\citep{li2022competition} stands out as a code generation system, demonstrating notable performance in programming competitions. CodeRL~\citep{le2022coderl} merges pre-trained language models with reinforcement learning for program synthesis. ERNIE-Code~\citep{chai2022ernie} employs unique pre-training methods, focusing on both monolingual and cross-lingual learning.
Pangu-Coder \citep{christopoulou2022pangu} adopts a two-stage training strategy, initially focusing on raw programming language data and subsequently on text-to-code generation. FIM~\citep{bavarian2022efficient} demonstrates the efficacy of autoregressive language models in text infilling.~\citet{zan2022cert} introduce CERT, a model comprising a sketcher and generator for detailed code creation, trained on unlabelled data. 
InCoder~\citep{fried2022incoder} focuses on code file generation from a large, permissively licensed code corpus, enabling code infilling with bidirectional context. ~\citet{nijkamp2022codegen} present CodeGen, a family of large language models for both natural language and programming, accompanied by the JAXFORMER training library.

CodeGeeX~\citep{zheng2023codegeex} is a multilingual model for code generation, trained on a vast dataset of programming languages. SantaCoder~\citep{allal2023santacoder} is a model with 1.1 billion parameters, trained on Java, JavaScript, and Python subsets from The Stack~\citep{kocetkov2022stack}, and assessed using the MultiPL-E text-to-code benchmark. This research revealed that intensifying the filtering of near-duplicates enhances performance, and interestingly, choosing files from repositories with more than five GitHub stars tends to reduce performance significantly. In contrast, StarCoder~\citep{li2023starcoder} is a more robust model with 15.5 billion parameters and an 8K context length. It boasts infilling capabilities and rapid large-batch inference, enabled by multi-query attention, and is trained on a vast dataset of one trillion tokens from The Stack~\citep{kocetkov2022stack}.
WizardCoder~\citep{luo2023wizardcoder} enhances Code LLMs with intricate instruction fine-tuning, adapting the Evol-Instruct method for the code domain. AceCoder~\citep{li2023acecoder} incorporates two innovative solutions to address coding challenges: firstly, it employs guided code generation, prompting LLMs to initially analyze requirements and produce preliminary outputs like test cases; secondly, it features example retrieval, selecting similar programs as prompt examples to provide relevant content such as algorithms and APIs.
CodeGen2~\citep{nijkamp2023codegen2} aims to make the training of LLMs for program synthesis more efficient by integrating four essential elements: model architectures, learning methods, infill sampling, and data distributions. CodeT5+~\citep{wang2023codet5+} forms a family of encoder-decoder LLMs for code, characterized by flexible module combinations to address a broad spectrum of downstream code tasks. CodeTF~\citep{bui2023codetf} is an open-source Transformer-based library dedicated to cutting-edge Code LLMs and code intelligence applications.
Code Llama~\citep{roziere2023code} represents a family of large language models for code, based on Llama 2, and offers top-tier performance among open models, along with infilling capabilities, support for large input contexts, and the ability to follow instructions in a zero-shot manner for programming tasks. 
CodeFuse~\citep{di2023codefuse} is tailor-made for code-related tasks and is unique in its support for both English and Chinese prompts, accommodating over 40 programming languages. 

\subsubsection{Long-Chain Reasoning}
\label{subsubsec:long-chain-reasoning}
Long-chain reasoning refers to the ability to connect and reason about a series of multiple, often complex pieces of information or events in a long sequential, and extended manner.
Long-chain reasoning is often required in complex problem-solving, decision-making, and understanding of intricate systems.

\citet{ho2022large} introduce Fine-tune-CoT, a method that leverages very large teacher models to generate reasoning samples for fine-tuning smaller models. By employing Fine-tune-CoT, smaller models acquire significant reasoning capabilities, surpassing prompt-based baselines and even outperforming the teacher model in numerous tasks.

Before the emergence of foundational models, the reasoning capabilities of earlier models were notably limited~\citep{sun2022plate}. This limitation primarily stemmed from the tendency of learning-based models to rapidly forget previous information.
Long-chain reasoning has great potential for application in AI Agent Reasoning or Embodied Reasoning, enabling them to handle more intricate and nuanced tasks. 
Despite the emergence of foundational models like GPT-4, mastering long-chain reasoning continues to be a significant challenge. We emphasize the immense utility of long-chain reasoning in applications such as decision-making, planning, and question-answering. With this in mind, we aim to draw attention to this area, encouraging researchers in foundational models to further investigate and advance in this field.

\subsubsection{Abstract Reasoning}
\label{subsubsec:abstract_reasoning}
Abstract reasoning refers to the cognitive ability to analyze and manipulate abstract concepts, ideas, or symbols without relying on specific contexts or concrete examples. It involves transcending immediate sensory input and specific instances to identify underlying patterns, relationships, and fundamental principles. Abstract reasoning requires the identification and application of general patterns based on limited data.

\citet{gendron2023large} extensively evaluate state-of-the-art LLMs in abstract reasoning tasks. Their research reveals that these models demonstrate notably limited performance compared to their performance on other natural language tasks. The findings suggest that LLMs face challenges when it comes to effectively tackling abstract reasoning, highlighting the need for further advancements in this area.

\subsubsection{Defeasible Reasoning}~\label{sec:defeasible_reasoning}

Defeasible reasoning refers to a mode of reasoning in which conclusions can be overturned or revised based on new evidence or information~\citep{madaan2021think}. 
CURIOUS~\citep{madaan2021think} is a framework that supports defeasible reasoning for humans, utilizing an inference graph~\citep{pollock2009recursive}.
In the context of defeasible inference, \citet{rudinger-etal-2020-thinking} have provided three noteworthy datasets: $\delta$-ATOMIC, $\delta$-SNLI, and $\delta$-SOCIAL. These datasets exhibit diversity by covering different domains, offering unique challenges for studying defeasible inference. $\delta$-ATOMIC pertains to commonsense reasoning, presenting scenarios that require drawing defeasible inferences based on background knowledge and understanding of everyday situations.
$\delta$-SNLI focuses on natural language inference, requiring reasoning about the relationships between premises and hypotheses. 
$\delta$-SOCIAL involves reasoning about social norms and conventions, providing a platform for investigating the application of defeasible reasoning in understanding and interpreting social behavior.
\citet{zhou2020kacc} introduce a testbed aimed at evaluating models' abilities to simulate human cognitive processes such as knowledge abstraction, concretization, and completion (KACC). These cognitive abilities play a crucial role in understanding the world and effectively managing acquired knowledge. The testbed includes new datasets characterized by larger concept graphs, ample cross-view links, and dense entity graphs, providing a more comprehensive representation of knowledge.
Within this experimental framework, the authors introduce innovative challenges, specifically multi-hop knowledge abstraction (MKA) and multi-hop knowledge concretization (MKC). These tasks necessitate intricate reasoning capabilities from models, involving the abstraction or concretization of knowledge across multiple sequential steps. 
\citet{kazemi2023boardgameqa} frame the problem of \emph{reasoning with contradictory information}, guided by source preferences, as a classical problem of \emph{defeasible reasoning}. This formulation allows for a comprehensive exploration of models' abilities to handle conflicting information and prioritize different sources in the reasoning process. BoardgameQA~\citep{kazemi2023boardgameqa} is a dataset designed to assess the defeasible reasoning capabilities of models. The dataset consists of 1000 training examples, 500 validation examples, and 1000 testing examples for each variation. 

Each of these datasets presents distinct challenges and opportunities for studying and advancing defeasible inference within various domains. Researchers can leverage these datasets to explore the capabilities and limitations of defeasible reasoning models in different contexts, contributing to the development of robust and adaptable reasoning systems with foundation model technologies.

\newcommand{\tabincell}[2]{\begin{tabular}{@{}#1@{}}#2\end{tabular}}

\subsection{Benchmarks, Datasets, and Metrics}
Benchmarks, datasets, and metrics play a crucial role in evaluating and advancing reasoning capabilities in various domains, driving innovation, and fostering the development of more capable and reliable reasoning systems. These resources provide standardized frameworks and tasks that enable researchers and developers to objectively assess the performance of reasoning models and compare different approaches. 
Representative datasets are summarized in Table~\ref{tab:dataset_summary_1} and~\ref{tab:dataset_summary_2}.

\subsubsection{Commensense Reasoning}
\begin{table*}[]
    \centering
    \begin{tabular}{c|c|c|c}
    \toprule
      Dataset & Choices & Knowledge Types & Questions \\
    \midrule
Swag~\cite{zellers2018swag} & 4 & Temporal, Physical & 113,000 \\
PHYRE~\cite{bakhtin2019phyre} & / & Physical & 25  \\
HellaSwag~\cite{zellers2019hellaswag} & 4 &  Temporal, Physical & 70,000 \\
WinoGrande~\cite{sakaguchi2021winogrande} & 2 & Social, Physical & 44,000 \\
Social IQA~\cite{sap-etal-2019-social} & 3  & Social & 35,350 \\
PIQA~\cite{bisk2020piqa} & 2 & Physical & 21,020 \\
SummEdits~\cite{laban2023llms} & 2 & Social & 6,348 \\
CConS~\cite{kondo-etal-2023-probing}  & / & Physical & 1,112 \\
    \bottomrule
    \end{tabular}
    \caption{Commonsense Reasoning Benchmark Statistics. Choices: the number
of choices for each question; Questions: the
number of questions.}
\label{tab:multiple_choice_benchmarks}
\end{table*}

In addition to CQA~\citep{talmor-etal-2019-commonsenseqa} and CoS-E~\citep{rajani2019explain}, there are several other benchmarks available for evaluating commonsense reasoning (Table~\ref{tab:multiple_choice_benchmarks}):
PHYRE (PHYsical REasoning) benchmark~\citep{bakhtin2019phyre} consists of 25 task templates that focus on physical reasoning.
CConS (Counter-commonsense Contextual Size comparison) dataset~\citep{kondo-etal-2023-probing} investigates the impact of physical commonsense on the contextualized size comparison task. It includes both contexts that align with physical commonsense and those that deviate from it. The dataset comprises 139 templates and automatically generates 1,112 examples.
SummEdits~\citep{laban2023llms} is a benchmark spanning 10 domains. It is designed to be more cost-effective per sample compared to previous benchmarks, offering a 20-fold improvement in efficiency. The benchmark is highly reproducible and aims to evaluate the performance of Language Model-based Systems (LLMs) on complex tasks, addressing issues with existing evaluation benchmarks.

Furthermore, commonsense knowledge encompasses various categories, including physical commonsense, social commonsense, and temporal commonsense. Benchmarks in this domain generally fall into two tasks: multiple-choice evaluation and generative evaluation. Multiple-choice benchmarks, such as SWAG~\citep{zellers2018swag}, HellaSWAG~\citep{zellers2019hellaswag}, Social IQA~\citep{sap-etal-2019-social}, and PIQA~\citep{bisk2020piqa}, require models to select the correct answer from a set of options. Generative evaluation~\citep{lin2020differentiable}, as seen in benchmarks like ProtoQA~\citep{boratko2020protoqa} and CommonGen~\citep{lin-etal-2020-commongen}, involves generating answers based on provided questions and context.
Rainbow~\citep{lourie2021unicorn} is a universal commonsense reasoning benchmark that integrates six existing tasks: 1) $\alpha$NLI~\citep{bhagavatula2019abductive}; 2) Cosmos QA~\citep{huang2019cosmos}; 3) HellaSWAG~\citep{zellers2019hellaswag}; 4) PIQA~\citep{bisk2020piqa}; 5) Social IQA~\citep{sap-etal-2019-social}; and 6) WinoGrande~\citep{sakaguchi2021winogrande}. It covers both social and physical commonsense reasoning and provides a comprehensive evaluation platform.

\paragraph{Metrics}

In multiple-choice benchmarks, accuracy is the primary metric used to evaluate a model's ability to select the correct answer. However, in language generation evaluations, automated metrics like BLEU~\citep{papineni2002bleu} may not always align perfectly with human judgment, so they should be used with caution.

In the case of the PHYRE benchmark~\citep{bakhtin2019phyre}, a metric measuring performance called AUCCESS is computed. AUCCESS aggregates the success percentages across different attempts by using a weighted average. The formula for AUCCESS is AUCCESS = $\sum_k w_k\cdot s_k/\sum_k w_k$. Here, $w_k$ represents weights that place more emphasis on tasks with fewer attempts, and it is calculated as $w_k=\log(k+1) - \log(k)$. The variable $s_k$ denotes the success percentage at the $k$-th attempt.
AUCCESS takes into account the performance across multiple attempts and provides a more comprehensive evaluation that rewards models for solving tasks with fewer attempts. 

\subsubsection{Mathematical Reasoning}

\paragraph{Math Word Problems (MWPs)}
There have been several benchmark datasets introduced for math word problem-solving. One such dataset is Alg514~\citep{kushman-etal-2014-learning}, which is also used by ~\citep{Zhou2015LearnTS} for evaluation. Alg514 consists of 514 algebra word problems sourced from online platforms. Each problem in the dataset is annotated with linear equations, and the template of each problem must appear at least six times within the entire set. Another dataset, Verb395~\citep{hosseini-etal-2014-learning} is a collection of addition and subtraction problems. 
The DRAW dataset~\citep{Upadhyay2015DRAWAC} features 1,000 algebraic word problems, each accompanied by linear equation annotations, collected from \texttt{algebra.com}. Meanwhile, SingleEQ~\citep{koncel-kedziorski-etal-2015-parsing} is comprised of 508 problems,
with each problem corresponding to a single equation. MaWPS~\citep{KoncelKedziorski2016MAWPSAM} repository provides interfaces for adding new word problems, which allows for the further extension of the dataset. These benchmark datasets cover various levels of difficulty and are useful for evaluating math word problem-solving approaches.
Dolphin18K~\citep{huang2016well} consists of over 18,000 annotated math word problems in the field of elementary mathematics. The dataset includes both the unedited text of the problem and either a single or multiple pieces of response text supplied by the individuals who answered the problems. MATH~\citep{hendrycksmath2021} comprises 12,500 challenging competition mathematics problems. Each problem in this dataset is accompanied by a full step-by-step solution. This rich annotated information allows models to be trained to generate detailed answer derivations and explanations. TabMWP~\citep{lu2022dynamic} features a collection of 38,431 grade-level, open-domain problems that necessitate mathematical reasoning through both text and tables. This dataset is divided into training, dev, and testing subsets, following a 6:2:2 distribution. In TabMWP, every query is associated with a tabular context displayed as an image, semi-structured text, and a structured table. The average length of these questions is 22.1 words, with the solutions averaging 49.5 words. The problems in TabMWP can be of two types: free-text questions and multiple-choice questions. 
Every problem comes with annotated gold-standard solutions that illustrate the multi-step reasoning involved. 

GSM8K~\citep{cobbe2021gsm8k} is a math word problem dataset that consists of 8.5K grade school math word problems. These problems exhibit varying levels of linguistic complexity and difficulty. With problem lengths ranging from 2 to 8 steps, they require a diverse set of mathematical skills and strategies to solve effectively.
Multilingual Grade School Math (MGSM) benchmark~\citep{shi2023language} consists of 250 grade-school math problems that have been manually translated from the GSM8K dataset~\citep{cobbe2021gsm8k} into ten languages with diverse linguistic typologies. The MGSM benchmark serves as an evaluation tool to assess the reasoning abilities of language models across multiple languages. It helps to identify areas where models may face challenges, such as cross-lingual reasoning and handling linguistic variations between languages. By incorporating typologically diverse languages, the benchmark ensures its relevance and applicability to real-world multilingual scenarios.

There are two Chinese datasets, Math23K~\citep{wang-etal-2017-deep} and HMWP~\citep{Qin2020SemanticallyAlignedUT}, specifically designed for math word problems at the elementary school level. Math23K~\citep{wang-etal-2017-deep} consists of 23,161 problems that are annotated with structured equations and corresponding answers.
The Hybrid Math Word Problems dataset (HMWP)~\citep{Qin2020SemanticallyAlignedUT} includes three types of math word problems extracted from a Chinese K12 math word problem bank. The dataset comprises 5,491 math word problems, categorized as follows: 2,955 one-unknown-variable linear MWPs, 1,636 two-unknown-variable linear MWPs, and 900 one-unknown-variable non-linear MWPs.
Additionally, there is the DRAW1K dataset~\citep{upadhyay-chang-2017-annotating}, which contains 1,000 general algebra word problems. This dataset includes human-annotated derivations, which serve as information structures for problem-solving. The authors have also provided derivation annotations for over 2,300 algebraic word problems to facilitate future evaluations. They suggest evaluating solvers based on ``derivation accuracy''.
Math23K-F and MAWPS-F~\citep{10.1145/3580305.3599375} are datasets that provide high-quality, precise annotations of formula usage in each reasoning step for Math Word Problems. These datasets aim to enhance the understanding of how formulas are utilized throughout the problem-solving process. In conjunction with these datasets, the authors propose the Formulamastered Solver (FOMAS) system~\citep{10.1145/3580305.3599375}, which incorporates insights from the dual process theory and consists of two components: the Knowledge System and the Reasoning System. The former is responsible for learning and acquiring formula knowledge, while the latter leverages this knowledge to solve math word problems. This dual-component architecture enables FOMAS to leverage formula knowledge in the reasoning process effectively.
The Academia Sinica Diverse MWP Dataset (ASDiv)~\citep{miao-etal-2020-diverse} consists of 2,305 English math word problems (MWPs). The Math Word Problems (MWPs) in this dataset display a diverse range of textual patterns, encompassing the majority of problem types that are typically introduced in elementary education. Additionally, every problem within the collection is meticulously categorized based on its type and educational grade level, providing a clear indication of its difficulty level.

Existing MWP corpora can be categorized into four main groups: (1) Number Word Problem corpora, which contain problems related to numbers exclusively; (2) Arithmetic Word Problem corpora, which involve the four basic arithmetic operations and can be either single-step or multi-step problems; (3) Algebraic Word Problem corpora, which focus on algebraic MWPs; and (4) Mixed-type MWP corpora, which are large-scale collections of MWPs from daily algebra or GRE/GMAT examinations. SVAMP~\citep{patel-etal-2021-nlp}, is a collection of 1,000 math word problems (MWPs), created by introducing variations to initial examples from the ASDiv-A dataset. This compilation features 26 distinct equation models, with each problem incorporating an average of 1.24 operations. Although SVAMP's Corpus Lexicon Diversity (CLD)~\citep{miao-etal-2020-diverse}, falls short when compared to ASDiv-A, it presents a higher level of difficulty. The creators of SVAMP challenge the notion that lexical diversity is a definitive measure of quality in MWP datasets. SVAMP's target audience is students at the elementary school level.

\paragraph{Geometry Problem Solving}

GeoS~\citep{seo-etal-2015-solving} comprises 186 shaded area problems in geometry. This dataset combines text understanding and diagram interpretation. 
In contrast, GeoShader~\citep{alvin2017synthesis} is a smaller dataset containing 102 shaded area problems. These problems are sourced from standard mathematics textbooks from the United States and released exams from the Indian Class X examination. 
Another benchmark, GEOS++~\citep{sachan-etal-2017-textbooks} includes 1,406 questions mirroring the style of SAT exams, covering content from grades 6 through 10. This dataset is segmented into training (350 questions), development (150 questions), and testing (906 questions) subsets, ensuring a balanced representation of questions from each grade level. The authors provide ground-truth logical forms for the 500 annotated questions in the training and development sets. Similarly, GEOS-OS~\citep{sachan-xing-2017-learning} comprises 2,235 geometry problems with demonstrations sourced from a set of grade 6-10 Indian high school math textbooks.
As a numerical reasoning benchmark that incorporates multi-modality, GeoQA~\citep{chen2021geoqa} stands out. GeoQA includes 4,998 geometric problems, each accompanied by annotated programs. Notably, GeoQA surpasses previous benchmarks such as GeoS~\citep{seo-etal-2015-solving} and GEOS++~\citep{sachan-etal-2017-textbooks} in terms of size and diversity.

UniGeo~\citep{chen-etal-2022-unigeo} is a comprehensive and large-scale benchmark for geometry problems. It includes 4,998 calculation problems sourced from GeoQA~\citep{chen2021geoqa}, along with an additional 9,543 proving problems. The proving problems are split into train, validation, and test sets in the proportion of 7.0:1.5:1.5. Each problem is labeled with reasons and mathematical expressions in a way of constituting a multi-step proof. To evaluate model performance, the authors define five sub-tasks: Parallel, Triangle, Quadrangle, Congruent, and Similarity, providing detailed insights into model capabilities. Geometry3K~\citep{lu2021inter}, on the other hand, consists of 3,002 multiple-choice geometry problems with dense annotations in formal language. 
This dataset encompasses a wide range of geometric shapes and objectives, featuring SAT-like problems sourced from two high-school textbooks.

\paragraph{Math Question Answering Datasets}

The AQuA dataset~\citep{ling-etal-2017-program} comprises 100,000 samples of questions, answers, and rationales, with a focus on program induction through rationale generation. Each question in the dataset is divided into four parts: the problem description (the ``question''), the answer ``options'' (in a multiple-choice format), the ``rationale'' description used to arrive at the correct answer, and the label indicating the ``correct option''. 
MathQA~\citep{amini2019mathqa}, an extension of the AQuA dataset~\citep{ling-etal-2017-program}, enhances it by including fully-specified operational programs. MathQA contains 37,000 English multiple-choice math word problems spanning various mathematical domains. The dataset is randomly divided into training, development, and test sets using an 80/12/8\% split ratio.
Specifically tailored for  LLMs, Advanced Reasoning Benchmark (ARB)~\citep{sawada2023arb} is designed to provide more challenging problems in advanced reasoning across multiple fields. ARB includes problems from diverse domains such as mathematics, physics, biology, chemistry, and law. Its purpose is to serve as a benchmark that surpasses the difficulty levels of previous benchmarks, pushing the boundaries of advanced reasoning tasks.

IconQA~\citep{lu2021iconqa}, is an expansive question-answering dataset featuring 107,439 questions. It includes three distinct sub-tasks: choosing from multiple images, selecting from various text options, and completing blank spaces in sentences. The dataset is split into train, validation, and test sets at the proportion of 6:2:2. Icon645~\citep{lu2021iconqa} contains 645,687 colored icons belonging to 377 classes. These icon-based question-answering pairs enable the evaluation of various reasoning skills, including visual reasoning and commonsense reasoning. 
MultiHiertt~\citep{zhao-etal-2022-multihiertt} is a dataset expertly annotated with 10,440 QA pairs, centered on questions and answers pertaining to Multi Hierarchical Tabular and Textual data. This dataset is derived from a variety of financial reports. Documents within MultiHiertt include several tables, mostly hierarchical, accompanied by substantial unstructured text. The complexity and challenge of the reasoning required for each question in MultiHiertt surpass those in existing benchmarks. Detailed annotations of the reasoning steps and supporting facts are included to highlight complex numerical reasoning.

In the realm of Textual QA datasets, DROP (\underline{D}iscrete \underline{R}easoning \underline{O}ver the content of \underline{P}aragraphs)~\citep{dua-etal-2019-drop} comprises 96,567 questions curated from a diverse range of Wikipedia categories, with a particular focus on sports game summaries and historical narratives. It aims to support methods that combine distributed representations with symbolic and discrete reasoning techniques.

Moving on to Tabular QA datasets, WTQ (\underline{W}iki \underline{T}able \underline{Q}uestions) or WikiTableQA~\citep{pasupat-liang-2015-compositional} consists of 22,033 complex question-answer pairs derived from 2,108 HTML tables extracted from Wikipedia. WTQ is tailored to support question-answering tasks specifically focused on semi-structured tables.
Additionally, WikiSQL~\citep{zhong2018seqsql} concentrates on simple SQL queries and single tables. This dataset includes 80,654 meticulously annotated examples of questions and corresponding SQL queries. Spanning 24,241 tables from Wikipedia, WikiSQL stands out for its substantial scale, surpassing comparable datasets in size. Spider~\citep{yu-etal-2018-spider} is tailored for complex, cross-domain semantic parsing and text-to-SQL challenges. This dataset consists of 10,181 questions and 5,693 unique, intricate SQL queries spread across 200 databases. These databases consist of multiple tables covering 138 different domains. Furthermore, \citet{yu-etal-2018-spider} propose a novel task for the text-to-SQL problem using the Spider dataset. AIT-QA (\underline{A}irline \underline{I}ndustry \underline{T}able QA)~\citep{katsis-etal-2022-ait} is tailored for complex and domain-specific Table QA tasks, with a specific focus on the airline industry. The resulting test dataset comprises 515 questions generated from 116 tables. These tables are selected from the 10-K forms of 13 airlines, covering the years between 2017 and 2019.
HiTab~\citep{cheng-etal-2022-hitab} focuses on question-answering (QA) and natural language generation (NLG) tasks specifically tailored for hierarchical tables. This cross-domain dataset is constructed from a rich collection of statistical reports and Wikipedia pages, exhibiting unique characteristics: Firstly, the majority of tables in HiTab are hierarchical, adding complexity to the dataset. Secondly, the questions in HiTab are not generated from scratch by annotators but rather revised from real and meaningful sentences authored by analysts. Lastly, to uncover intricate numerical reasoning in data analysis, fine-grained annotations of quantity and entity alignment are provided. The dataset consists of 3,597 tables, divided into train (70\%), dev (15\%), and test (15\%) sets with no overlap. 

For Hybrid QA Dataset, HybridQA~\citep{chen-etal-2020-hybridqa} is a comprehensive and extensive question-answering dataset designed to challenge reasoning abilities on heterogeneous information sources. This dataset encompasses approximately 70,000 question-answering pairs that are aligned with 13,000 Wikipedia tables. It aims to evaluate the ability to reason and extract information from diverse and varied data sources. Free917~\citep{cai-yates-2013-large} comprises 917 questions sourced from 81 domains within the Freebase database. Freebase is an online, user-contributed, relational database that covers a wide range of knowledge domains. The dataset involves 635 Freebase relations, which have been annotated with lambda calculus forms. However, due to the requirement for logical forms, scaling up the Free917 dataset becomes challenging as it necessitates expertise in annotating logical forms. In contrast, WebQuestions~\citep{berant2013semantic} contains question-answer pairs gathered from non-experts. It presents a higher number of word types compared to datasets like ATIS~\citep{hemphill-etal-1990-atis}, posing greater difficulties in lexical mapping. Nevertheless, WebQuestions exhibits simpler structural complexity, with many questions consisting of a unary, a binary, and an entity. This dataset comprises 5,810 question-answer pairs. The questions were collected using the Google Suggest API, while the answers were curated from Freebase with the assistance of Amazon MTurk. WebQuestionsSP (WebQSP)~\citep{yih-etal-2016-value} is derived from WebQuestions~\citep{berant2013semantic} that includes semantic parses for questions answerable using Freebase. It provides SPARQL queries for 4,737 questions, making it possible to directly execute them on Freebase. WebQSP is larger in size compared to Free917~\citep{cai-yates-2013-large} and offers semantic parses in SPARQL format with standard Freebase entity identifiers. For evaluating reading comprehension (RC) and question-answering (QA) tasks, the dataset WebQComplex or ComplexWebQuestions~\citep{talmor-berant-2018-web} proves valuable. It consists of 34,689 complex examples of broad and intricate questions, accompanied by answers, web snippets, and SPARQL queries. The dataset automatically generates more complex queries involving function composition, conjunctions, superlatives, and comparatives. MetaQA (\underline{M}ovi\underline{E} \underline{T}ext \underline{A}udio QA)~\citep{Zhang2017VariationalRF} is a comprehensive dataset comprising over 400,000 questions designed for both single and multi-hop reasoning. It also provides more realistic versions in text and audio formats. MetaQA expands upon WikiMovies~\citep{miller-etal-2016-key} and serves as a comprehensive extension to it.
These datasets, WebQuestionsSP~\citep{yih-etal-2016-value}, WebQComplex~\citep{talmor-berant-2018-web}, and MetaQA~\citep{Zhang2017VariationalRF}, offer valuable resources for various question answering and reasoning tasks, catering to different complexities and domains.

In the realm of text-centric datasets featuring singular passages, the Stanford Question Answering Dataset (SQuAD)~\citep{rajpurkar-etal-2016-squad} stands out as a significant reading comprehension collection. It contains over 100,000 questions, all crafted by crowdworkers using a range of Wikipedia articles. SQuAD uniquely pairs each question with a specific reading passage, and the answer to each question is a segment extracted directly from that passage. The dataset encompasses a total of 107,785 question-answer pairs across 536 articles. A distinctive feature of SQuAD is its absence of pre-defined answer choices for the questions, unlike some other datasets in this category. Instead, systems are required to select the answer from all possible spans within the passage, posing the challenge of dealing with a relatively large number of candidate answers.

For open-domain text-only datasets, TriviaQA~\citep{joshi-etal-2017-triviaqa} is another reading comprehension dataset that includes over 650,000 question-answer-evidence triples. TriviaQA consists of 95,000 question-answer pairs contributed by trivia enthusiasts. Furthermore, for each question in the dataset, an average of six independent evidence documents are compiled, offering robust distant supervision that enhances the quality of question-answering support. This makes TriviaQA an invaluable tool for assessing the capabilities of systems to understand and respond to questions within an open-domain context.
HotpotQA~\citep{yang-etal-2018-hotpotqa} is a dataset comprising 113,000 question-answer pairs sourced from Wikipedia. It exhibits four key features: The questions in HotpotQA necessitate the identification and reasoning based on various supporting documents to deduce answers. The dataset includes a broad spectrum of questions that are not confined to established knowledge bases or specific knowledge frameworks. HotpotQA offers crucial sentence-level supporting facts necessary for the reasoning process. This level of granularity enables QA systems to reason with robust supervision and provide explanations for their predictions. Additionally, HotpotQA introduces a new type of factoid comparison question. These questions evaluate the ability of QA systems to extract relevant facts and effectively perform necessary comparisons.
By incorporating these four key features, HotpotQA offers a comprehensive and challenging dataset for evaluating QA systems' capabilities in multi-document reasoning, generalization, explanation generation, and factoid comparison. Natural-QA~\citep{kwiatkowski-etal-2019-natural} is a question-answering dataset comprising real anonymized queries that were aggregated from interactions with the Google search engine. Natural-QA includes a total of 307,373 training examples that are publicly available. For the development data, there are 7,830 examples that have been annotated with 5 possible answers. Additionally, the test data consists of another 7,842 examples, also annotated with 5-way annotations.

MultiModalQA (MMQA)~\citep{talmor2021multimodalqa} is an intricate question-answering dataset designed to challenge models in joint reasoning across multiple modalities, including text, tables, and images. The dataset comprises 29,918 questions, and a notable 35.7\% of these questions require cross-modality reasoning.
GeoTSQA~\citep{Li2021TSQATS} is a dataset that focuses on the tabular scenario-based question answering (TSQA) task within the domain of geography. It consists of 556 scenarios accompanied by 1,012 real multiple-choice questions that are contextualized within these tabular scenarios.

TheoremQA~\citep{chen2023theoremqa} is a question-answering dataset that revolves around the concept of theorems. It consists of 800 high-quality questions, which cover 350 theorems spanning various disciplines such as Mathematics, Physics, Electrical Engineering and Computer Science (EE\&CS), and Finance.
TAT-QA~\citep{zhu2021tat} serves as a question-answering benchmark specifically focused on the domain of finance. The dataset evaluates the ability of models to answer questions based on both tables and text. The questions in TAT-QA often require numerical reasoning skills, such as performing arithmetic operations, counting, comparing or sorting values, and combining multiple reasoning steps. The benchmark encompasses 16,552 questions associated with 2,757 hybrid contexts derived from real-world financial reports. It covers a wide range of finance-related topics and scenarios, including stock prices, financial reports, bank transactions, and currency exchange rates. FinQA~\citep{chen2021finqa} is a dataset specifically designed to facilitate numerical reasoning tasks with financial data. The dataset comprises 8,281 question-answer pairs that revolve around financial calculations. Importantly, each pair is accompanied by detailed reasoning steps that provide insights into the process of arriving at the answer.

\paragraph{Metrics}

GeoS++~\citep{sachan-etal-2017-textbooks} employs Normalized Mutual Information (NMI)~\citep{strehl2002cluster} to assess the quality of axiom mention clustering. 
To measure the lexicon usage diversity of a given MWP corpus, \citet{miao-etal-2020-diverse} introduced the use of BLEU~\citep{papineni2002bleu}. They also proposed the Corpus Lexicon Diversity (CLD) metric to assess the lexical diversity of a given corpus~\citep{miao-etal-2020-diverse}.
\citet{cheng-etal-2022-hitab} adopt Execution Accuracy (EA) as their evaluation metric. This approach follows the methodology proposed by \citet{pasupat-liang-2015-compositional}, which measures the percentage of samples with correct answers. 
To evaluate the performance of TSQA models on GeoTSQA, \citet{Li2021TSQATS} employ two standard information retrieval evaluation metrics: Mean Average Precision (MAP) and Mean Reciprocal Rank (MRR). These metrics provide quantitative measures of the models' retrieval effectiveness and ranking accuracy when answering questions based on the tabular scenarios.

\subsubsection{Logical Reasoning}
There are four notable logical reasoning datasets: ProofWriter~\citep{tafjord-etal-2021-proofwriter}, 
PrOntoQA~\citep{SaparovHe2023}, 
FOLIO~\citep{han2022folio}, and LogicalDeduction from BIG-Bench~\citep{srivastava2023beyond}.

ProofWriter~\citep{tafjord-etal-2021-proofwriter} builds upon the original RuleTaker D* datasets~\citep{clark2020transformers} and introduces two additional variants. The closed-world assumption (CWA) variant addresses minor inconsistencies related to negation, while the open-world assumption (OWA) variant incorporates an open-world assumption during reasoning. The RuleTaker D* datasets~\citep{clark2020transformers} consist of five subsets (D0, D1, D2, D3, and D5), each containing 100k questions.
PrOntoQA (\underline{Pr}oof and \underline{Onto}logy-Generated \underline{Q}uestion-\underline{A}nswering)~\citep{SaparovHe2023} contains examples generated from synthetic world models represented in first-order logic.
FOLIO~\citep{han2022folio} is a human-annotated, open-domain dataset that covers a wide range of logical complexities and diversities. It provides first-order logic (FOL) annotations and consists of 1,435 examples. Additionally, the dataset includes 487 sets of premises that serve as rules for deductive reasoning to evaluate the validity of conclusions.
The LogicalDeduction task~\citep{srivastava2023beyond} within BIG-Bench serves as an evaluation benchmark for assessing the ability to perform multi-step logical reasoning. This task involves deducing the order of a sequence of objects based on a minimal set of given conditions. Each instance includes a naturally ordered context with three to seven similar objects, such as differently colored books on a shelf. Alongside the context, a set of simple clues is provided. More relevant datasets and their corresponding statistics are presented in Table~\ref{tab:logical_reasoning_benchmarks}.

These logical reasoning datasets contribute to the development and evaluation of models and systems that aim to enhance logical reasoning capabilities. They provide diverse scenarios and challenges, ranging from synthetic world models to real-world contexts, enabling researchers to explore and advance logical reasoning in various domains.

\begin{table}[]
    \centering
    \begin{tabular}{
    >{\centering}m{0.26\textwidth}
    |>{\centering}m{0.07\textwidth}
    |>{\centering}m{0.06\textwidth}
    |>{\centering}m{0.06\textwidth}
    |>{\centering}m{0.09\textwidth}
    |>{\centering}m{0.09\textwidth}
    |c
    }
    \toprule
    Dataset  & Train Size  & Dev size & Test size & Task Type  &  Synthetic  & Type\\
\midrule
\makecell{$\alpha$NLI \\ \cite{bhagavatula2019abductive}} & 169,654 & - & 1532 & NLI & \XSolidBrush & Abductive \\
\rowcolor{mygray}
\makecell{ProofWriter \\ \cite{tafjord-etal-2021-proofwriter}}  & 69,814 & 10,158 & 20,058 & FV  & \Checkmark & Deductive \\
\makecell{FOLIO \\ \cite{han2022folio}}  &  1,004 &  204 &  227 & FV & \XSolidBrush  & Deductive  \\
\rowcolor{mygray}
\makecell{LogicalDeduction \\ \cite{srivastava2023beyond}} & -  & -  & 1300 & FV & \XSolidBrush  & Deductive \\
\makecell{PrOntoQA \\ \cite{SaparovHe2023}}  & -  &  - & 200 & MCQA & \XSolidBrush  & Deductive  \\
\bottomrule
    \end{tabular}
    \caption{Logical Reasoning Benchmarks~\cite{luo2023logiglue}. There are three types of tasks: multiple choice question answer (MCQA); natural language inference (NLI); and fact verification (FV).}
    \label{tab:logical_reasoning_benchmarks}
\end{table}

\subsubsection{Causal Reasoning}
The T{\"u}bingen cause-effect pairs dataset~\citep{mooij2016distinguishing} encompasses 108 cause-effect pairs obtained from 37 datasets spanning diverse domains, including meteorology, biology, medicine, engineering, and economics. This dataset serves as a benchmark for evaluating causal reasoning abilities. In contrast, the Neuropathic Pain dataset~\citep{tu2019neuropathic} focuses on the relationships between nerves and the corresponding symptoms observed in patients. Due to its specialized medical terminology and domain-specific knowledge, interpreting the variable names within this dataset requires expertise in the field.
The Arctic sea ice dataset~\citep{huang2021benchmarking} presents a graph derived from domain knowledge, featuring 12 variables and 48 edges. It offers valuable insights into the dynamics of Arctic sea ice.

Counterfactual reasoning, even in the absence of actual causality, is a valuable capability for language models. It aids in decision-making, planning, and uncovering hidden insights that may not be immediately apparent in the original context. CRASS (Counterfactual Reasoning Assessment)~\citep{frohberg2021crass} is a benchmark specifically developed to evaluate the proficiency of language models in dealing with counterfactual queries. This benchmark comprises 275 instances in which the language model is presented with counterfactual conditional questions. In each instance, the model is tasked with selecting the most suitable response from a provided set of multiple-choice options.

In evaluating the performance of language models on causal reasoning benchmarks and datasets like CRASS, the commonly used metric is top-$k$ accuracy~\citep{frohberg2021crass}. This metric quantifies the model's ability to make correct predictions by considering the top $k$ ranked choices. It serves as a quantitative measure of the model's proficiency in causal reasoning tasks.
Percentage of Preference~\citep{li2023counterfactual} is a metric used to evaluate logical completions in both counterfactual and factual scenarios. This metric provides a quantitative measure of the extent to which a language model's generated completions align with human preferences and judgments in terms of logical consistency.

\subsubsection{Visual Reasoning}
In order to establish a benchmark for grounded grammar induction, researchers have curated a large-scale dataset known as PARTIT~\citep{hong2021vlgrammar}. This dataset consists of human-written sentences that provide detailed descriptions of part-level semantics for 3D objects.
PTR~\citep{hong2021ptr} is an extensively curated dataset tailored for visual reasoning analysis. It includes around 70,000 synthetic RGB-D images, each accompanied by detailed ground truth data on objects and part-level annotations. These annotations cover a range of aspects such as spatial and geometric relationships, semantic instance segmentation, color attributes, and key physical properties like stability. PTR is designed to facilitate research on part-based conceptual, relational, and physical reasoning. Compositional Language and Elementary Visual Reasoning (CLEVR)~\citep{johnson2017clevr} is a widely used diagnostic benchmark that evaluates a wide array of visual reasoning abilities. It consists of 100,000 rendered images and around one million automatically generated questions, with 853,000 unique questions. CLEVR offers a challenging set of images and questions designed to assess various aspects of visual reasoning, including tasks like counting, comparing, logical reasoning, and memory retention. This dataset provides a robust platform for testing and advancing visual reasoning algorithms and models, enabling researchers to explore and enhance their capabilities in this field. Outside Knowledge Visual Question Answering (OK-VQA)~\citep{okvqa} is a dataset specifically designed for visual question-answering tasks that necessitate the utilization of external knowledge to generate accurate answers. The dataset comprises 14,055 open-ended questions, each associated with five ground truth answers. During the annotation process, the questions were carefully filtered to ensure that they all required external knowledge, such as information from sources like Wikipedia. Additionally, efforts were made to mitigate dataset bias by reducing questions with frequently occurring answers. This dataset serves as a valuable resource for developing and evaluating methods that can effectively leverage external knowledge for visual question-answering tasks.

\subsubsection{Audio Reasoning}

The most widely adopted benchmark datasets for different aspects of audio reasoning, i.e., the Speech processing Universal PERformance Benchmark (SUPERB)~\citep{yang21superb} for discriminative tasks and the enhanced Speech processing Universal PERformance Benchmark (SUPERB-SG)~\citep{tsai2022superb} for generative tasks, have been introduced in Section~\ref{sec:audio_reasoning}. The evaluation metrics for these tasks are listed in Table~\ref{tab:audio_metrics}.

\begin{table*}[]
    \centering
    \renewcommand\tabcolsep{20pt}
    \resizebox{\textwidth}{!}{
    \begin{tabular}{c|c|c}
    \toprule
      Tasks & Cat. & Evaluation Metric \\
    \midrule
        phone recognition & discr. & phone error rate (PER) \\
        automatic speech recognition & discr. & word error rate (WER) \\
        keyword spotting & discr. & accuracy (ACC) \\
        query by example spoken term detection & discr. & maximum term weighted value (MTWV) \\
    \midrule
        speaker identification & discr. & accuracy (ACC) \\
        automatic speaker verification & discr. & equal error rate (EER) \\
        speaker diarization & discr. & diarization error rate (DER) \\
    \midrule
        intent classification & discr. & accuracy (ACC) \\
        slot filling & discr. & F1-score and character error rate (CER) \\
    \midrule
        emotion recognition & discr. & accuracy (ACC) \\
    \midrule
        voice conversion & gen. & mel-cepstrum distortion (MCD) \\
        speech separation & gen. & \tabincell{c}{scale-invariant signal-to-distortion\\ratio improvement (SI-SDRi)} \\
        speech enhancement & gen. & \tabincell{c}{perceptual evaluation of speech quality (PESQ)\\short time objective intelligibility (STOI)} \\
    \bottomrule
    \end{tabular}
}
    \caption{Metrics of Audio Reasoning Tasks. Here ``Cat.'' denotes the category of the tasks. ``discr.'' and ``gen.'' stand for discriminative and generative tasks.}
\label{tab:audio_metrics}
\end{table*}

The availability of datasets in a wide variety of languages contributes to the success of self-supervised learning (SSL) of speech representations, which lays the crucial foundation for audio reasoning. One of the largest and most widely utilized speech corpora for foundation model pre-training is the Libri-light~\citep{kahn2020libri} dataset, which contains approximately 60,000 hours of speech in English originating from audiobooks. Didi Dictation and Didi Callcenter~\citep{jiang2021further} are large-scale corpora in Chinese, each containing roughly 10,000 hours of data collected from mobile dictation applications or phone calls. Apart from English and Chinese, multilingual corpora of substantial sizes are available as well, including VoxPopuli~\citep{babu2021xls} (400,000 hours, 23 languages), Multilingual LibriSpeech~\citep{pratap2020mls} (50,000 hours, 8 languages) and Common Voice~\citep{ardila2020common} (11,000 hours, 76 languages). Since there is typically no ground-truth transcription available, it is impractical to utilize these datasets to train the conventional hidden Markov models (HMMs) and the supervised deep neural network (DNN) or end-to-end (E2E) models. The advance of SSL-based models leverages the power of data and provides a good and universal starting point for further fine-tuning using labeled data for the aforementioned downstream tasks.

\subsubsection{Multimodal Reasoning}

In their comprehensive study, \citet{liu2023hidden} conducted an in-depth evaluation of publicly accessible multimodal models, focusing on their efficacy in a range of text-centric tasks. These tasks include text recognition, encompassing scene text, artistic text, and handwritten text; text-based visual question answering, which involves document text, scene text, and bilingual text; key information extraction from various sources such as receipts, documents, and nutrition facts labels; and the recognition of handwritten mathematical expressions. The study identified both strengths and weaknesses in these models. While they excel in word recognition through semantic understanding, they struggle when it comes to perceiving combinations of characters lacking semantic meaning. Additionally, the models exhibit consistent performance regardless of text length and have limited capabilities in detecting intricate image details. Overall, the study concludes that even the most powerful existing multimodal models fall short compared to domain-specific methods in traditional text tasks. These findings underscore the necessity for innovative strategies to enhance zero-shot multimodal techniques and improve model performance in complex tasks.

Regarding evaluation benchmarks, \citet{vedantam2015cider} contribute to the field by introducing two datasets: PASCAL-50S and ABSTRACT-50S. These datasets are designed for evaluating image caption generation methods. They serve as valuable resources that enable researchers to assess the performance and quality of image captioning models. By utilizing these datasets, researchers can advance image caption generation techniques, fostering progress and innovation in this area of research. LVLM-eHub~\citep{xu2023lvlm} serves as a comprehensive and extensive evaluation benchmark for publicly available large multimodal models. It rigorously assesses the performance of eight LVLMs across six categories of multimodal capabilities. The evaluation process involves the utilization of 47 datasets and 1 arena online platform, providing a robust and standardized framework for evaluating LVLMs. \citet{odouard2022evaluating} present a concept-based approach to systematic evaluations, with a focus on assessing the proficiency of AI systems in utilizing a given concept across different instances. Their evaluation approach, as described in the cited work, entails conducting case studies within two specific domains: RAVEN, which is influenced by the Raven's Progressive Matrices~\citep{raven1938raven}, and the Abstraction and Reasoning Corpus (ARC)~\citep{acquaviva2021larc}. These are frequently utilized for assessing and advancing the capacity for abstraction in AI systems. This methodology provides valuable information about the AI systems' understanding and application of abstract reasoning abilities, shedding light on their ability to grasp and utilize concepts effectively. In a related study, \citet{yin2023lamm} expanded the research on Multimodal Large Language Models (MLLMs) by incorporating point clouds. They introduced the LAMM-Dataset and LAMM-Benchmark, which specifically focus on improving 2D image and 3D point cloud understanding. 

Hallucination is a well-known issue and has long been present in multimodal foundation models as well. Recent studies~\citep{dai-etal-2023-plausible,li2023evaluating} have investigated the performance of Visual Language Pretraining (VLP) models and Vision and Language Models (VLMs) in terms of object hallucination.
\citet{dai-etal-2023-plausible} discovered that despite advancements in VLP models, hallucinations remain a common issue. Interestingly, the study revealed that models with higher scores on conventional metrics like CIDEr~\citep{vedantam2015cider} tended to exhibit more unfaithful results. The authors also found that patch-based features yielded the best results, with smaller patch resolutions reducing object hallucination. To tackle this issue, They proposed a straightforward yet effective VLP loss called Object Masked Language Modeling (ObjMLM)~\citep{dai-etal-2023-plausible}, which further mitigates object hallucination. By decoupling various VLP objectives, the authors demonstrated the importance of token-level image-text alignment and controlled generation in reducing hallucination.
Similarly, ~\citet{li2023evaluating} conducted evaluation experiments on representative VLMs and discovered widespread object hallucination issues. They also found that visual instructions can influence hallucination, with objects that frequently appear in the instructions or co-occur with image objects being more susceptible to hallucination by VLMs. Furthermore, the authors observed that existing evaluation methods may be influenced by the input instructions and generation styles of VLMs. To address this concern, they proposed an improved evaluation method called POPE (Polling-based Object Probing Evaluation) to assess object hallucination more effectively.
\citet{zhao2023evaluating} put forth a methodology for examining the robustness of open-source large VLMs in realistic and high-risk scenarios, where adversaries have limited black-box system access and aim to deceive the model into producing targeted responses. The authors begin by crafting targeted adversarial examples against pre-trained models like CLIP~\citep{radford2021learning} and BLIP~\citep{li2022blip}. They later transfer these adversarial examples to other VLMs, including MiniGPT-4~\citep{zhu2023minigpt4}, LLaVA~\citep{liu2023visual}, UniDiffuser~\citep{bao2023transformer}, BLIP-2~\citep{li2023blip2}, and Img2Prompt~\citep{guo2023images}. The authors discovered that the effectiveness of targeted evasion in large VLMs can be significantly enhanced by employing black-box queries. This approach yields a surprisingly high success rate in generating targeted responses, thereby highlighting the vulnerability of these models to adversarial attacks.
In addition, \citet{huang2023t2i} introduce T2I-CompBench, a comprehensive benchmark for assessing text-to-image generation models' capabilities in processing compositional prompts. It evaluates how these T2I models interpret and represent compositional concepts, such as attribute binding, object relationships, and complex compositions. T2I-CompBench also delves into effectively utilizing multimodal LLMs for evaluation in this context.

Regarding the metrics, DePlot~\citep{liu-2022-deplot} introduces a metric called Relative Number Set Similarity (RNSS) for comparing table similarity. RNSS takes into account the table's structure and numeric values while remaining unaffected by column/row permutations.
CIDEr, as introduced by \citet{vedantam2015cider}, is presented as an automated metric for the evaluation of image captioning. 

In summary, these studies contribute to the development of innovative models, benchmarks, and evaluation metrics in the field of multimodal understanding. They explore various aspects of multimodal reasoning, including visual instruction-following, reasoning on charts, object detection, and image-related tasks. The proposed approaches and evaluations shed light on the strengths and weaknesses of existing models and highlight the need for further advancements in multimodal techniques.

\begin{table*}[]
    \centering
    \renewcommand\tabcolsep{20pt}
    \resizebox{\textwidth}{!}{
    \begin{tabular}{c|c|c}
    \toprule
      Dataset & Tasks & Size \\
    \midrule
Swag~\cite{zellers2018swag} & Commonsense& 113,000 \\
PHYRE~\cite{bakhtin2019phyre} & Commonsense & 25  \\
HellaSwag~\cite{zellers2019hellaswag}&  Commonsense & 70,000 \\
WinoGrande~\cite{sakaguchi2021winogrande} & Commonsense & 44,000 \\
Social IQA~\cite{sap-etal-2019-social} & Commonsense & 35,350 \\
PIQA~\cite{bisk2020piqa} & Commonsense & 21,020 \\
SummEdits~\cite{laban2023llms} & Commonsense & 6,348 \\
CConS~\cite{kondo-etal-2023-probing}  & Commonsense & 1,112 \\
    \midrule
Alg514~\cite{kushman-etal-2014-learning} & Math & 514 \\
Verb395~\cite{hosseini-etal-2014-learning} & Math & 395 \\
Dolphin1878~\cite{Shi2015AutomaticallySN} & Math & 1878 \\
DRAW~\cite{Upadhyay2015DRAWAC} & Math& 1000 \\
SingleEQ~\cite{koncel-kedziorski-etal-2015-parsing} & Math & 508 \\
Dolphin18K~\cite{huang2016well} & Math & 18,000 \\
MATH~\cite{hendrycksmath2021} & Math & 12,500 \\
TabMWP~\cite{lu2022dynamic} & Math & 38,431 \\
GSM8K~\cite{cobbe2021gsm8k} & Math & 8,500 \\
MGSM~\cite{shi2023language} & Math & 250 \\
Math23K~\cite{wang-etal-2017-deep} & Math & 23,161 \\
HMWP~\cite{Qin2020SemanticallyAlignedUT} & Math & 5,491 \\
ASDiv~\cite{miao-etal-2020-diverse} & Math & 2,305 \\
SVAMP~\cite{patel-etal-2021-nlp} & Math & 1,000 \\
GeoS~\cite{seo-etal-2015-solving} & Math & 186 \\
GeoShader~\cite{alvin2017synthesis} & Math & 102 \\
GEOS++~\cite{sachan-etal-2017-textbooks} & Math & 1,406 \\
GEOS-OS~\cite{sachan-xing-2017-learning} & Math & 2,235 \\
GeoQA~\cite{chen2021geoqa} & Math & 4,998 \\
UniGeo~\cite{chen-etal-2022-unigeo} & Math & 4,998 \\
Geometry3K~\cite{lu2021inter} & Math & 3,002 \\
AQuA dataset~\cite{ling-etal-2017-program} & Math & 100,000 \\
MathQA~\cite{amini2019mathqa} & Math & 37,000 \\
ARB~\cite{sawada2023arb} & Math (physics, biology, chemistry, law) & 1,207  \\
IconQA~\cite{lu2021iconqa} & Math & 107,439 \\
MultiHiertt~\cite{zhao-etal-2022-multihiertt} & Math & 10,440 \\
DROP~\cite{dua-etal-2019-drop} & Textual QA & 96,567 \\
WTQ~\cite{pasupat-liang-2015-compositional} & Textual QA & 22,033 \\
WikiSQL~\cite{zhong2018seqsql} & Textual QA & 80,654 \\
Spider~\cite{yu-etal-2018-spider} & Textual QA & 10,181 \\
HybridQA~\cite{chen-etal-2020-hybridqa} & Hybrid QA & $\sim$ 70,000 \\
MetaQA~\cite{Zhang2017VariationalRF} & Hybrid QA & $\sim$ 400,000 \\
SQuAD~\cite{rajpurkar-etal-2016-squad} & Hybrid QA & $\sim$ 100,000 \\
TriviaQA~\cite{joshi-etal-2017-triviaqa}  & Hybrid QA & 95,000 \\
HotpotQA~\cite{yang-etal-2018-hotpotqa} &Hybrid QA & 113,000 \\
MMQA~\cite{talmor2021multimodalqa} &Hybrid QA & 29,918 \\
TheoremQA~\cite{chen2023theoremqa} & Hybrid QA & 800 \\
TAT-QA~\cite{zhu2021tat} & Hybrid QA& 16,552 \\
FinQA~\cite{chen2021finqa}  & Hybrid QA & 8,281 \\
    \bottomrule
    \end{tabular}
    }
    \caption{Summary of Some Reasoning Datasets 1}
\label{tab:dataset_summary_1}
\end{table*}

\begin{table*}[]
    \centering
    \resizebox{\textwidth}{!}{
    \begin{tabular}{c|c|c}
    \toprule
      Dataset & Tasks & Size \\
    \midrule
    APPS~\cite{hendrycksapps2021} & Program Synthesis & 10,000 \\
HumanEval~\cite{chen2021evaluating} & Program Synthesis & 164 \\
MathQA-Python~\cite{austin2021program} & Program Synthesis & 23,914 \\
MBPP~\cite{austin2021program} & Program Synthesis & 974 \\
\midrule 
$\alpha$NLI~\cite{bhagavatula2019abductive} & Logical & 171,186 \\
ProofWriter ~\cite{tafjord-etal-2021-proofwriter} & Logical & 100,030 \\
FOLIO~\cite{han2022folio} & Logical & 1,435 \\
PrOntoQA~\cite{SaparovHe2023}~\cite{srivastava2023beyond} & Logical & 200 \\
LogicalDeduction~\cite{srivastava2023beyond} & Logical & 1,300 \\
\midrule
T{\"u}bingen cause-effect pairs dataset~\cite{mooij2016distinguishing} & Causal & 108 \\
Neuropathic Pain dataset~\cite{tu2019neuropathic} & Causal & N/A \\
Arctic sea ice dataset~\cite{huang2021benchmarking} & Causal & N/A \\
CRASS~\cite{frohberg2021crass} & Causal & 275 \\
\midrule 
PARTIT~\cite{hong2021vlgrammar} & Visual &  $\sim$ 10,000 \\
PTR~\cite{hong2021ptr} & Visual & 70,000 \\
CLEVR~\cite{johnson2017clevr} & Visual & 100,000 \\
OK-VQA~\cite{okvqa} & Visual & 14,055 \\
\midrule
VoxPopuli~\cite{babu2021xls} & Audio & 400,000 hours \\
Libri-light~\cite{kahn2020libri} & Audio & 60,000 hours \\
Multilingual LibriSpeech~\cite{pratap2020mls} & Audio & 50,000 hours \\
Common Voice~\cite{ardila2020common} & Audio & 11,000 hours \\
Didi Dictation~\cite{jiang2021further} & Audio & 10,000 hours \\
Didi Callcenter~\cite{jiang2021further} & Audio & 10,000 hours \\
\midrule
PASCAL-50S~\cite{vedantam2015cider} & Multimodal & 1,000 \\
ABSTRACT-50S~\cite{vedantam2015cider} & Multimodal & 500 \\
LVLM-eHub~\cite{xu2023lvlm} & Multimodal & 47 sub-dataset \\
LAMM-Dataset~\cite{yin2023lamm} & Multimodal & 25 sub-datasets \\
\midrule
RoboTHOR~\cite{deitke2020robothor} & Embodied & / \\
VirtualHome~\cite{puig2018virtualhome} & Embodied & / \\
Gibson~\cite{xiazamirhe2018gibsonenv} & Embodied & / \\
BEHAVIOR-1K~\cite{li2022behaviork} & Embodied & / \\
Habitat~\cite{habitat19iccv} & Embodied & / \\
\midrule
DriveLM~\cite{DriveLM_Contributors_Drive_on_Language_2023} & Driving & 360,000 \\
nuScenes QA~\cite{qian2023_nuscenes_qa} & Driving & 460,000 \\
HAD~\cite{kim2019CVPR_had_honda} & Driving & 
5,675 \\
    \bottomrule
    \end{tabular}
    }
    \caption{Summary of Some Reasoning Datasets 2}
\label{tab:dataset_summary_2}
\end{table*}

\subsubsection{Embodied Reasoning}
RoboTHOR~\citep{deitke2020robothor} is a platform designed to develop and test embodied AI agents in both simulated and physical environments.
VirtualHome~\citep{puig2018virtualhome} is another platform that focuses on modeling complex activities occurring in typical household settings. It offers support for program descriptions that cover a wide variety of activities found in people's homes.
Gibson~\citep{xiazamirhe2018gibsonenv} places emphasis on real-world perception for embodied agents. To bridge the gap between simulation and reality, iGibson~\citep{li2021igibson} and BEHAVIOR-1K~\citep{li2022behaviork} extend the simulation capabilities to encompass a more diverse range of household tasks and achieve high levels of realism. These platforms provide researchers with tools to explore and evaluate embodied AI approaches in realistic simulated environments.
Habitat~\citep{habitat19iccv} boasts high performance, reaching several thousand frames per second (fps) even when running single-threaded. Habitat-Lab~\citep{szot2021habitat} is a high-level, modular library that supports comprehensive development in the realm of embodied AI. It enables the specification of a range of embodied AI tasks, including navigation, interaction, following instructions, and answering questions. This platform allows researchers to tailor embodied agents with particular physical attributes, sensors, and functionalities, and to evaluate their performance on these tasks using established metrics.

These simulation platforms hold great potential for evaluating LLMs on robotics tasks. By leveraging these simulators, researchers can assess the performance and capabilities of LLMs in the context of real-world scenarios, further advancing the field of embodied AI.

\subsubsection{Autonomous Driving}
DriveLM~\citep{DriveLM_Contributors_Drive_on_Language_2023} is a comprehensive driving benchmark to investigate the role of LLMs in various aspects. 
It introduces the reasoning ability of Large Language Models in autonomous driving to guarantee explainable planning and thus make safe decisions.
The questions and answers (QA) in perception, prediction, and planning modules are connected in a graph-style structure, with QA pairs as nodes, and objects' relationships as edges.
Compared to predecessors, such as nuScenes QA~\citep{qian2023_nuscenes_qa}
and HAD~\citep{kim2019CVPR_had_honda}, DriveLM draws many merits from them and improves the logical reasoning in more tasks and a wide diversity of scenarios.

\subsubsection{Code Generation}
Code generation~\citep{pmlr-v155-sun21a} encompasses several datasets and benchmarks that contribute to the advancement of code generation and evaluation. 
The APPS dataset~\citep{hendrycksapps2021}, consists of 10,000 problems derived from coding competitions. It functions as a standard for assessing code generation tasks that are guided by natural language descriptions. This dataset provides a platform for researchers to measure and contrast the effectiveness of models in producing code in response to natural language prompts. Additionally, the study highlights concerns with using BLEU~\citep{papineni2002bleu} as a metric for code generation, suggesting that it may not be reliable in this context. Similarly, HumanEval~\citep{chen2021evaluating} consists of 164 handwritten programming problems and serves as a benchmark for evaluating the performance of Codex~\citep{chen2021evaluating}. Each problem in HumanEval includes a function signature, docstring, body, multiple unit tests, and each problem has 7.7 tests on average. 
MathQA-Python~\citep{austin2021program} is a Python version of the MathQA~\citep{amini2019mathqa} benchmark. It contains 23,914 problems that evaluate models' ability to synthesize code from complex textual descriptions. Notably, the study found that providing natural language feedback from humans resulted in a significant reduction in error rates compared to the models' initial predictions.
The Mostly Basic Programming Problems (MBPP) dataset~\citep{austin2021program} consists of 974 programming tasks specifically designed to be solvable by entry-level programmers. In MBPP, there is a greater emphasis on the usage of imperative control flow structures like loops and conditionals. On the other hand, MathQA-Python~\citep{austin2021program} contains more intricate natural language descriptions, offering a higher level of complexity in the problem statements.

\section{Foundation Model Techniques}
\label{sec:technique}
In this section, we provide a concise overview of various foundation model techniques. Here, we present distinct categories of reasoning techniques:
\begin{itemize}
    \item Pre-Training (Section~\ref{sec: pre-training}): Exploring data and architecture of reasoning foundation models.
    \item Fine-tuning (Section~\ref{sec: fine-tuning}): Focusing on reasoning foundation models' fine-tuning data and techniques.
    \item Alignment Training (Section~\ref{sec: alignment-training}): Examining the alignment techniques employed by reasoning foundation models. 
     \item Mixture-of-Expert (Section~\ref{sec: mixture-of-experts}): Introducing the mixture-of-expert techniques in the context of reasoning.
    \item In-Context Learning (Section~\ref{sec: in-context learning}): Introducing in-context learning in reasoning foundation models.
    \item Autonomous Agent (Section~\ref{sec: llm with agent}): Focusing on the reasoning foundation model as an agent for multiple tasks.
\end{itemize}

\subsection{Pre-Training}
\label{sec: pre-training}
In the pre-training part, LLMs can acquire essential language understanding and generation skills. Here, the data and architecture are critical for the foundation model. Therefore, we will discuss them in the following sections.

\subsubsection{Data Source}
\begin{figure}[tbp]
	\begin{center}
\includegraphics[width=1.0\columnwidth]{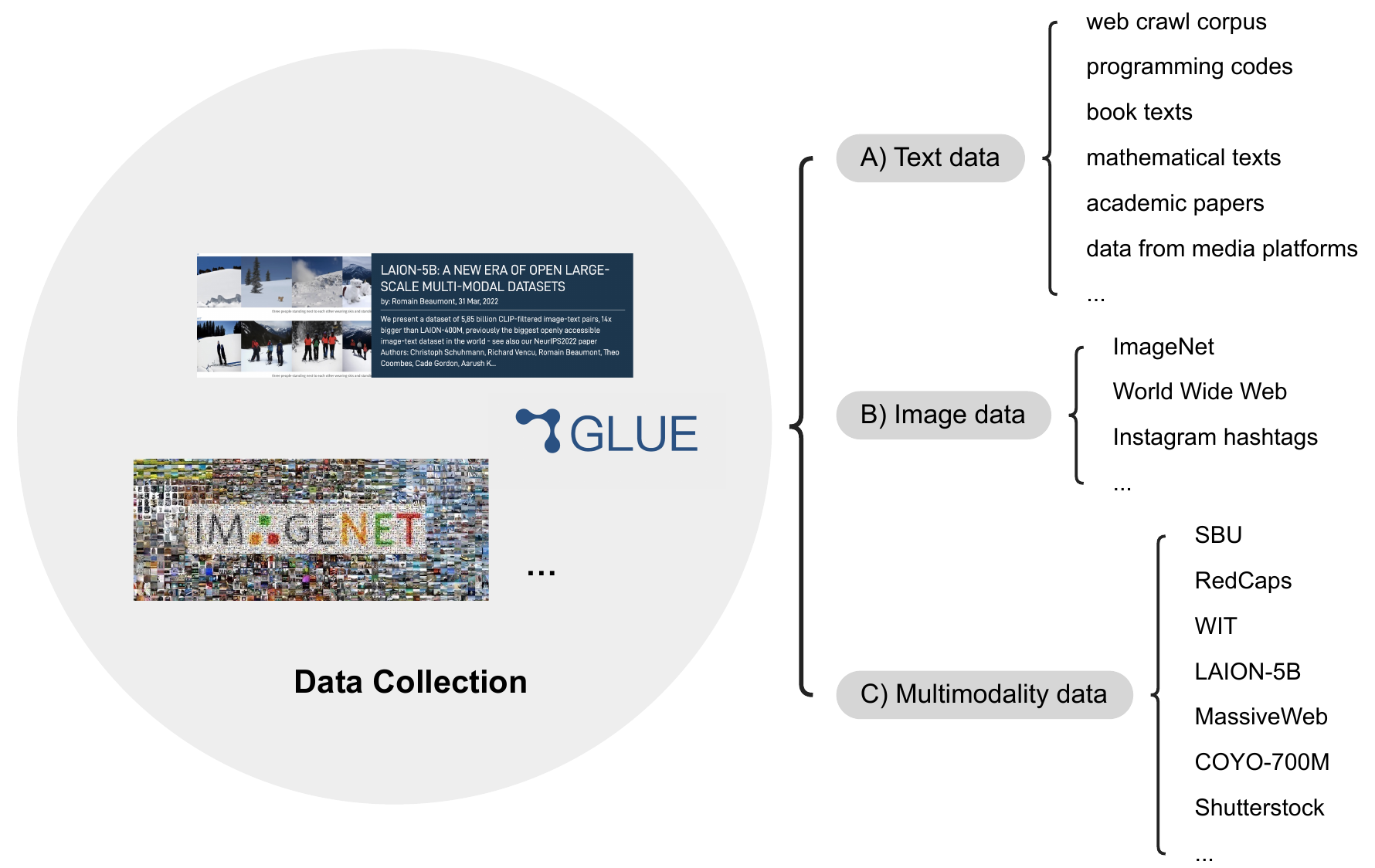}
	\end{center}
        \vspace*{2mm}
	\caption{A diverse suite of data sources and datasets for pre-training foundation models, mainly including text data, image data, and multimodality data.}
\label{fig:pretraining_data}
\end{figure}

Foundation models are data-driven, and both quality and quantity of data lie at the core of foundation model development. Figure~\ref{fig:pretraining_data} presents three broad types of data sources for foundation model pre-training.

\paragraph{Text Data}
The realm of publicly accessible large-scale text datasets has seen a considerable expansion, presenting a rich variety of resources for myriad applications. A prime example is the Pile~\citep{gao2020pile}, an extensive English text corpus, notable for its impressive volume of 825 GB, and specifically curated to facilitate the training of large-scale language models. This corpus is comprised of 22 diverse subsets, recognized for their variety and quality, amalgamating both existing and newly created content, with a significant portion sourced from scholarly and professional domains. A considerable amount of this data is amassed through web crawling initiatives, akin to the CommonCrawl project. It is crucial to acknowledge that such web crawling produces a spectrum of content, from high-caliber material like Wikipedia entries to lower-tier content such as spam emails, necessitating rigorous filtering and processing to elevate data quality.
Another notable dataset in this field is the C4 dataset~\citep{2019t5}, representing an expansive and refined version of the Common Crawl web corpus, extensively utilized in various sectors.
In contrast, the ROOTS dataset~\citep{laurenccon2022bigscience} emerges as an immense resource, encompassing 1.6TB and spanning 59 languages across 46 natural languages, derived from three macro regions and nine language families. It also includes material in 13 programming languages, with Java, PHP, and C++ comprising the majority of its content.
The Gutenberg project~\citep{lahiri:2014:SRW} offers a selection of 3,036 English books by 142 authors. This collection, a subset of the larger Project Gutenberg corpus, has been diligently cleaned to remove metadata, licensing details, and transcribers' notes to the fullest extent.
The CLUECorpus~\citep{xu-etal-2020-clue} stands out in the Chinese text domain as a substantial 100GB resource. This community-led project integrates nine varied tasks, ranging from single-sentence/sentence-pair classifications to machine reading comprehension, all rooted in authentic Chinese text.
Additionally, the Proof-Pile dataset~\citep{azerbayev2023llemma}, with its impressive 8 billion tokens, is notable in the mathematical text sphere. It is distinguished for being among the few open-source language models specifically tuned for the general mathematics field.
The peS2o dataset~\citep{peS2o}, consisting of around 40 million open-access academic papers, is an invaluable asset. It has undergone thorough cleaning, filtering, and formatting, making it ideal for pre-training language models. Originating from the Semantic Scholar Open Research Corpus (S2ORC), it expands the availability of academic text resources.
Furthermore, researchers have access to various public conversation datasets, like the Reddit corpus~\citep{roller2020recipes}. Data from online social media platforms also offers a wealth of conversational content. Scientific text collections typically focus on aggregating materials such as arXiv papers, scientific textbooks, mathematical websites, and related scientific materials. The complex nature of scientific data, often laden with mathematical symbols and protein sequences, requires specialized tokenization and preprocessing methods for standardization and uniform processing by language models.
Recent research~\citep{austin2021program} highlights the benefits of training Large Language Models (LLMs) on extensive code corpora, leading to marked enhancements in generated program quality. These corpora are often sourced from platforms like StackOverflow and GitHub.
Lastly, the RedPajama project~\citep{together2023redpajama} deserves mention for its remarkable feat in reproducing LLaMA's training dataset, encompassing an impressive 1.2 trillion tokens. This dataset includes a vast array of tokens from CommonCrawl, C4, GitHub, Books, ArXiv, Wikipedia, and StackExchange, presenting a comprehensive and diverse resource for the development and refinement of language models.

\paragraph{Image Data}
The methodology of supervised pre-training using extensive, human-curated datasets like ImageNet~\citep{deng2009imagenet} and ImageNet21K~\citep{ridnik2021imagenet} has become a prevalent approach in developing transferable visual representations. This process is structured to create a linkage between an input image and a distinct label, each corresponding to a specific visual concept. With the growing need for large-scale pre-training, the generation of copious amounts of noisy labels from image-text pairings sourced from the World Wide Web has become increasingly relevant.
Leveraging these noisy labels, numerous leading industrial research labs have skillfully assembled vast classification datasets using semi-automatic data pipelines. Notable examples of such endeavors include JFT~\citep{sun2017revisiting} and I2E~\citep{wu2023mofi}. Additionally, they have utilized proprietary data sources, such as Instagram hashtags~\citep{singh2022revisiting}, to enrich their datasets further and augment the precision of their pre-trained models. This strategy has significantly contributed to the advancement of sophisticated visual recognition systems, equipping them with the ability to effectively identify and categorize a wide spectrum of visual concepts and objects.
\paragraph{Multimodality Data} 
The domain of large-scale datasets features several notable examples. SBU~\citep{ordonez2011im2text}, for instance, executes an extensive number of Flickr queries and then rigorously filters the results to produce 1 million images, each paired with a caption that is visually pertinent. Conversely, RedCaps~\citep{desai2021redcaps} is a substantial dataset encompassing 12 million image-text pairs, sourced from Reddit.
The WIT dataset~\citep{srinivasan2021wit} is distinguished by its curated compilation of 37.6 million image-text instances, enhanced with entity information, covering 108 Wikipedia languages, and incorporating 11.5 million unique images. Other relatively large datasets in this field include Shutterstock~\citep{nguyen2022quality}, LAION-400M~\citep{schuhmann2021laion}, and COYO-700M~\citep{byeon2022coyo}. OpenAI's CLIP~\citep{radford2021learning} was refined through an impressive collection of 400 million image-text pairs, meticulously sourced from the web.
Recently, the emergence of datasets at the billion-scale level has been observed. The LAION-5B dataset~\citep{schuhmann2022laion}, for instance, comprises 5.85 billion CLIP-filtered image-text pairs, of which 2.32 billion are in the English language. DataComp~\citep{gadre2023datacomp} functions as a platform for dataset experiments, focusing on a new pool of 12.8 billion image-text pairs collected from Common Crawl. Flamingo~\citep{alayrac2022flamingo} introduces the MultiModal MassiveWeb (M3W) dataset, aggregating text and images from about 43 million web pages, and aligning images with text according to the Document Object Model (DOM).
A noteworthy project in this context is ImageBind~\citep{girdhar2023imagebind}, which aims to develop a joint embedding covering six distinct modalities, including images, text, audio, depth, thermal, and IMU data, with potential extension to other modalities such as point clouds~\citep{guo2023point}. This ambitious endeavor signifies a major step forward in fostering a deeper comprehension of multimodal data by establishing meaningful links across diverse data types. As multimodal learning advances, these developments in dataset creation and application are crucial to the ongoing innovation in the field.

\paragraph{Data for Reasoning}
The significance of code and paper data in enhancing the reasoning abilities of foundation models is paramount. Discussing code data first, research by CoCoGen \citep{madaan-etal-2022-language} indicates that when structured commonsense reasoning tasks are approached as code generation problems, pre-trained language models (LMs) for code exhibit superior reasoning capabilities compared to those trained on natural language. This holds true even for tasks that do not involve source code. Such code data are readily accessible on GitHub and through various filtered datasets available to the public. Highlighting this, StarCoder \citep{li2023starcoder} has released an extensive pretraining dataset (783GB) to further refine LMs' proficiency in coding. In terms of paper data, Galactica \citep{taylor2022galactica} stands out, having been trained on a vast corpus of scientific papers, reference material, knowledge bases, and other diverse sources. This model demonstrates superior performance across a spectrum of scientific tasks compared to existing models. Paper data primarily originates from academic platforms like Arxiv, with a notable emphasis on mathematics papers. Additionally, the peS2o \citep{peS2o} dataset, encompassing over 40 million open-access academic papers from the Semantic Scholar Open Research Corpus (S2ORC), provides a substantial resource for the pretraining of models.
\subsubsection{Network Architecture}
\begin{figure}[tbp]
	\begin{center}	
\includegraphics[width=1.0\columnwidth]{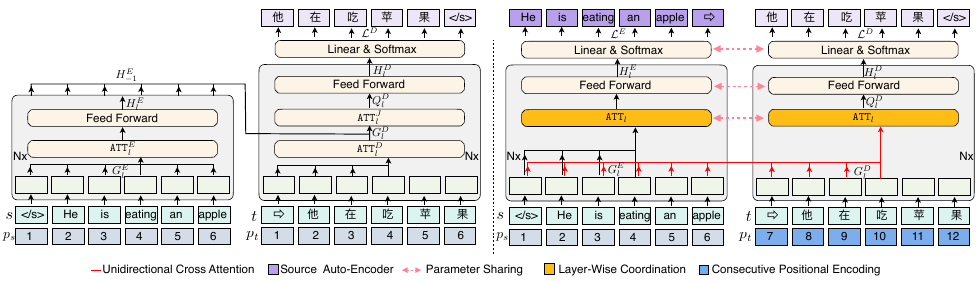}
	\end{center}
        \vspace*{2mm}
	\caption{Illustration of Encoder-Decoder framework (left) and Decoder-Only framework (right). The figure credits from~\citep{fu2023decoder}.}
\label{fig:encoder_decoder}
\end{figure}

The foundation model architecture is essential. We discuss different network architectures in the following and show them in Figure~\ref{fig:encoder_decoder}.

\paragraph{Encoder-decoder Architecture}
The seminal Transformer model, as delineated by \citet{vaswani2017attention}, is founded on the encoder-decoder framework. This paradigm employs dual stacks of Transformer blocks, wherein one functions as the encoder and the other as the decoder. The encoder phase involves utilizing multi-head self-attention layers in a stacked arrangement to decode the intrinsic information within the input sequence, thereby yielding latent representations. In the subsequent phase, the decoder applies cross-attention mechanisms to these representations, facilitating the generation of the target sequence. This innovative architecture is extensively applied in sequence-to-sequence modeling tasks, such as neural machine translation.
In a distinctive approach, BERT~\citep{kenton2019bert} is engineered for the pretraining of deep bidirectional representations from the unlabeled text. It uniquely processes both left and right contexts concurrently across all layers, rendering it exceptionally versatile for a plethora of NLP tasks.
Conversely, BART~\citep{lewis-etal-2020-bart} incorporates a conventional Transformer-based architecture for neural machine translation. While its structure may appear straightforward, BART can be regarded as an evolution of BERT, amalgamating the bidirectional encoder of BERT and the unidirectional, left-to-right decoder of GPT, along with other advanced pretraining methodologies.
Furthermore, Pre-trained Language Models (PLMs) following the encoder-decoder architecture paradigm, exemplified by T5~\citep{2019t5}, have consistently showcased remarkable performance across a wide array of NLP tasks.

\paragraph{Decoder-only Architecture}
The decoder-only architecture is characterized by its strategic use of an attention mask, a pivotal element that ensures each input token is exclusively attentive to preceding tokens, including itself. This unique configuration facilitates a unidirectional flow of information from antecedent tokens to the current token within the decoder, thereby streamlining the processing of input and output tokens. This approach not only simplifies the learning mechanism but also bolsters the model's coherence and consistency.
In the domain of language modeling, the GPT (Generative Pre-trained Transformer) series epitomizes the decoder-only architecture. This series encompasses GPT-1~\citep{radford2018improving}, GPT-2~\citep{radford2019language}, and the notably advanced GPT-3~\citep{brown2020language}. GPT-3, in particular, serves as a quintessential model within this paradigm, exemplifying the architectural efficacy, especially in in-context learning, a distinguishing feature of Large Language Models (LLMs).
The decoder-only architecture's influence transcends the GPT lineage, significantly impacting the broader field of LLMs. Numerous cutting-edge language models have adopted this architectural framework as their foundational structure. For instance, OPT~\citep{zhang2022opt} employs the decoder-only architecture to achieve commendable natural language understanding capabilities. Gopher~\citep{rae2021scaling} also leverages this unidirectional flow to escalate the complexity and scale of language modeling tasks.
Moreover, the decoder-only architecture has been instrumental in the evolution of models like BLOOM~\citep{scao2022bloom}, which utilize its unidirectional information flow for tasks necessitating contextual comprehension. LLaMA~\citep{touvron2023llama} and its successor, LLaMA-2~\citep{touvron2023llama2}, have integrated this architectural style to propel advancements in language modeling, achieving remarkable performances across various NLP benchmarks. GLM~\citep{zeng2022glm} further underscores the decoder-only architecture's efficacy in a range of language understanding tasks, underscoring its vital role in the contemporary landscape of language modeling.

\paragraph{CLIP Variants}
CLIP~\citep{radford2021learning} employs an innovative approach by simultaneously training an image encoder and a text encoder to infer the correct pairings among a set of $<$image, text$>$ pairs. This strategy forms the bedrock of its learning process.
In contrast, FILIP~\citep{yao2021filip} enhances alignment at a finer granularity by incorporating a cross-modal late interaction mechanism. This mechanism employs token-wise maximum similarity measurements between visual and textual tokens to provide guidance for the contrastive objective, resulting in more precise alignments.
FLIP~\citep{li2023scaling} introduces a groundbreaking training technique that involves randomly masking and removing a significant portion of image patches. This approach increases the number of image-text pairs that can be learned within the same wall-clock time, enabling more samples to be contrasted per iteration without significantly increasing memory usage.
On the language encoder side, K-Lite~\citep{shen2022k} suggests incorporating external knowledge in the form of Wiki definitions for entities in combination with their original alt-text for contrastive pre-training. Empirical evidence indicates that enriching text descriptions in this manner leads to improved CLIP performance.
LaCLIP~\citep{fan2023improving} leverages the in-context learning ability of large language models to rewrite text descriptions for their associated images, further enhancing the model's performance by aligning descriptions more effectively with the visual content.  DetCLIP, as introduced in \citet{yao2022detclip}, represents a pioneering approach in parallel visual-concept pre-training for open-world detection. It leverages knowledge enrichment from a meticulously crafted concept dictionary. Meanwhile, its successor, DetCLIPv2~\citep{yao2023detclipv2} capitalizes on the maximum word-region similarity between region proposals and textual words to steer the contrastive objective.

\paragraph{Other Architectures}
Traditional Transformer architectures are often limited by their quadratic computational complexity. To address this, recent research has focused on developing more efficient language modeling architectures. The S4 model~\citep{gu2021efficiently} offers an innovative solution by applying a low-rank correction to condition the state matrix, thus stabilizing its diagonalization and reducing the complexity of the state space model (SSM) to operations akin to a Cauchy kernel.
Similarly, GSS~\citep{mehta2022long} emerges as a compelling alternative to the S4 and DSS~\citep{gupta2022diagonal} models, with the advantage of markedly faster training times. In contrast, H3~\citep{dao2022hungry} is designed to excel in specific functions like recalling earlier tokens in the sequence and comparing tokens across the sequence, further enhancing its efficiency through the integration of FlashCov.
For those exploring subquadratic alternatives to attention mechanisms, Hyenra~\citep{poli2023hyena} offers a notable solution. This model is crafted by combining implicitly parametrized long convolutions with data-controlled gating, significantly diminishing computational requirements.
RWKV~\citep{peng2023rwkv} utilizes a linear attention mechanism, allowing the model to function as either a Transformer or an RNN. This approach not only facilitates parallelized computations during training but also ensures constant computational and memory complexity during inference, marking it as the first non-transformer architecture scalable to tens of billions of parameters.
RetNet~\citep{sun2023retentive} represents another significant contribution, striking an optimal balance between training parallelism, cost-effective inference, and robust performance.
LongNet~\citep{ding2023longnet} introduces dilated attention, a technique that significantly widens the attention field as the distance between tokens increases, thereby enabling effective scaling of sequence length to over a billion tokens.
Lastly, Streaming-LLM~\citep{xiao2023efficient} presents an efficient framework that allows Language Models (LLMs) trained with a finite-length attention window to adapt to infinite sequence lengths without additional fine-tuning. This breakthrough has extended the sequence length capability of these models to 4 million tokens.

\subsection{Fine-Tuning}
\label{sec: fine-tuning}

A fundamental strategy employed by Large Language Models (LLMs) revolves around the concept of pre-training on extensive general domain data, followed by customizing the model to suit particular tasks or domains. This approach endows LLMs with a comprehensive understanding of language patterns, enabling them to subsequently fine-tune their performance across a broad spectrum of downstream tasks, including natural language understanding, generation, and translation. The process of adaptation assumes paramount significance in achieving exceptional results in these specific tasks, as it empowers the LLM to leverage its previously acquired knowledge and apply it to new instances. The adaptation process encompasses a variety of techniques, ranging from thorough fine-tuning of the pre-trained model to the incorporation of task-specific layers or modules, as well as the utilization of transfer learning methods like knowledge distillation.

\subsubsection{Data Source}
\paragraph{Benchmark Data} 
A natural step in the process of data collection entails the adaptation of pre-existing NLP benchmarks. Given that these benchmarks are open-source, researchers find it both more convenient and cost-effective to utilize reasoning benchmarks to bolster the model's reasoning capabilities. However, challenges arise concerning the availability of benchmarks in terms of quantity and scale, and the manual creation of new benchmarks proves to be a resource-intensive task. To tackle this issue, researchers are devising strategies to generate fine-tuning data for reasoning synthesis using an advanced language model.

\paragraph{Synthesis Data} 
This section delves into the synthesis of reasoning data utilizing advanced Large Language Models (LLMs) and subsequently harnesses the generated data for fine-tuning. The core of this research revolves around the application of Chain-of-Thought (CoT) techniques to LLMs, leading to the creation of reasoning paths. Subsequently, the generated data is leveraged for the model fine-tuning~\citep{fu2023specializing,hsieh-etal-2023-distilling,huang2022large,li2022explanations,magister-etal-2023-teaching}.
Additionally, the Finetune-CoT method, as introduced by \citet{ho2022large}, involves the sampling of multiple reasoning paths from LLMs, which are then used for fine-tuning student models with the correct ones. The Distilling step-by-step approach, proposed by \citet{hsieh-etal-2023-distilling}, introduces a novel mechanism with two primary objectives: (a) training smaller models surpassing LLMs and (b) achieving this feat with reduced training data requirements for fine-tuning or distillation. Furthermore, the Self-Improve approach, as detailed in \citet{huang2022large}, includes the selection of rationale-augmented answers with the highest confidence for unlabeled questions using Chain-of-Thought prompting and self-consistency. Subsequently, the LLM is fine-tuned using these self-generated solutions as target outputs, with the additional step of feeding the question and ground-truth label to LLMs to prompt their reasoning path.
An alternative approach involves leveraging several examples with human-written explanations as demonstrations of LLMs, followed by the generation of explanations for the training set~\citep{li2022explanations}. Notably, this research provides evidence supporting the feasibility of fine-tuning a student model based on the chain of thought outputs generated by a larger teacher model, resulting in improved task performance across various types of reasoning datasets, including arithmetic, commonsense, and symbolic reasoning~\citep{magister-etal-2023-teaching}.
In the domain of mathematics, the WizardMath framework~\citep{luo2023wizardmath}, introduces a novel method termed Reinforcement Learning from Evol-Instruct Feedback (RLEIF). This approach initially generates diverse math instruction data using math-specific Evol-Instruct. Subsequently, it involves the training of an instruction reward model (IRM) and a process-supervised reward model (PRM)~\citep{yuan2023scaling,lightman2023lets}. The IRM assesses the quality of the evolved instruction, while the PRM receives feedback for each step in the solution.
Furthermore, MetaMath~\citep{yu2023metamath} introduces an innovative question bootstrapping method (such as forward-backward augmentation ~\citep{jiang2023backward}) to augment the training dataset, resulting in MetaMathQ. This method entails the rewriting of questions with both forward and backward reasoning paths and utilizes LLMs to rephrase the question text. Lastly, MAmmoTH, as introduced by \citet{yue2023mammoth}, presents a new math hybrid instruction-tuning dataset named MathInstruct. This dataset boasts two significant characteristics: extensive coverage of diverse math fields and complexity levels, as well as the incorporation of hybrid Chain-of-Thought (CoT) and Process-of-Thought (PoT) rationales. The paper "Orca" by \citet{mukherjee2023orca} introduces a method called explanation tuning. This approach involves fine-tuning a model using pairs of queries and responses. The responses are augmented with detailed explanations from GPT-4, clarifying the teacher model's reasoning process as it generates each response. In a subsequent work, "Orca2" by \citet{mitra2023orca}, a technique named Prompt Erasing is proposed. This method involves modifying the training process by replacing the specific instructions provided to the student system with generic ones, omitting specific details about how to execute the task.

\subsubsection{Parameter-Efficient Fine-tuning}

\begin{figure}[tbp]
	\begin{center}	
\includegraphics[width=1.0\columnwidth]{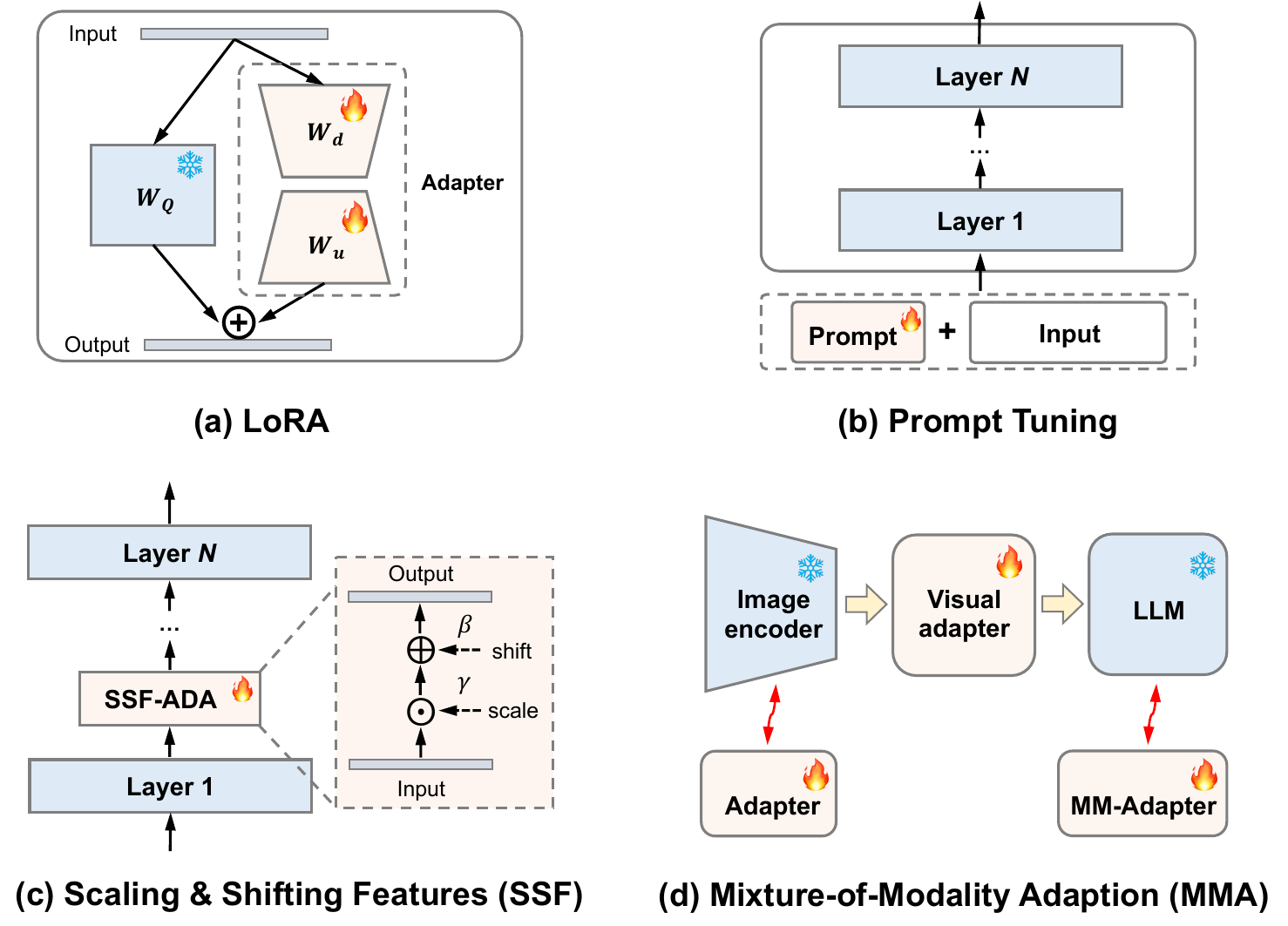}
	\end{center}
	\caption{Illustrations of different parameter-efficient training approaches. (a) Low-Rank Adaptation (LoRA) maintains the original weights of the pre-trained model unchanged, while integrating trainable matrices based on rank decomposition into every layer of the network for adjusting parameters. The figure credits from LoRA~\citep{hu2022lora}. (b) Prompt tuning incorporates trainable prompt vectors at the input layer and uses the prompt-augmented input to tackle specific downstream issues. (c) SSF only needs to scale and shift the deep features extracted by a pre-trained network for parameter-efficient fine-tuning. The figure credits from SSF~\citep{lian2022scaling}. (d)~MMA trains lightweight adapters to bridge the gap between large language models and vision-language tasks to enable the joint optimization of the vision and language models. The figure credits from MMA~\citep{luo2023cheap}.}
\label{fig:42_finetuning}
\end{figure}

One of the fundamental paradigms in building foundation models entails thorough pre-training on general domain data, succeeded by customization for specific tasks or domains. As model sizes continue to grow, conducting comprehensive fine-tuning that alters all model parameters becomes progressively unfeasible. Hence, the importance of parameter-efficient fine-tuning in efficiently refining foundation models cannot be emphasized enough. Some representative approaches of different types of techniques are shown in Figure \ref{fig:42_finetuning}.

\paragraph{Adapter Tuning}
Adapter tuning, a technique that employs specialized neural network modules called ``adapters'' within Transformer models, is discussed in \citet{houlsby2019parameter}. An innovative adaptation method, LLaMA-Adapter~\citep{zhang2023llamaadapter}, has been developed to effectively fine-tune LLaMA models for instruction-following tasks. LLaMA-Adapter showcases its efficiency by introducing only 1.2 million learnable parameters into the pre-trained LLaMA 7B model, utilizing 52,000 self-instruct demonstrations, and completing the fine-tuning process in under an hour using 8 A100 GPUs.
MAD-X~\citep{pfeiffer2020mad} is an adapter-based framework which is designed to learn modular language and task representations that can be adapted to various tasks and languages with high portability and transfer of high parameter-efficiency. On the other hand, AdaMix~\citep{wang2022adamix} fine-tunes a mixture of adaptation modules in each transformer layer while keeping the majority of PLM weights frozen. Compacter~\citep{karimi2021compacter} integrates task-specific weight matrices into the weights of a pre-trained model, which can be efficiently obtained as a sum of Kronecker products between shared ``slow" weights and ``fast" rank-one matrices as defined in each Compacter layer. Lastly, \citet{he2021towards} introduce a unified framework that establishes connections between these approaches.

\paragraph{Low-Rank Adaptation}
Low-Rank Adaptation (LoRA)~\citep{hu2022lora}, shown in Figure~\ref{fig:42_finetuning}(a), offers a distinctive approach aimed at reducing the number of trainable parameters in pre-trained Transformer models when applied to downstream tasks. This technique involves the freezing of pre-trained model weights and the introduction of trainable rank decomposition matrices into each layer of the Transformer architecture. While low-rank decomposition has limitations in terms of representation power, KronA~\citep{edalati2022krona} opts for the Kronecker product as an alternative to low-rank representation. AdaLoRA~\citep{zhang2023adaptive} parametrizes incremental updates through singular value decomposition, allowing for the effective pruning of unimportant singular values.
DyLoRA~\citep{valipour2022dylora} adopts an alternative method, focusing on training LoRA blocks across a spectrum of ranks instead of just one. This is done by organizing the representations acquired by the adapter module at different ranks throughout the training process. For those in search of an efficient fine-tuning solution, ``Efficient Fine-tuning of Quantized LLMs" (QLoRA)~\citep{dettmers2023qlora} provides an appealing option. QLoRA enables the fine-tuning of models with up to 65 billion parameters on a single 48GB GPU, making it a practical and accessible choice for both researchers and practitioners. To enable more complex reasoning tasks, LongLoRA~\citep{longlora} offers a novel method to expand the input context sizes of large language models while maintaining computational efficiency and performance integrity.

\paragraph{Prompt Tuning}
Prefix tuning, initially introduced in \citet{li-liang-2021-prefix}, extends Transformer-based language models by appending a sequence of trainable continuous vectors, known as ``prefixes'' to each layer. It lays the foundation for prompt tuning, a concept akin to ``prefix tuning''~\citep{lester-etal-2021-power}, with a primary focus on integrating trainable prompt vectors exclusively at the input layer. Prompt tuning represents a simple yet highly effective approach for obtaining ``soft prompts'' that empower fine-tuned language models to excel in specific downstream tasks, e.g., classification~\citep{yang2022parameter}, which has been illustrated in Figure~\ref{fig:42_finetuning}(b).
In a similar context, OptiPrompt~\citep{zhong-etal-2021-factual} operates within the continuous embedding space to optimize performance. On the other hand, P-tuning~\citep{liu2023gpt} leverages trainable continuous prompt embeddings alongside discrete prompts, demonstrating effectiveness across both pre-trained and fine-tuned language models, whether in fully supervised or few-shot settings. An evolution of this concept, P-tuning V2~\citep{liu2021p}, proposes the integration of continuous prompts into every layer of the pre-trained model, not restricting itself solely to the input layer. This extension provides a comprehensive approach to harnessing continuous prompts throughout the model's architecture.

\paragraph{Partial Parameter Tuning}
In contrast to the aforementioned approaches that emphasize parameter efficiency, partial parameter tuning distinguishes itself by not introducing any additional components but rather by selectively fine-tuning specific parameters within the original model. Bitfit~\citep{zaken2021bitfit} exemplifies this concept as a method for sparse fine-tuning, concentrating solely on adjusting the bias terms of the model.
Child-Tuning~\citep{xu2021raise} adopts a strategic approach to parameter adaptation. It targets a subset of parameters known as the ``child network" within large pre-trained models while carefully masking out gradients from the non-child network during the backward pass.
In the case of SSF~\citep{lian2022scaling}, corresponding to Figure~\ref{fig:42_finetuning}(c), the method introduces learnable parameters in training. These extra parameters can be seamlessly integrated into the original pre-trained model weights through re-parameterization at inference, with modifications applied to either the complete set or a subset of these parameters.
DiffFit~\citep{xie2023difffit}, on the other hand, presents a parameter-efficient fine-tuning strategy tailored for large pre-trained diffusion models. This method enables rapid adaptation to new domains by fine-tuning bias terms and incorporating newly introduced scaling factors into specific layers of the model.
\citet{fu2023effectiveness} theoretically analyze the parameter sparsity in fine-tuning approaches and design SAM to optimize the selection of suitable parameters.

\paragraph{Mixture-of-Modality Adaption}
\citet{luo2023towards,luo2023cheap} have developed a pioneering method for fine-tuning vision-language models, termed Mixture-of-Modality Adaptation (MMA). Illustrated in Figure~\ref{fig:42_finetuning}(d), MMA serves as a comprehensive optimization framework that unifies the image encoder with Large Language Models (LLMs) via efficient adapters. This work also introduces a cutting-edge routing algorithm in MMA, enabling the model to dynamically modify its reasoning pathways for both single- and multimodal instructions.
Utilizing MMA, the authors have created LaVIN~\citep{luo2023cheap}, a significant vision-language instructed model that exhibits enhanced training efficiency and improved reasoning abilities across a range of instruction-following tasks. LaVIN demonstrates superior performance compared to existing multimodal LLMs. The MMA methodology and LaVIN model hold considerable potential in augmenting the utility of vision-language models, particularly in applied fields like robotics and autonomous systems.
In a similar context, LLaMA-Adapter V2~\citep{gao2023llama} represents a visual instruction model that focuses on parameter efficiency and the seamless integration of visual information. This model incorporates several strategies to boost its performance, including expanding its learnable parameter set, adopting an early fusion approach to integrate visual tokens into the initial layers of LLMs, and applying a joint training approach for both image-text pairings and instruction-following datasets.
Alternatively, LLaVA~\citep{liu2023visual} presents itself as an integrated multimodal model that undergoes an end-to-end training process. LLaVA links a vision encoder and an LLM to process a wide spectrum of tasks involving both vision and language comprehension.
LLaVA-1.5~\citep{liu2023improved} introduces relatively straightforward adjustments, like utilizing CLIP-ViT-L-336px with an MLP projection and integrating task-specific VQA data with basic response formatting prompts. These modifications enable LLaVA-1.5 to set a robust baseline performance, achieving top-tier results across 11 benchmark tasks.

\subsection{Alignment Training}
\label{sec: alignment-training}
The methodology of alignment training introduces an innovative approach that employs learning techniques to optimize language models using human feedback directly. This concept has initiated a new paradigm in which language models are fine-tuned to correspond with intricate human values more closely. While Large Language Models (LMs) can be prompted to execute a variety of natural language processing (NLP) tasks based on given examples, they often manifest unintended behaviors. These include generating fictitious information, creating biased or offensive text, or failing to comply with user directives. Such discrepancies stem from the divergence between the traditional language modeling objective—predicting the next token from the web-based text—and the goal of ``following user instructions in a manner that is both helpful and safe.'' This incongruity suggests a misalignment in the language modeling objective. Rectifying these unintentional behaviors is critically important, especially given the widespread application of language models in numerous domains.

\subsubsection{Data Source}

We define the data as $d_k = (i_k, y_k)$, where $i_k$ represents the instruction and $y_k$ denotes the corresponding response.

\paragraph{Human Data}

Databricks has curated a comprehensive crowd-sourced instruction dataset known as ``databricks-dolly-15k"~\citep{conover2023free}, containing a total of 15,000 instructions. In addition to this, the OpenAssistant corpus~\citep{köpf2023openassistant} consists of more than 10,000 dialogues, involving the participation of over 13,000 international annotators. UnifiedQA~\citep{khashabi2020unifiedqa} has undergone evaluation across 20 diverse datasets, covering various linguistic phenomena. CrossFit~\citep{ye2021crossfit} has been established as an NLP benchmark, encompassing 160 tasks converted from publicly available NLP datasets into a unified text-to-text format. P3~\citep{sanh2021multitask} has collected over 2,000 English prompts from more than 270 datasets, while MetaICL~\citep{min2022metaicl} has conducted experiments across 142 NLP datasets with seven different meta-training and target splits. ExMix~\citep{aribandi2022ext} offers a diverse set of 107 supervised NLP tasks. The Natural Instructions dataset~\citep{naturalinstructions} comprises 61 tasks, and Super-NaturalInstructions~\citep{supernaturalinstructions} expands upon it with over 1.5k tasks. Flan 2022~\citep{longpre2023flan} combines various sources for instruction tuning, adapting templates to achieve strong evaluation performance.
xP3 (Crosslingual Public Pool of Prompts)~\citep{muennighoff2022crosslingual} is a collection of prompts and datasets spanning 46 languages and 16 NLP tasks, which aids multilingual models BLOOMZ and mT0 in zero-shot instruction-following. LongForm~\citep{koksal2023longform} selects 15,000 target text examples from the C4 and English Wikipedia corpus.
Furthermore, ShareGPT, a website, actively encourages users to share their engaging ChatGPT/GPT4 conversations, resulting in a wealth of diverse, human-authored instructions capable of eliciting high-quality ChatGPT/GPT4 responses. To create non-English datasets, the Open Instruction Generalist (COIG)~\citep{zhang2023chinese} translates English instructions into Chinese and utilizes annotators to rectify and reorganize the instructions.

\paragraph{Synthesis Data}

Gathering data from human sources can be a resource-intensive and time-consuming process. Given the remarkable success of Large Language Models (LLMs) like GPT-4, utilizing LLM responses to formulate instructions for training other LLMs in Reinforcement Learning from Human Feedback (RLHF) has become increasingly viable.

Pioneering work in this area, as demonstrated by Self-Instruct~\citep{selfinstruct}, harnesses the in-context learning capability of ChatGPT to generate a substantial volume of instructions. These instructions are drawn from a predefined set of human-annotated examples, spanning a wide range of topics and task types. Building upon this approach, Aplaca~\citep{taori2023alpaca} and its various iterations~\citep{peng2023instruction, vicuna2023} employ LLMs to generate numerous training pairs for RLHF. Instruction Backtranslation~\citep{li2023self} leverages Self-augmentation to create responses along with instructions and utilizes Self-curation to generate instructions based on responses.
Unnatural Instructions~\citep{honovich2022unnatural} stands out as a substantial dataset of innovative instructions, comprising 64,000 examples generated by LLMs through seed examples and rephrasing, resulting in a dataset of approximately 240,000 instances.
The OPT-IML Bench~\citep{iyer2022opt} serves as a benchmark for Instruction Meta-Learning (IML), featuring 2,000 tasks derived from eight existing benchmarks. It evaluates model generalizations using the vanilla GPT-3's Self-Instruct approach, yielding over 52,000 instructions and 82,000 instances.
Koala~\citep{koala_blogpost_2023} is a small yet high-quality dataset curated from various sources, including ChatGPT Distillation Data, resulting in a comprehensive and diverse dataset.
GPT4All~\citep{gpt4all} comprises approximately one million prompt-response pairs from the GPT-3.5-Turbo OpenAI API, spanning the period from March 20, 2023, to March 26, 2023.
Alpaca-GPT4~\citep{peng2023instruction} includes 52,000 examples of instruction-following in both English and Chinese. It incorporates feedback data from GPT-4 to enhance zero-shot performance.
LaMini-LM~\citep{lamini-lm} contains a vast dataset of 2.58 million instruction-response pairs generated by the GPT-3.5-Turbo model. These pairs are drawn from various prompt sources to ensure diversity.
CoEdIT~\citep{raheja2023coedit} is a system that offers an 82,000 dataset of \textless instruction: source, target\textgreater \ pairs for text editing model training and evaluation.
UltraChat~\citep{ding2023enhancing} is an open-source collection of multi-round dialogues, including a million-scale multi-turn instructional conversation data.
CoT-Collection~\citep{kim2023cot} augments Chain-of-Thought (CoT) rationales with 1.88 million instances from the FLAN Collection~\citep{longpre2023flan}.
Dynosaur~\citep{yin2023dynosaur} is a dynamic paradigm for data curation in instruction tuning, continuously expanding by incorporating new datasets from the Huggingface Datasets Platform.

\subsubsection{Training Pipeline}
\begin{figure}[tbp]
	\begin{center}
		
\includegraphics[width=0.9\columnwidth]{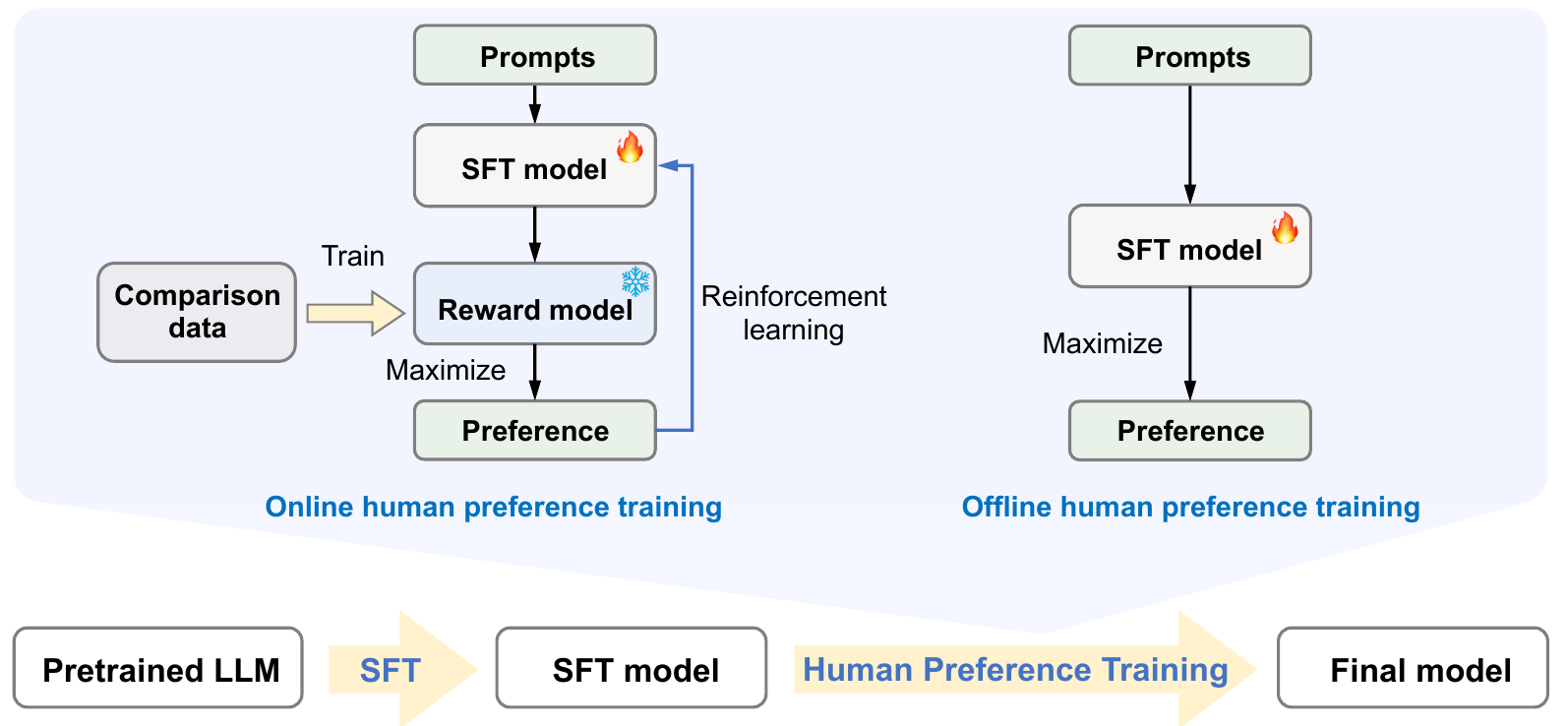}
	\end{center}
        
	\caption{The development process for large language model's (LLM's) alignment training. First, LLM is conventionally optimized via Supervised Fine-Tuning (SFT) using high-quality instruction data. Then, it may be further adjusted through Human Preference Training. The related techniques include online human preference training (left) that needs reinforcement learning and offline ones (right) that directly optimizes the policy to satisfy the preferences best.}
\label{fig:rlhf}
\end{figure}

A common method for enhancing Large Language Models (LLMs) to more accurately interpret and respond to human intentions through specific guidance is known as Supervised Fine-Tuning (SFT). This technique involves processing an instructional input, labeled as $x$, and then calculating the cross-entropy loss in relation to the actual correct response, denoted as $y$. The main role of SFT is to assist LLMs in understanding the deeper meanings within text prompts and to produce appropriate replies. However, a significant drawback of SFT is its lack of capacity to make detailed distinctions between the best and less ideal responses. Overcoming this challenge necessitates additional training strategies, such as incorporating human preference training.The overall training pipeline is presented in Figure~\ref{fig:rlhf}.

\paragraph{Online Human Preference Training}

Reinforcement Learning from Human Feedback (RLHF)\citep{ouyang2022training} represents a strategy developed to interpret human preferences by incorporating additional reward models within the framework of Proximal Policy Optimization (PPO)\citep{schulman2017proximal}. RLHF is divided into three primary phases:
1) The initial stage includes the creation of a comprehensive set of guidelines and the application of Supervised Fine-Tuning (SFT) on pre-existing Large Language Models (LLMs);
2) The next phase involves human evaluators who manually grade pairs of responses, aiding in the development of a reward model that evaluates the effectiveness of the responses generated;
3) Lastly, the SFT model (policy) undergoes refinement through PPO, leveraging the rewards determined by the reward model.

While the PPO framework is known for its effectiveness in learning human preferences, it can present challenges and exhibit less stability during training. An alternative approach, Reward Ranked Fine-Tuning (RAFT)~\citep{dong2023raft}, initially involves sampling a substantial batch of instructions. Subsequently, responses are generated by the current LLMs, and the resulting data is ranked using a reward model. Only the top instances, as determined by the reward model, are then used for SFT.
Additionally, Advantage-Induced Policy Alignment (APA)~\citep{zhu2023fine} employs a squared error loss function based on estimated advantages, offering an alternative perspective on policy alignment within the RLHF framework.

\paragraph{Offline Human Preference Training}

The implementation of those online algorithms can often be challenging due to the intricate interactions required between policy, behavior policy, reward, and value models. This complexity necessitates the adjustment of numerous hyperparameters to strengthen performance. To mitigate this problem, offline learning of human preferences has been studied.

One such approach is Direct Preference Optimization (DPO)~\citep{rafailov2023direct}, which aims to implicitly optimize the same objective as existing Reinforcement Learning from Human Feedback (RLHF) algorithms. Preference Ranking Optimization (PRO)~\citep{song2023preference} takes this further by fine-tuning Large Language Models (LLMs) to better align with human preferences and introduces Supervised Fine-Tuning (SFT) training objectives for regularization. Sequence Likelihood Calibration (SLiC)~\citep{zhao2022calibrating} focuses on adjusting the probability of sequences created by the model to more closely match those of reference sequences within the model's latent space.
In contrast, Rank Responses to align Human Feedback (RRHF)~\citep{yuan2023rrhf} aligns model probabilities of multiple responses with human preferences using ranking loss, providing a simpler yet effective alternative that retains the performance of the Proximal Policy Optimization (PPO) algorithm. \citet{wang2023making} present the Alignment Fine-Tuning (AFT) method, which involves fine-tuning Large Language Models (LLMs) with Chain-of-Thought training data, categorizing generated responses into positive or negative based on correctness, and adjusting response scores using a novel constraint alignment loss.
\subsection{Mixture of Experts (MoE)}
\label{sec: mixture-of-experts}

\begin{figure}[tbp]
	\begin{center}
\includegraphics[width=0.9\columnwidth]{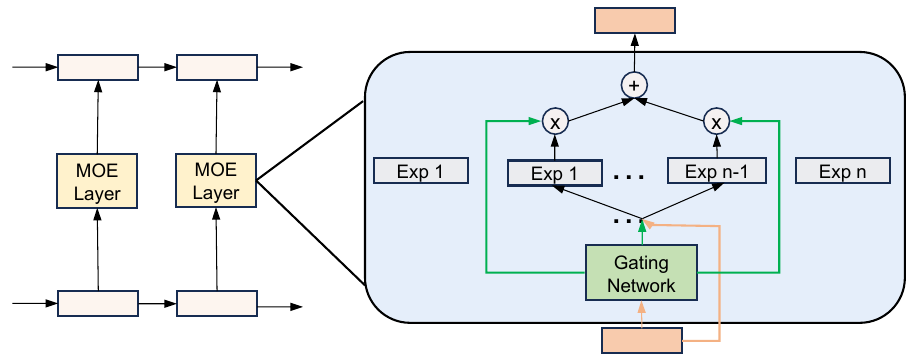}
	\end{center}
	\caption{A Mixture of Experts (MoE) layer within a recurrent language model. In this scenario, we employ a sparse gating function to select a pair of experts for performing the required calculations. The figure credits from~\citep{shazeer2017outrageously}.}
\label{fig:moe}
\end{figure}

As depicted in Figure~\ref{fig:moe}, the Mixture of Experts (MoE) model represents a sophisticated supervised learning framework consisting of an array of networks, each fine-tuned to process a specific segment of the complete training dataset~\citep{jacobs1991adaptive}. In this architecture, individual examples are processed by their respective expert networks. The Sparsely-Gated Mixture-of-Experts model~\citep{shazeer2017outrageously} integrates thousands of feed-forward sub-networks and employs a selective mechanism to engage a sparse array of these experts for each data instance. This methodology culminates in a model with an astounding 137 billion parameters, assigning a singular expert to every example. The model achieves sparsity through a gating function that directs each input to the top-$K$ experts, where $K$ is at least 2.
Expanding upon this concept, GShard~\citep{lepikhin2020gshard} adapts the MoE paradigm for transformers by substituting each feed-forward layer with a pairwise MoE layer, equipped with a Top-2 gating network. In a different approach, Switch Transformers~\citep{fedus2022switch} refine the MoE's sparsity by selecting either the optimal experts or a single best expert (where $K$ equals 1) for each input.
Additionally, GaLM~\citep{du2022glam} leverages a sparsely activated MoE architecture to amplify model capacity while substantially reducing training costs compared to denser models. The largest variant of GaLM boasts a remarkable 1.2 trillion parameters, significantly surpassing GPT-3 in scale. MoE has also been effectively implemented to enhance the capabilities of vision models~\citep{chen2023efficient,chen2023mixed,chen2023mod}.
Moreover, MoE finds application in network compression strategies. WideNet~\citep{xue2022go} represents a parameter-efficient method that utilizes parameter sharing for compression along the network's depth. To optimize modeling capacity, WideNet scales the model's width by replacing standard feed-forward networks with a MoE structure and incorporating distinct layer norms to effectively process diverse semantic representations. MoEBERT~\citep{zuo2022moebert} adopts a similar strategy, transforming the feed-forward neural networks in a pre-trained model into multiple experts. This modification maintains the robust representational abilities of the pre-trained model while integrating layer-wise distillation during training. In inference, a single expert is activated to optimize performance.
\subsection{In-Context Learning}
\begin{figure}[tbp]
	\begin{center}
		
\includegraphics[width=1.0\columnwidth]{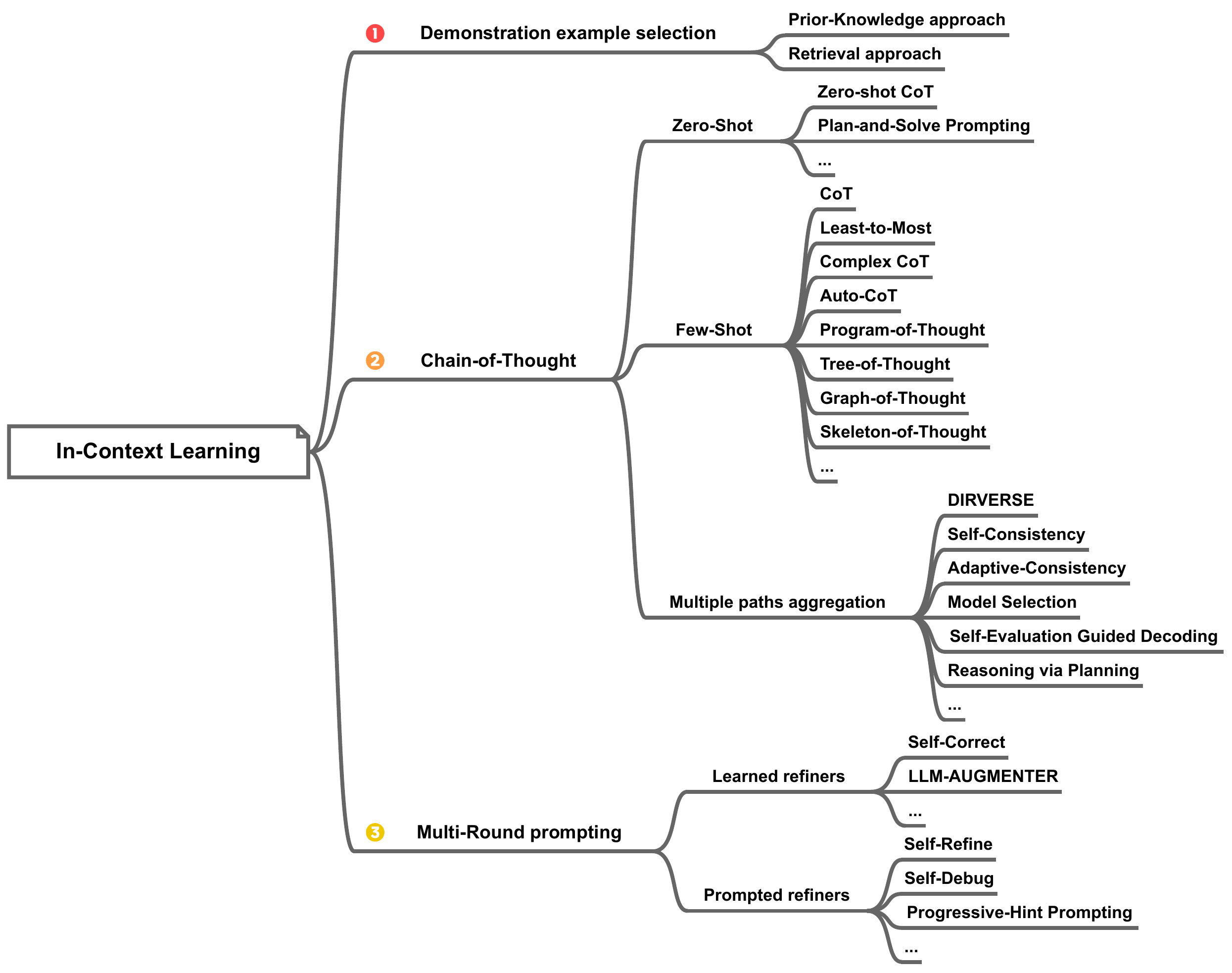}
	\end{center}
        
	\caption{Common techniques for In-Context Learning.}
\label{fig:icl}
\end{figure}

\label{sec: in-context learning}
As described in \citet{brown2020language}, In-Context Learning (ICL) is a method that utilizes a meticulously crafted natural language prompt encompassing both the task description and a subset of task examples to provide demonstrations. This process commences with the task description, followed by the careful selection of a few examples from the task dataset to serve as demonstrations. These chosen instances are then intricately arranged into natural language prompts using thoughtfully designed templates. Subsequently, the test instance is combined with these demonstrations as input for Language Models or Vision-Language Models~\citep{chen2023lightweight} to generate the desired output. Leveraging the provided task demonstrations, LLMs can effectively discern and execute novel tasks without the need for explicit gradient updates.
It is important to note that ICL shares a fundamental connection with instruction tuning, as both methods use natural language to structure tasks or instances.
Nevertheless, instruction tuning necessitates the fine-tuning of LLMs to adapt models, whereas ICL relies purely on prompting LLMs for their applications.
Moreover, it's important to note that instruction tuning can improve the ICL abilities of LLMs for executing specific tasks, particularly in zero-shot situations where only task descriptions are provided~\citep{chung2022scaling}.
A diverse set of common techniques are introduced next and listed in Figure~\ref{fig:icl}.

\subsubsection{Demonstration Example Selection}

The effectiveness of In-Context Learning (ICL) often exhibits considerable variability based on the choice of demonstration examples. Therefore, it becomes crucial to carefully select a subset of examples that can truly harness the ICL capacity of Language Models (LLMs). Two primary methods for demonstration selection are prevalent: heuristic approaches and LLM-based approaches, as explored in the works of \citet{liu2022makes} and \citet{lee2022does}.

\textbf{Prior-Knowledge Approach} 
Due to their cost-effectiveness and simplicity, heuristic techniques have been widely adopted in previous research for the selection of demonstrations. Many studies have integrated k-NN-based retrievers to identify semantically relevant examples for specific queries, as evidenced by \citet{liu2022makes} and \citet{lee2022does}. However, it is important to note that these approaches typically operate on a per-example basis, lacking a holistic evaluation of the entire example set.
To overcome this limitation, diversity-centric selection strategies have been introduced to curate a subset of examples that collectively represent the spectrum of specific tasks, as explored in the works of \citet{levy2022diverse} and \citet{hongjin2022selective}. Moreover, research conducted by \citet{ye2022complementary} takes into account both relevance and diversity in the demonstration selection process.
Intriguingly, Complex CoT~\citep{fu2022complexity} advocates the inclusion of intricate examples that involve extensive reasoning steps, while Auto-CoT~\citep{zhang2022automatic} suggests the sampling of a more diverse set of examples for demonstration.

\textbf{Retrieval Approach}
Another area of research is dedicated to harnessing the capabilities of Language Models (LLMs) in selecting demonstrations. For instance, LLMs can be employed to directly assess the informativeness of each example by quantifying the performance improvement resulting from its inclusion, as demonstrated by \citet{li2023finding}.
In a related vein, \citet{rubin2022learning} introduce an approach called EPR, which involves a two-stage retrieval process. Initially, EPR recalls similar examples through an unsupervised method and subsequently ranks them using a dense retriever. Building upon this, Dr.ICL~\citep{luo2023dr} applies the EPR approach to a broader spectrum of evaluation tasks, encompassing QA, NLI, MathR, and BC.
Within the context of in-context learning, Compositional Exemplars for In-context Learning (CEIL)~\citep{ye2023compositional} utilizes Determinantal Point Processes (DPPs) to learn the interaction between input and in-context examples. This model is optimized using a well-crafted contrastive learning objective.
Additionally, LLM-R~\citep{wang2023learning} adopts a ranking method for retrieved candidates, relying on the conditional LLM log probabilities of the ground-truth outputs. It employs a cross-encoder-based reward model for capturing fine-grained ranking signals from LLMs, and a bi-encoder-based dense retriever trained through knowledge distillation. The Unified Demonstration Retriever (UDR)~\citep{li-etal-2023-unified} utilizes a shared demonstration retrieval model to overcome the issue of non-transferability among retrievers across different tasks. UDR ranks candidate examples based on LLM's feedback.
With trained retrievers, DQ-LoRe~\citep{Xiong2023DQLoReDQ} utilize Dual Queries and Low-rank approximation Re-ranking to automatically select exemplars for in-context learning.

\subsubsection{Chain-of-Thought}

\paragraph{Zero-Shot CoT}
Zero-shot CoT~\citep{kojima2022large} introduces a novel approach to enhance model reasoning abilities by incorporating additional sentences. For instance, empirical evidence has demonstrated that including the phrase ``Let's think step by step" can significantly boost the model's reasoning skills.
In a similar vein, Plan-and-Solve (PS) Prompting~\citep{wang2023plan} presents a two-fold strategy. First, it involves formulating a plan to break down the overall task into smaller, manageable subtasks. Subsequently, these subtasks are executed according to the devised plan. More precisely, PS prompting replaces the original ``Let’s think step by step" from Zero-shot CoT with a new prompt that encourages a more detailed approach: ``Let’s first understand the problem and devise a plan to solve it. Then, let’s proceed to execute the plan and solve the problem step by step."

\paragraph{Few-Shot CoT}
Chain-of-Thought (CoT)~\citep{wei2022chain} has charted a significant course for enhancing the reasoning capabilities of Language Models (LLMs) by employing detailed reasoning paths as prompts. This directional trend has given rise to various CoT variants, such as least-to-most~\citep{zhou2022least}, complex CoT~\citep{fu2022complexity}, program-of-thought~\citep{chen2022program}, equation-of-thought~\citep{liu2023plan},  program-aid-language~\citep{gao2023pal}, mathprompter~\citep{imani2023mathprompter}, and code prompting~\citep{hu2023code}. However, it is worth noting that all these methods require annotations, which impose practical limitations on their application.
To address this constraint, Auto-CoT~\citep{zhang2022automatic} proposes a novel approach that utilizes Zero-Shot-CoT~\citep{kojima2022large} to generate CoT reasoning paths. Furthermore, Auto-CoT divides these reasoning paths into different clusters and selects questions that are most closely aligned with the centroid of each cluster. Memory-of-Thought~\citep{li2023mot} selects relevant, high-quality thoughts from external memory during the reasoning process. 
Taking a step further, Tree-of-Thought~\citep{yao2023tree} models the human thought process not only as a chain but also as a tree, whereas Graph-of-Thought~\citep{yao2023chainofthought} extends this concept to represent human thought processes as both chains and graphs. Additionally, Skeleton-of-Thought~\citep{ning2023skeletonofthought} guides LLMs to first create the basic structure of the answer and then uses batched decoding to simultaneously fill in the details of each skeleton. 

\paragraph{Multiple Paths Aggregation}
The DIVERSE approach~\citep{li2022advance} employs a voting verifier to consolidate final answers derived from multiple reasoning paths. In a similar vein, the Self-Consistency method~\citep{wang2023selfconsistency} suggests sampling multiple reasoning paths and making a majority vote to determine the ultimate results. Building on this direction, the concept of complexity-based voting has been introduced, retaining reasoning paths with high complexity for majority voting~\citep{fu2022complexity}.
Furthermore, Model Selection~\citep{zhao2023automatic} takes a different approach by sampling two answers via Chain-of-Thought (CoT) and Plan-of-Thought (PoT) and then employing a Language Model (LLM) to select the correct one. Instead of generating complete reasoning paths, Self-Evaluation Guided Decoding~\citep{xie2023decomposition} samples various reasoning steps at the step level and utilizes beam search to complete the search tree.
One notable limitation of Self-Consistency is its relatively high cost. To mitigate this drawback, Adaptive-Consistency~\citep{aggarwal2023lets} progressively samples reasoning paths until predefined criteria are met. Two concurrent approaches related to Tree-of-Thought~\citep{yao2023tree,long2023large} gradually sample reasoning steps rather than complete reasoning paths.
Additionally, Reasoning via Planning (RAP)~\citep{hao2023reasoning} repurposes the LLM as both a world model and a reasoning agent. It incorporates a principled planning algorithm, based on Monte Carlo Tree Search, to facilitate strategic exploration within the extensive reasoning space.
Exchange-of-Thought~\citep{yin2023eot} and X-of-Thoughts~\citep{liu2023plan} introduce a variety of external reasoning insights and reasoning methods to enhance reasoning performance.

\subsubsection{Multi-Round Prompting}
Multi-round prompting enhances the response through iterative refinement, unlike single-round prompting methods such as Chain of Thought or self-consistency, which do not employ this process of progressive improvement.

\paragraph{Learned Refiners}
The Learned Refiner necessitates a training process, and the acquisition of supervised refinement typically involves pairs of feedback and refinement~\citep{schick2022peer,du2022read,yasunaga2020graph,madaan2021think}. CURIOUS~\citep{madaan2021think} initially constructs a graph that represents relevant influences. This graph is then integrated as an additional input for responding to the question.
PEER~\citep{schick2022peer} is an advanced collaborative language model that replicates the entire writing process, encompassing drafting, suggesting modifications, proposing edits, and providing explanations for its actions.
In contrast, Read, Revise, Repeat (R3)~\citep{du2022read} aims to achieve superior text revisions with minimal human intervention. It achieves this by analyzing model-generated revisions and user feedback, making document revisions, and engaging in repeated human-machine interactions.
DrRepair~\citep{yasunaga2020graph} introduces a program feedback graph that connects symbols relevant to repairing source code with diagnostic feedback. It then employs a graph neural network to model the reasoning process.
Self-Correction~\citep{welleck2022generating} takes an innovative approach by decoupling an imperfect base generator, such as a standard language model or supervised sequence-to-sequence model, from a separate corrector. This corrector learns to refine outputs progressively.
Furthermore, LLM-Augmenter~\citep{peng2023check} continuously enhances LLM prompts to improve model responses by incorporating feedback generated by utility functions, such as the factuality score of an LLM-generated response.

\paragraph{Prompted Refiners}
The REFINER framework~\citep{paul2023refiner} is a comprehensive system designed to fine-tune Language Models (LMs) with the specific goal of generating intermediate reasoning steps, aided by a critic model that automates feedback on the reasoning process. The Self-Refine framework~\citep{madaan2023selfrefine} comprises two vital components: it first generates an output using an LLM and then employs the same LLM to provide feedback on its output through an iterative self-refinement process. Self-Debugging~\citep{chen2023teaching} integrates both LLM and tool feedback to enhance performance. Similarly, Progressive-Hint Prompting (PHP)~\citep{zheng2023progressive} utilizes previous answers as references for generating subsequent responses.
Furthermore, employing distinct prompts for LLMs allows for the assignment of different roles in handling various aspects~\citep{dong2023selfcollaboration,fu2023improving,du2023improving}. \citet{du2023improving} introduce a complementary approach to enhance language responses, involving multiple instances of language models engaging in discussions about their individual responses and reasoning processes over multiple rounds to arrive at a shared final answer. Self-collaboration~\citep{dong2023selfcollaboration} utilizes multiple LLMs as distinct ``experts", each responsible for specific subtasks within complex assignments, defining strategies for collaboration and interaction. \citet{fu2023improving} observe that only a subset of the considered language models demonstrate proficiency in self-improvement through AI feedback, as weaker models may struggle with understanding game rules or incorporating AI feedback for further enhancements.
In conclusion, models exhibit varying abilities to learn from feedback based on their roles, and interactions between LLMs and tools can further enhance reasoning performance~\citep{chen2023teaching,gou2023critic,zhang-etal-2023-self,yang2023intercode,olausson2023demystifying}.

\subsection{Autonomous Agent}
\label{sec: llm with agent}

\begin{figure}[tbp]
	\begin{center}
\includegraphics[width=1.0\columnwidth]{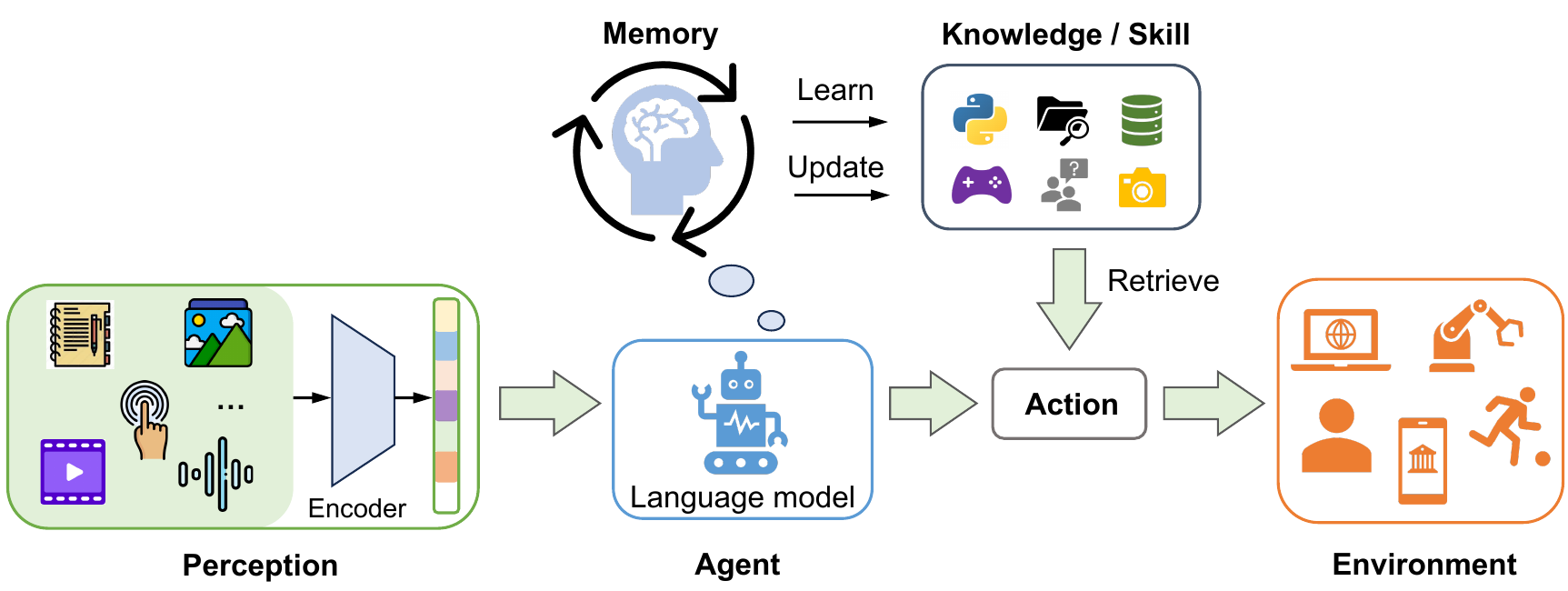}
	\end{center}
	\caption{The general pipeline for LLM with autonomous agent. An LLM agent leverages an LLM as its digital brain mastering multiple abilities and possessing a high-level of intelligence. The agent can receive a diverse set of encoded data as input and correspondingly construct or have access to knowledge bases and skill libraries. With sufficient knowledge and prompts, the agent can work semi-autonomously to operate a spectrum of tasks.}
\label{fig:agent}
\end{figure}

Agents that operate autonomously have often been considered a key route to achieving Artificial General Intelligence (AGI). These agents are adept at performing tasks by independently formulating plans and following instructions. At present, these autonomous entities primarily rely on Large Language Models (LLM) to control and orchestrate various tools~\citep{xi2023rise,wang2023interactive}, including web browsers and code interpreters, to complete their designated tasks, as shown in Figure~\ref{fig:agent}.

VISPROG~\citep{gupta2022visual} is a neuro-symbolic approach for complex visual tasks, using large language models to generate Python-like modular programs without task-specific training. It provides a comprehensive and interpretable rationale. ToolFormer~\citep{schick2023toolformer} is a self-supervised model that decides which APIs to call, when, and with what arguments to incorporate the results into token prediction based on demonstrations. ART~\citep{paranjape2023art} introduces a framework for automatic reasoning and tool use, utilizing frozen LLMs to generate intermediate reasoning steps and seamlessly integrating external tools. CAMEL~\citep{li2023camel} presents a pioneering communicative agent framework known as ``role-playing'', which employs inception prompting to direct chat agents towards achieving tasks while upholding alignment with human intentions.
GPT4Tools~\citep{yang2023gpt4tools} empowers LLMs with multimodal tool capabilities to solve multiple vision tasks.
HuggingGPT~\citep{shen2023hugginggpt} connects AI models for solving tasks, using ChatGPT for task planning and selecting models based on function descriptions in Hugging Face. Chameleon~\citep{lu2023chameleon} augments LLMs with plug-and-play modules for complex reasoning, synthesizing programs by composing various tools for tasks. \citet{wang2023guiding} propose to learn ``planning tokens'', a soft prompt. 
TRICE~\citep{qiao2023making} addresses the challenge of teaching language models when and how to use tools, proposing a two-stage framework for learning through feedback from tool execution. ChatCoT~\citep{chen2023chatcot} presents a tool-augmented chain-of-thought reasoning framework for chat-based LLMs, using multi-turn conversations to integrate thought chains and tool usage naturally. MultiTool-CoT~\citep{inaba2023multitoolcot} leverages chain-of-thought prompting to incorporate multiple external tools during the reasoning process. AssistGPT~\citep{gao2023assistgpt} introduces a multimodal AI assistant with an interleaved code and language reasoning approach, including planning, execution, inspection, and autonomous learning. OpenAGI~\citep{ge2023openagi} is an open-source AGI research platform for real-world tasks, using natural language queries to select and execute appropriate models and proposing a Reinforcement Learning from Task Feedback mechanism. ToolkenGPT~\citep{hao2023toolkengpt} combines the benefits of finetuning LLMs with tool demonstration data and in-context learning, representing tools as tokens (``toolkens'') for flexible tool calls. AutoGPT~\citep{auto2023autogpt} decomposes problems into subproblems and employs tools to solve them. ReAct~\citep{yao2023react} explores the interleaved generation of reasoning traces and task-specific actions for enhanced synergy in language tasks, improving interpretability and trustworthiness. Reflexion~\citep{shinn2023reflexion} reinforces language agents through linguistic feedback and episodic memory, improving decision-making in subsequent trials. CREATOR~\citep{qian2023creator} endows LLMs to create their own tools through documentation and code realization, addressing limitations in tool-using ability. Voyager~\citep{wang2023voyager} is an LLM-powered agent in Minecraft for lifelong learning, incorporating automatic curriculum, skill library, and iterative prompting mechanisms. AutoAgents~\citep{chen2023autoagents} can adaptively generate specialized agents to build a team of agents based on task definitions. SwiftSage~\citep{lin2023swiftsage} is an agent framework inspired by human cognition's dual-process theory, integrating behavior cloning and LLMs for complex reasoning tasks, enhancing problem-solving efficiency. These references cover a wide range of approaches and frameworks for enhancing the capabilities of large language models across various domains.

\section{Discussion: Challenges, Limitations, and Risks}~\label{sec:discussion}

Foundation models have shown promising capabilities in reasoning tasks, opening up new possibilities for the field. It is also essential to acknowledge the challenges, limitations, and risks associated with their use.

\paragraph{Hallucinations}
Despite the promising progress made in foundation models, it is important to acknowledge that these models still face challenges, specifically in relation to the issue of hallucinations~\citep{li2023evaluating,mündler2023selfcontradictory, chen2023can}. Hallucination refers to the generation of outputs by foundation models that contain fabricated or incorrect information, deviating from the intended or expected outputs. These hallucinations can be problematic, as they undermine the reliability and accuracy of the model's generated content.

The hallucination problem in foundation models arises due to various factors. One key factor is the reliance on large-scale pre-training data, which can contain biased or erroneous information. This can lead to the model learning and propagating false patterns or generating unrealistic outputs. Another significant factor contributing to the hallucination issue in foundation models is the models' lack of ability to acknowledge their own knowledge limitations. When confronted with questions beyond their understanding, these models tend to fabricate seemingly plausible answers instead of admitting their lack of knowledge~\citep{yin-etal-2023-large}.

Addressing the hallucination problem in foundation models is an ongoing area of research. Techniques such as fine-tuning task-specific data, incorporating external knowledge sources, and developing advanced evaluation metrics have been explored to mitigate hallucinations. Researchers are also exploring methods to enhance reasoning capabilities in foundation models, enabling them to make more informed and accurate predictions.

It is worth noting that while progress has been made in reducing hallucinations, completely eliminating them remains a challenge due to the inherent complexity of language understanding and generation.

\paragraph{Context Length}
Another limitation is to optimize context length and context construction. For example, GPT models start with 2K window size (GPT-3~\citep{brown2020language}) and go all the way to 32K (GPT-4~\citep{openai2023gpt4}). A longer context window is useful for working with long sequence data, such as gene sequences. By having a larger context window, LLM is capable of handling more lengthy inputs such as entire documents, or comprehending the full scope of an article. This ability enables LLM to produce more contextually relevant responses by leveraging a more comprehensive understanding of the input.

Increasing the context window size in foundation models can bring several benefits, such as capturing longer-range dependencies and improving the model's understanding of context. However, it also comes with certain challenges and costs. In earlier studies, it was observed that the costs associated with larger context window sizes exhibited a quadratic increase as the number of tokens grew~\citep{aryan2023costly}. This means that the computational resources required to process and train the models become significantly higher as the window size grows.
LongNet~\citep{ding2023longnet} represents a modified version of the Transformer model, capable of handling sequences exceeding 1 billion tokens in length, while still maintaining its effectiveness on shorter sequences. LongNet also has a linear computation complexity. Position Interpolation~\citep{chen2023extending} implements a linear downscaling of input position indices to align with the initial context window size during inference. This approach prevents extending beyond the context length trained for, which might otherwise result in abnormally high attention scores and interfere with the self-attention mechanism.

Indeed, while increasing the context window size in language models offers benefits, it is important to consider the tradeoff between window size and generalization ability. Researchers have highlighted that there can be a tradeoff between them~\citep{liu2023lost}.
One challenge worth exploring is how to increase the context window length without sacrificing the model's performance and generalization capabilities. It is crucial to find strategies that allow models to capture longer-range dependencies and context while maintaining their ability to generalize well to new or unseen inputs.

\paragraph{Multimodal Learning}
Multimodal learning is an incredibly powerful but often underappreciated aspect of reasoning. It finds applications in numerous fields where multimodal data is essential, including healthcare (such as CT, X-ray, MRI scans, and gene sequences), robotics, e-commerce, retail, gaming, and entertainment. The integration of different modalities in these domains enables a more comprehensive understanding of the data and facilitates more sophisticated reasoning processes.

One of the key advantages of multimodal reasoning is its potential to significantly improve model performance. While some prior works have delved into multimodal reasoning, such as the multimodal language Model for embodied reasoning proposed by PaLM-E~\citep{driess2023palme} and the visual language model for fear-shot learning known as Flamingo~\citep{alayrac2022flamingo}, there is still ample room for exploring additional data modalities. Incorporating modalities like video, audio, 3D data, and multiple images not only enriches the information available to the models but also opens up exciting possibilities for a more nuanced and comprehensive understanding of the world. 
Other potential applications of foundation model reasoning lie in the domain of Electronic Design Automation (EDA) for program design~\citep{huang2021machine} and Formal Methods~\citep{woodcock2009formal}.

Formal methods, intrinsically linked to logical reasoning, are mathematical strategies employed in the realm of computer science for the design, specification, verification, and analysis of both software and hardware. These techniques are anchored in structured logic, the theory of automata, and other comprehensive mathematical frameworks. They are used to meticulously examine the behavior, accuracy, and dependability of systems. The utilization of formal methods empowers researchers and professionals to guarantee the integrity and precision of intricate systems, establishing them as indispensable in the creation and assessment of software and hardware. The amalgamation of formal methods with foundational models opens doors to augmenting reasoning abilities in the design of software and hardware systems. Formal methods bring to the table precise mathematical methods for defining and confirming system characteristics, whereas foundational models contribute potent language comprehension and reasoning skills. The synthesis of these methodologies can foster the development of more dependable and resilient software and hardware systems.

By leveraging multimodality reasoning and further expanding the exploration of various data modalities, we can unlock new insights and capabilities in reasoning systems. 
It is crucial to recognize and harness the power of multimodal reasoning to fully exploit the potential of reasoning in diverse domains.

\paragraph{Efficiency and Cost}

Efficiency and cost are significant challenges for foundation models for reasoning. Foundation models, especially those with large architectures and extensive training data, can be computationally expensive to train and deploy. The large number of parameters requires more memory and computational resources for processing. This poses challenges in terms of scalability, accessibility, and cost-effectiveness. Efficient reasoning models should be capable of performing fast and real-time inference to meet the demands of interactive applications. However, the complex computations involved in reasoning tasks can lead to slower inference times, hindering real-time performance and user experience. Therefore, it is crucial to enhance the speed and cost-effectiveness of foundation models, making them cheaper and faster.

There are several techniques that can be employed to improve the efficiency of foundation models, including:
\begin{itemize}
    \item Model Pruning~\citep{sun2023simple,Wang_2020}: removing unnecessary connections, parameters, or layers from the model. This results in a more compact architecture, reducing computational requirements.
    \item Compression~\citep{zhu2023survey} and Quantization~\citep{tao2022compression}: reduce the size of the model or reduce the precision of model parameters, using fewer bits to represent them. This reduces memory usage and computational complexity.
    \item Knowledge Distillation~\citep{gu2023knowledge}: training a smaller model (student) to mimic the behavior and predictions of a larger model or ensemble of models (teacher). This transfer of knowledge allows for efficient inference with reduced computational resources.
    \item Low-Rank Factorization~\citep{ren2023lowrank,hsu2022language}: replace high-dimensional tensors with lower-dimensional tensors. By reducing the number of parameters, these methods improve efficiency without significant loss in performance.
\end{itemize}

By employing these techniques, it is possible to enhance the efficiency of foundation models, making them faster and more cost-effective for various reasoning tasks and applications.

\paragraph{Human Preference}
Addressing the risks and potential harms associated with foundation models, such as bias, unfairness, manipulation, and misinformation, requires careful consideration and proactive measures. One approach is to focus on improving learning from human preference and feedback to ensure more responsible and accurate model behavior.

To mitigate these risks, we can explore several strategies. First, we need mechanisms to incorporate diverse perspectives and mitigate bias during the training and fine-tuning phases of foundation models. This can involve diverse data collection, representative sampling, and careful annotation processes that involve input from a wide range of human perspectives. 
Continual learning and adaptation, informed by human feedback, can also play a crucial role. By enabling ongoing interactions between models and human annotators or users, we can gather feedback and iteratively refine the models' behavior. This iterative process helps identify and rectify potential biases, unfairness, or misinformation, allowing the models to improve over time. 
Furthermore, ensuring that the outputs of foundation models align with real-world evidence, experimental findings, and explicit knowledge is essential. This requires incorporating robust fact-checking mechanisms and validation processes into the model training pipeline. Additionally, leveraging external sources of information, such as trusted databases or expert knowledge, can help verify and validate the outputs generated by the models.

Constitutional AI, as proposed by \citet{bai2022constitutional}, offers an approach that involves both supervised learning and reinforcement learning phases like ``RL from AI Feedback" (RLAIF). 
Similarly, \citet{bakker2022finetuning} explore the use of fine-tuning a large language model (LLM) with 70 billion parameters to generate statements that maximize the expected approval for people with different and diverse perspectives. This approach emphasizes the importance of incorporating human preferences and diverse viewpoints during the model training process.

By integrating these techniques and approaches, we can work towards mitigating the risks and potential harms associated with foundation models. Improving learning from human preference, continual learning informed by feedback, and ensuring fidelity to real-world evidence are challenging steps in building more responsible and trustworthy reasoning systems.

\paragraph{Multilingual Support}

While reasoning itself is a language-agnostic process, the availability of comprehensive knowledge sources is often limited to a few languages, primarily English. Historically, language foundation models have demonstrated exceptional reasoning performance, primarily in English, with relatively limited support for other languages such as Chinese and Japanese. Currently, there is a lack of robust multilingual reasoning language foundation models that excel across various languages.

\citet{fang-etal-2022-leveraging} propose utilizing English as a pivot language in their commonsense reasoning framework. They employ a translate-retrieve-translate (TRT) strategy, leveraging English knowledge sources to enhance their reasoning capabilities.
Furthermore, \citet{huang2023languages} introduce cross-lingual thought prompting (XLT) as a systematic approach to improving the multilingual capabilities of Language and Reasoning Models (LLMs).

Given these advancements, there is a growing interest in developing foundation models dedicated to multilingual reasoning. Building robust models that excel in multilingualism presents an intriguing avenue for future research and development.

In summary, to address these challenges, ongoing research and development efforts are necessary. This includes advancing the deployment of reasoning models.

\section{Future Directions}~\label{sec:future}

Further research and development in this area have the potential to unlock even more advanced reasoning abilities in foundation models.

\subsection{Safety and Privacy}
The rise of foundation models and their application to reasoning tasks has highlighted the critical need to ensure their safety and trustworthiness~\citep{huang2023survey}. 

Various intended attacks have been identified, including the robustness gap~\citep{shreya2022survey}, backdoor attacks~\citep{shen2021backdoor,kurita2020weight}, poisoning~\citep{carlini2023poisoning}, disinformation~\citep{nelson2008exploiting}, privacy leakage~\citep{li2023multi}, and unauthorized disclosure of information~\citep{perez2022ignore}. Specifically, backdoor attacks involve the injection of malicious knowledge into foundation models through techniques such as poisoning the training data~\citep{shen2021backdoor} or modifying model parameters~\citep{kurita2020weight}.

As one of the most principled techniques for training machine learning models with privacy, differential privacy allows for training on datasets without revealing any details of individual training examples, providing enhanced privacy protection~\citep{shi2022just,behnia2022ew}. Another effective way of defensing adversarial attacks is by adversarial training, which can provide another layer of security when facing malicious yet human invisible perturbations added in model inputs~\citep{li2023aroid,li2023data}.

In response to some copyright concerns, \citet{kirchenbauer2023watermark} introduce a watermarking framework specifically designed for proprietary language models. This framework enables the embedding of watermarks with minimal impact on text quality and facilitates their detection using an efficient open-source algorithm, eliminating the need for accessing the language model API or parameters.

\subsection{Interpretability and Transparency}

Additionally, there is a need for increased transparency and interpretability of foundation models~\citep{liao2023ai}. As these models become more complex and sophisticated, understanding their reasoning processes and the factors influencing their outputs becomes increasingly important.

Sometimes, foundation models generate toxic content, which may incite violence and cause infordemic~\citep{Bender2021OnThe, weidinger2021ethical}. They can inadvertently disclose sensitive information, thereby jeopardizing privacy and security. Additionally, LLMs can contribute to the dissemination of misinformation, both intentionally and unintentionally~\citep{pan2023risk, buchanan10truth, kreps2022all, zhou2023synthetic}.
The complex and uncertain nature of foundation models further compounds these challenges. These models exhibit a remarkable capacity to perform a wide range of tasks across diverse contexts~\citep{bommasani2021opportunities}. However, their massive and opaque architectures hinder a comprehensive understanding of their capabilities and behaviors, making it difficult to ascertain their decision-making processes and potential biases. This lack of transparency raises concerns regarding model interpretability, control, and accountability.

Developing techniques and frameworks for model interpretability can help address concerns regarding transparency and accountability.

\subsection{Autonomous Language Agents}
The capacity for logical reasoning is crucial in achieving complex tasks in embodied environments, and it plays a significant role in embodied intelligence~\citep{dasgupta2022collaborating}. Foundation Models have exhibited powerful capabilities for reasoning and flexibility through the process of in-context learning~\citep{yang2023foundation}.
Recent studies, such as Voyager and DEPS, have explored the use of LLMs for planning in Minecraft~\citep{wang2023voyager,wang2023describe}. DEPS specifically proposes an interactive planning approach based on LLMs~\citep{wang2023describe}.
LLMs have also shown promise in generating action sequences directly based on natural language instructions without requiring extra domain knowledge~\citep{li2022pre}. 
Equipping embodied agents with commonsense knowledge is crucial for their successful completion of complex human instructions in diverse environments~\citep{wu2023embodied}.

In the context of reasoning for autonomous agents, there are key characteristics:
\begin{itemize}
    \item Infinite Task Capability: Foundation models empower agents with the capacity to handle an extensive range of tasks, even those that are not pre-defined or anticipated in advance. This flexibility allows agents to dynamically generate tasks based on their understanding of the context and the specific needs of the users. 
    \item Autonomous Task Generation: 
    Foundation model reasoning enables agents to autonomously generate new tasks within a given context. This capability empowers agents to take initiative, identify opportunities, and propose relevant tasks to users. They can adapt and respond to changing circumstances, making them more versatile, proactive, and efficient in fulfilling user requirements.
    \item Value System: Autonomous agents are driven by a value system empowered by a pre-trained foundation model, which serves as the foundation for task generation. This value system guides the agent's decision-making process, taking into account factors such as priorities, preferences, and ethical considerations. By leveraging the capabilities of the foundation model, agents can make informed decisions aligned with human values, ensuring responsible and ethical behavior.
    \item World Model: The foundation model can also be utilized as a world model that represents the real world and serves as the basis for agents' interactions and reasoning. This comprehensive model enables agents to understand the context, interpret natural language inputs, and generate appropriate responses or actions. With the foundation model as their world model, agents can effectively navigate and operate within their environment, enhancing their ability to interact intelligently and respond to user needs.
\end{itemize}

By leveraging foundation models, autonomous agents can facilitate more meaningful and effective interactions with users, better understand their intents and needs, and generate relevant tasks accordingly. This approach opens up promising avenues for research in areas such as contextual understanding, human-like reasoning, and personalized assistance. Ultimately, it enhances the overall user experience and enables the development of more sophisticated and intelligent AI systems.

Given their reasoning capabilities, Foundation Models hold significant potential for applications in human-computer interaction and embodied intelligence. can be leveraged to create interactive and adaptive systems that can dynamically respond to user input and adapt their behavior accordingly. This involves developing models that can learn from user interactions and update their knowledge and behavior over time. By enabling Foundation Models to actively engage with users and adapt to their preferences and needs, we can create more personalized and user-centric human-computer interaction experiences.

\subsection{Reasoning for Science}
Future work can also build upon the research on temporal reasoning in multimodal question-answering tasks or sound reasoning~\citep{brandt2011sound}, as demonstrated by Audio Question Answering (AQA)~\citep{fayek2019temporal}. 
Researchers can delve deeper into understanding and developing foundation models that can reason and make inferences based on auditory information. 
This can have implications in areas such as audio-based decision-making systems, environmental monitoring, and audio scene understanding.

Furthermore, the application of multimodal reasoning can be extended to domains like medical reasoning and diagnosis, particularly in the context of gene sequence analysis. This can aid in the identification of genetic disorders, personalized medicine, and the exploration of potential treatments.

Overall, future work can focus on advancing multimodal reasoning abilities in foundation models.
These endeavors can contribute to the development of more intelligent and context-aware systems in various fields.

\subsection{Super Alignment}
Superintelligence alignment, according to OpenAI\footnote{\url{https://openai.com/blog/introducing-superalignment}}, is the next machine-learning question of utmost importance.
However, ensuring the control and alignment of potentially superintelligent AI systems poses significant challenges. Current techniques, such as Reinforcement Learning from Human Feedback (RLHF), heavily rely on human supervision and reasoning. As AI systems surpass human intelligence, human supervision becomes inadequate, necessitating new scientific and technical breakthroughs in alignment research. Existing alignment techniques will not scale to superintelligence due to the limitations of human reasoning and supervision. The prospect of controlling and steering highly intelligent AI systems to prevent them from going rogue remains an unsolved challenge. Without reliable means of supervising these reasoning systems surpassing human capabilities, ensuring their alignment with human intent becomes increasingly difficult.

One approach to address the challenge of ensuring that reasoning systems surpassing human intelligence adhere to human intent is to develop a roughly human-level automated alignment researcher. By creating such a system, it becomes possible to leverage extensive computational resources to scale alignment efforts and iteratively align superintelligence.

\section{Conclusion}\label{sec:conclusion}
This survey illuminates the evolutionary path of foundation models in the field of reasoning, showcasing a discernible progression in complexity and efficacy from their initial stages to current advancements. While we acknowledge the remarkable strides made in data-driven thinking, it is crucial for us to objectively recognize both the strengths and limitations of large models. Emphasizing the importance of enhancing their interpretability and security becomes imperative in this context. We also note that with all the papers surveyed in this work, a consensus is yet to be reached on how to push forward the reasoning ability of foundation models to a consistently superhuman level (which can for instance win an IMO medal or even solve open mathematical problems).

In conclusion, while foundation models offer exciting possibilities in reasoning tasks, it is essential to approach their development and application with a critical perspective. It is crucial to acknowledge the challenges, limitations, and risks associated with LLM-based reasoning. By doing so, we can foster responsible and thoughtful advancements in this field, ensuring the development of robust and reliable reasoning systems.

\newpage

\nocite{*}
\bibliography{sn-bibliography}

\end{document}